%% file: main.tex
\documentclass[11pt]{article}
\usepackage[left=1in,top=1in,right=1in,bottom=1in,letterpaper]{geometry}

\usepackage[utf8]{inputenc} %
\usepackage[T1]{fontenc}    %
\usepackage{hyperref}       %
\usepackage{url}            %
\usepackage{booktabs}       %
\usepackage{amsfonts}       %
\usepackage{nicefrac}       %
\usepackage{microtype}      %
\usepackage{xcolor}         %
\usepackage[most]{tcolorbox}
\usepackage{enumitem}
\usepackage{natbib}
\usepackage{subfigure}
\usepackage{placeins}
\usepackage{tikz}
\usetikzlibrary{matrix,decorations.pathreplacing}
\usetikzlibrary{positioning, arrows.meta, shapes.geometric, fit}
\usepackage{float}
\usepackage[most]{tcolorbox}
\usetikzlibrary{shadows}

\newcounter{exa}

\usepackage{amsmath}
\usepackage{amssymb}
\usepackage{mathtools}
\usepackage{amsthm}

\usepackage{subcaption}
\usepackage{keyval}

\usepackage[capitalize,noabbrev]{cleveref}

\input{notations}
\theoremstyle{plain}
\newtheorem{theorem}{Theorem}[section]
\newtheorem{proposition}[theorem]{Proposition}
\newtheorem{lemma}[theorem]{Lemma}
\newtheorem{corollary}[theorem]{Corollary}
\theoremstyle{definition}
\newtheorem{definition}[theorem]{Definition}

\theoremstyle{remark}

\title{Transformers Provably Learn Graph Search: Training Dynamics and the Exponential Power of Depth}
\author{
Xutao Ma\qquad
Somayeh Sojoudi
\\
\\
UC Berkeley
\\
\\
\texttt{\{xutao\_ma,sojoudi\}@berkeley.edu}
}
\date{\today}

\begin{document}

\maketitle
\begin{abstract}
    Large Language Models (LLMs) have achieved remarkable progress on complex reasoning tasks, yet the mechanisms by which they acquire reasoning capabilities and the efficiency of their reasoning remain poorly understood. In this paper, we investigate the path-finding problem over directed graphs—a symbolic abstraction of multi-step reasoning. We first characterize the architectural requirements for this task, proving that one-layer transformers require $\Omega(N^2)$ size on dense graphs to implement the key DFS child-selection primitive. While a two-layer transformer implements full DFS with $O(N\log(N))$ size. Thus, a second layer is necessary and sufficient for near-linear DFS reasoning. We then extend this expressivity analysis to deeper models: for every $k\geq3$, a $k$-layer transformer implements a bounded-reachability DFS policy with look-ahead horizon $1+2^{k-3}$, establishing an exponential increase in reachability horizon with depth.
    Finally, going beyond expressivity, given appropriately curated training data, we show that two-layer transformers can provably learn to execute DFS via gradient flow and generalize to unseen graphs. Together, our results provide a mechanistic account of how transformers implement graph search, how depth expands bounded look-ahead, and how such reasoning algorithms can emerge through training. We provide the experiment code and  Lean 4 proofs of all the theoretical results.\footnote{ \url{https://github.com/xutaoma/transformer_search_code_lean}.}
\end{abstract}

\input{content/intro}

\input{content/setting}
\input{content/expressivity}
\input{content/learning_dynamics}
\input{content/experiment}
\input{content/conclusion}

\newpage
\input{acknowledgement}

\bibliography{references}
\bibliographystyle{plainnat}

\newpage
\appendix 

\input{content/related_work}
\input{content/appendix/proof_expressivity}

\input{content/appendix/proof_learning_dynamics}

\input{content/appendix/addition_exp}

\newpage

\end{document}

%% file: notations.tex
\usepackage{bbm, dsfont}

\usepackage{amsmath,amsfonts,bm}

\def\eqref#1{equation~\ref{#1}}

\def\1{\bm{1}}

\def\va{{\bm{a}}}
\def\vb{{\bm{b}}}
\def\vc{{\bm{c}}}
\def\vd{{\bm{d}}}
\def\ve{{\bm{e}}}

\def\vx{{\bm{x}}}
\def\vy{{\bm{y}}}

\def\mA{{\bm{A}}}
\def\mB{{\bm{B}}}

\def\mD{{\bm{D}}}
\def\mE{{\bm{E}}}

\def\mG{{\bm{G}}}

\def\mI{{\bm{I}}}

\def\mP{{\bm{P}}}

\def\mW{{\bm{W}}}
\def\mX{{\bm{X}}}

\DeclareMathAlphabet{\mathsfit}{\encodingdefault}{\sfdefault}{m}{sl}
\SetMathAlphabet{\mathsfit}{bold}{\encodingdefault}{\sfdefault}{bx}{n}

\def\gE{{\mathcal{E}}}
\def\gF{{\mathcal{F}}}
\def\gG{{\mathcal{G}}}

\def\gK{{\mathcal{K}}}
\def\gL{{\mathcal{L}}}

\def\gR{{\mathcal{R}}}

\def\gV{{\mathcal{V}}}

\def\sE{{\mathbb{E}}}

\def\sR{{\mathbb{R}}}

\newcommand{\dx}{\dot{x}}
\newcommand{\dy}{\dot{y}}
\newcommand{\ind}{\mathds{1}}
\newcommand{\relu}{\text{ReLU}}
\newcommand{\Atten}{\text{Atten}}
\newcommand{\TF}{\text{TF}}

\newcommand{\dd}{\,\mathrm d}

\newcommand{\Head}{\operatorname{Head}}

\newcommand{\INDEX}{\textsc{INDEX}}

\newcommand{\poly}{\operatorname{poly}}
\newcommand{\softmaxcol}{\operatorname{softmax}_{\rm col}}
\newcommand{\ones}{\bm{1}}

\newcommand{\Tar}{\text{Tar}}
\newcommand{\Root}{\text{Root}}
\newcommand{\Prompt}{\text{Prompt}}

\DeclareMathOperator*{\argmax}{arg\,max}

\DeclareMathOperator{\Tr}{Tr}

\newcommand{\Eoff}{{\mE}_{\text{off}}}
\newcommand{\bone}{\mathbf{1}}

\newif\ifdraft
\draftfalse

\ifdraft
    \newcommand{\comments}[1]{\textcolor{violet}{[xutao: #1]}}
\else 
    \newcommand{\comments}[1]{}
\fi

%% file: content/intro.tex
\section{Introduction}

Large Language Models (LLMs) have demonstrated striking reasoning capabilities across mathematics, code, scientific question answering, and planning, especially when equipped with chain-of-thought (CoT) prompting~\citep{wei2022chain} and test-time scaling of intermediate reasoning steps~\citep{guo2025deepseek}. Yet our theoretical understanding of these capabilities remains incomplete. We still lack precise answers to several basic questions: how can a transformer implement a multi-step reasoning algorithm, why does architectural depth improve reasoning efficiency, and under what conditions can such an algorithm be acquired through gradient-based training?

A growing body of theoretical work studies these questions through symbolic reasoning tasks, where the algorithmic structure is explicit enough to analyze. Prior results have clarified important aspects of transformer expressivity, showing both limitations of constant-depth models on inherently sequential tasks and the additional power obtained from chain-of-thought computation~\citep{feng2023towards,li2024chain,merrill2024expressive}. Other work has examined depth--width tradeoffs for graph algorithms~\citep{sanford2024understanding,yehudai2025depth}, sample complexity for reasoning tasks~\citep{wies2023subtask}, and training dynamics for algorithmic primitives such as induction heads, sparse token selection, in-context regression, and Markov-chain structure learning~\citep{tian2023scan,nichani2024transformers,chen2024unveiling,wang2024transformers}.

Despite this progress, graph search remains poorly understood. This setting is especially important because it captures a central difficulty of reasoning: the model must not merely follow a single predetermined chain, but must choose among alternatives, remember what has already been explored, and backtrack when a branch fails. In other words, graph search formalizes the kind of committed exploration and revision that appears in planning, proof search, and multi-step deduction.

Existing theory leaves three gaps. First, most provable learning results for path-finding focus on trees, where there is at most one path between two vertices and the need for genuine search is largely absent~\citep{yang2025multihead}. Directed graphs are qualitatively harder: they contain branching, cycles, and dead ends, so a successful algorithm must maintain a visited set and decide when to move forward or backtrack. Second, although prior work shows that constant-depth transformers can implement DFS~\citep{merrill2024expressive}, the architectural threshold for efficient graph search is not known. It is unclear whether one layer suffices, why two layers should be special, and how additional layers translate into improved search efficiency. Third, recent work has analyzed reachability with continuous chain of thought~\citep{zhu2025emergence}, but the corresponding learnability question for standard discrete CoT traces remains open.

In this paper, we give a theoretical account of how transformers implement, scale, and learn depth-first search (DFS) on directed graphs. We focus on DFS because it is the canonical sequential search algorithm: it combines branching, memory of visited states, forward exploration, and backtracking, while naturally producing a discrete CoT trace. Our results connect three aspects of reasoning---architecture, depth, and training dynamics---within a single graph-search framework.

\paragraph{Our contributions.}

\begin{itemize}[leftmargin=*]
    \item \textbf{We identify a sharp depth threshold for efficient DFS.}
    We prove that a one-layer transformer cannot implement even a basic DFS child-selection primitive on dense directed graphs without quadratic representation size, measured as heads $\times$ width $\times$ precision. In contrast, we construct a two-layer transformer that implements full DFS with near-linear representation size. The separation concerns the width a single attention layer needs, not parameter count or running time, and it shows that the second layer is not merely a convenience: it is necessary and sufficient for exact DFS with sub-quadratic width in our setting.

    \item \textbf{We show that depth gives an exponential increase in look-ahead horizon.}
    Beyond the two-layer DFS construction, we prove that additional layers can perform parallel bounded-reachability computation. After explicit neighbor aggregation, each propagation layer doubles the radius represented in source and child memories. For every $k\geq3$, a $k$-layer transformer implements a bounded-reachability DFS policy with horizon $1+2^{k-3}$. This establishes a depth--horizon tradeoff. We distinguish it from worst-case search-efficiency guarantees, which require additional control over the distances of the selected branches.

    \item \textbf{We prove that DFS behavior can emerge through gradient-based training.}
    For a two-layer transformer trained on a permutation-symmetric DFS curriculum, we analyze the gradient-flow dynamics of the relevant attention parameters. The dynamics preserve a low-dimensional invariant structure, and the resulting model generalizes from the training curriculum to unseen graphs and DFS traces. Thus, the construction is not only expressive: it can be learned under a mathematically analyzable training process.
\end{itemize}

Together, these results provide a mechanistic account of transformer-based graph search. They show that efficient DFS requires a genuine two-layer mechanism, that additional depth yields exponentially larger look-ahead, and that the corresponding search behavior can arise from gradient-based training on discrete chain-of-thought traces. More broadly, our analysis uses directed-graph path finding as a clean theoretical proxy for understanding how reasoning capabilities depend on architecture, depth, and training data.

%% file: content/setting.tex
\section{Preliminaries}
\label{sec:preliminaries}

\paragraph{Notation.}
For $N\in\mathbb{N}$, let $[N]=\{1,\ldots,N\}$. We use {$\mathcal{A}$} to denote the token vocabulary, reserving $\gV$ for the vertex set of a graph. For a finite set $\gK$, let $S_{\gK}$ denote the set of all permutations on $\gK$. We write $\tau_\Prompt$ for a prompt, viewed as a sequence of tokens. Throughout the paper, $\mI$ denotes the identity matrix, and $\mE$ denotes an all-ones matrix of the appropriate size.

\paragraph{Directed graphs and permutations.}
A directed graph of order $N$ is an ordered pair $\gG=(\gV,\gE)$, where $\gV=[N]$ is the vertex set and $\gE\subseteq \gV\times\gV$ is the edge set. A path from a source node $n_s$ to a target node $n_t$ is a sequence of vertices
\[
    p=(n_1,n_2,\ldots,n_k)
\]
with $n_1=n_s$ and $n_k=n_t$, such that $(n_i\to n_{i+1})\in\gE$ for every $i\in[k-1]$. A shortest path is any such path with the minimum number of vertices. For any permutation $\pi$ of $\gV$, we define the permuted graph by
\[
    \pi(\gG)=(\gV,\pi(\gE)),\qquad
    \pi(\gE)=\{(\pi(u)\to\pi(v)):(u\to v)\in\gE\},
\]
and define the permuted path as $\pi(p)=(\pi(n_1),\ldots,\pi(n_k))$.

\paragraph{Path-finding problem and depth-first search.}
Given a directed graph $\gG$, a source node $n_s$, and a target node $n_t$, the path-finding problem asks for a path from $n_s$ to $n_t$, if one exists. Depth-first search (DFS) is a recursive search procedure for this problem. Starting from $n_s$, DFS repeatedly explores outgoing edges from the current node $n$. If there exists an unvisited child $v$ with $(n\to v)\in\gE$, the search moves to $v$, marks it as visited to avoid cycles, and continues recursively. If no unvisited child remains, the search backtracks to the parent of $n$ in the current search stack and explores alternative branches. This forward-and-backtracking process terminates either when $n_t$ is found, yielding a valid $n_s\to n_t$ path, or when the search returns to $n_s$ after exhausting all reachable branches.

\paragraph{Multi-head transformer architecture.}
An $L$-layer transformer with $H$ attention heads is a sequence-to-sequence model. Given an input sequence $\mX=[\vx_1,\ldots,\vx_T]\in\sR^{d\times T}$, the model computes hidden states $\mX^{(l)}\in\sR^{d\times T}$ as follows. Let $\mX^{(1)}=\mX$, and for every $l\in[L]$,
\begin{align}
    \mX^{(l+1)} &= \mX^{(l)} + \sigma\!\left(\Atten^{(l)}(\mX^{(l)})\right),\\
    \Atten^{(l)}(\mX^{(l)}) &= \mW_O^{(l)}
    \begin{bmatrix}
        \mathrm{Head}_1^{(l)}(\mX^{(l)})\\
        \vdots\\
        \mathrm{Head}_H^{(l)}(\mX^{(l)})
    \end{bmatrix},\\
    \mathrm{Head}_h^{(l)}(\mX^{(l)})
    &= \mW_V^{(l,h)}\mX^{(l)}
    \phi^{(l)}\!\left((\mX^{(l)})^T(\mW_K^{(l,h)})^T\mW_Q^{(l,h)}\mX^{(l)}\right).
\end{align}
Here $\sigma$ is an element-wise activation function; throughout, we take $\sigma(x)=\relu(x)=\max\{0,x\}$. The map $\Atten^{(l)}$ denotes the attention block at layer $l$, $\mW_O^{(l)}$ is the output projection, $\mW_V^{(l,h)}$, $\mW_K^{(l,h)}$, and $\mW_Q^{(l,h)}$ are the value, key, and query matrices for head $h$, and $\phi^{(l)}$ is an element-wise attention activation, possibly with normalization.

The transformer output logits are
\begin{align}
    \TF(\mX;\theta)=\mW_R^T\mX^{(L+1)},
\end{align}
where $\theta$ denotes the trainable parameters and $\mW_R\in\sR^{d\times |\mathcal{A}|}$ is the readout matrix. The next-token distribution is the softmax over these logits:
\begin{align}
    p_\theta(v;\mX)=
    \frac{\exp(\TF(\mX;\theta)_v)}{\sum_{v'\in\mathcal{A}}\exp(\TF(\mX;\theta)_{v'})},
\end{align}
where $\TF(\mX;\theta)_v$ is the logit assigned to token $v$.

\paragraph{Prompt format and chain of thought.}
Following~\citet{zhu2025reasoning,zhu2025emergence}, the transformer path-finding prompt has three parts: (1) a beginning-of-sequence token \textsf{<s>}; (2) a graph-description segment containing $M$ edges, where each edge is represented by a source token $s_i$, a target token $t_i$, and a special edge token \textsf{<e>}; and (3) a task-description segment containing the target token $n_{\Tar}$, a start-reasoning token \textsf{<r>}, and the root token $n_{\Root}$. We denote the full prompt by $\tau_\Prompt$, \emph{i.e.},
\begin{align}
    \tau_\Prompt = \textsf{<s>}, s_1, t_1,  \textsf{<e>},\cdots,s_M, t_M,  \textsf{<e>},n_{\Tar},\textsf{<r>},n_{\Root}.
\end{align}
Starting from $n_{\Root}$, the transformer predicts tokens autoregressively, generating a chain-of-thought (CoT) trace $n_1,n_2,\ldots,n_T$. Generation terminates when the model reaches the target $n_{\Tar}$, or when it returns to the root $n_{\Root}$ after all of the root's child nodes have been explored.

%% file: content/expressivity.tex
\section{Expressivity}
\label{sec:expressivity}

Although~\citet{merrill2024expressive} showed that constant-depth transformers can perform DFS, the minimal architectural requirements remain unclear. In this section, we first prove that any one-layer transformer that performs DFS on directed graphs of order $N$ must have {$\Omega(N^2)$} \emph{representation size}, i.e., heads $\times$ width $\times$ precision, in the dense-graph regime. This lower bound shows that at least two layers are needed for DFS with sub-quadratic representation size. We then give a two-layer construction with near-linear representation size, and finally show that additional layers provide an exponential improvement in look-ahead depth.

\subsection{One-layer lower bound for DFS child selection}

To formalize the size measure, we first specify the numerical precision of a transformer.
\begin{definition}[$p$-bit precision]
A transformer uses $p$ bits of precision if all coordinates produced by $\mW_K$, $\mW_Q$, $\mW_V$, $\mW_O$, and the attention heads, after rounding, can be represented with $p$ bits without changing the computed input-output function. As in~\citet{sanford2024one}, attention sums are accumulated in $p$-bit fixed-point arithmetic: for an input of length $T$, every unnormalized attention weight $w_j$ and every summand $w_j\mW_V\vx_j$ of an attention numerator is rounded to a $p$-bit fixed-point number, and $p\geq\log_2T$. Consequently, the partial sum of an attention numerator or normalizer over any subset of positions is exactly representable with $p+\lceil\log_2T\rceil=O(p)$ bits per coordinate. The value dimension of every head is at most $d$.
\end{definition}
\paragraph{Size measure.}
As in~\citet{sanford2024one}, we quantify a transformer layer by $Hdp$, the number of heads times the embedding dimension times the precision, and call this its \emph{representation size}; it is the quantity that \citet{sanford2024one} call simply the \emph{size}, and we use the longer name because it is a measure of width and precision rather than of parameters or running time. It is the number of bits that one layer of attention can deliver to a single position, which is exactly the quantity controlled by the communication-complexity argument below, and we claim no separation in parameter count or running time. Dense $d\times d$ projection matrices have $\Theta(d^2)=\Theta(N^2)$ entries in the one-layer and in the two-layer setting alike (the matrices of our construction are built from identity and all-ones blocks, but they are still $N\times N$), and on a dense graph the prompt itself has length $\Theta(N^2)$, so any model processes $\Theta(N^2)$ tokens. The separation below is therefore a statement about depth versus width: a single attention layer must route $\Omega(N^2)$ bits to the output position, whereas with two layers $O(N\log N)$ bits per layer suffice, which is within a logarithmic factor of the $\Omega(N)$ bits needed to represent the visited set at all (\cref{app:token_embeddings}).

We next define a restricted task that isolates the retrieval of an outgoing child.
\begin{definition}[Unique outgoing-child task]
Fix $N,M\in\mathbb{N}$. Let $\gG=([N],\gE)$ be a directed graph with exactly $M$ distinct edges, $\gE=\{(s_i\to t_i):i\in[M]\}$. The prompt is $\textsf{<s>},s_1,t_1,\textsf{<e>},\ldots,s_M,t_M,\textsf{<e>},q$, an enumeration of $\gE$ followed by a query node $q$, under the promise that exactly one edge has source $q$. The desired output is the target of this unique outgoing edge.
\end{definition}
In the hard instances constructed below the unique outgoing edge of $q$ is not a self-loop, so its target is the mandatory first DFS move from $q$, and the reduction applies to the full DFS next-step task as well.

\begin{theorem}[One-layer representation-size lower bound]
\label{thm:lower_bound}
Let $M\geq 1$ and $N\geq 3\lceil\sqrt{M}\rceil+1$.
Fix any scalar weighting function $\phi:\mathbb{R}\to\mathbb{R}$.
Any one-layer transformer with the architecture of~\cref{sec:preliminaries}, with element-wise attention activation $\phi$, either normalized or unnormalized, with $H$ heads, embedding dimension $d$, and $p$-bit precision that is correct on every valid instance of the unique outgoing-child task whose graph has vertex set $[N]$ and exactly $M$ distinct edges satisfies $Hdp=\Omega(M)$.
\end{theorem}

\begin{proof}
See~\cref{app:proof_thm_lower_bound}.
\end{proof}

The proof adapts the communication-complexity argument for one-layer lower bounds on induction heads~\citep{sanford2024one}: we reduce from \textsc{INDEX}~\citep{kushilevitz1997communication} by placing the $M$ records in distinct cells of an $m\times m$ grid, $m=\lceil\sqrt M\rceil$, so that the source of a record encodes its row and the target encodes its column and Alice's bit; Bob changes only the source of the selected record to the query vertex $q$. This yields exactly $M$ distinct edges on $3m+1$ vertices, and since Alice controls only the target-token positions while the final query is fixed, each head contributes an $O(dp)$-bit summary to a one-way protocol.

\begin{corollary}[Dense-graph lower bound]
\label{cor:dense_graph_lower_bound}
For $N\geq 4$, set $m_N=\lfloor (N-1)/3\rfloor$ and $M_N=m_N^2$.
Any one-layer transformer in the architecture of~\cref{thm:lower_bound} that is correct on every valid instance of the unique outgoing-child task whose graph has vertex set $[N]$ and exactly $M_N$ distinct edges requires $Hdp=\Omega(N^2)$. Consequently, any one-layer transformer in this architecture that solves the full DFS next-step task on all $N$-vertex directed graphs requires $\Omega(N^2)$ representation size in the worst case, already on a family with $\Theta(N^2)$ distinct edges.
\end{corollary}
\begin{proof}
See~\cref{app:proof_cor_dense}.
\end{proof}

\subsection{Two-layer near-linear construction for exact DFS}

\paragraph{Token embeddings and the first layer.}
Each token is embedded in four $N$-dimensional node blocks (plus a fixed positional component): edge endpoints $s_i,t_i$ and the target $n_\Tar$ place their one-hot vector in the fourth block, the root and the CoT tokens place theirs in the first block, and the special tokens are zero (\cref{app:token_embeddings}). The $O(N)$ dimension is within a logarithmic factor of what exact DFS requires, since the model must distinguish $2^N$ visited sets (\cref{app:token_embeddings}). The first layer only copies information from the preceding token.
\begin{lemma}[Predecessor-copy layer, informal version of~\cref{lem:first-layer-softmax-copy-relu}]
Assume that the input embeddings are augmented with a sinusoidal positional encoding, or any positional encoding that permits a constant-size relative-position selector. Then, for every leakage parameter $\delta\in(0,1/8)$, there exists a one-layer transformer block with $O(1)$ attention heads, embedding dimension $O(N)$, and parameter precision $O(\log N)$ that copies the first block of the preceding token into the second block and its fourth block into the third block, each one-hot vector scaled by a factor in $[1-2\delta,1]$ and all other copied coordinates exactly zero after the ReLU. \end{lemma}
Concretely, up to the scalar factors in $[1-2\delta,1]$ on the copied one-hot vectors, the hidden states entering the second layer are
\begin{align}
\label{eq:hidden_state_after_first_layer}
\text{Edge source } s_i &: \begin{pmatrix} 0_N \\ 0_N \\ 0_N \\ s_i \end{pmatrix}, &
\text{Edge target } t_i &: \begin{pmatrix} 0_N \\ 0_N \\ s_i \\ t_i \end{pmatrix}, &
\textsf{<s>} &: \begin{pmatrix} 0_N \\ 0_N \\ 0_N \\ 0_N \end{pmatrix},\quad
\textsf{<e>} : \begin{pmatrix} 0_N \\ 0_N \\ t_i \\ 0_N \end{pmatrix},  \\
\text{Root } n_\Root &: \begin{pmatrix} n_\Root \\ 0_N \\ 0_N \\ 0_N \end{pmatrix}, &
\text{Target } n_{\Tar} &: \begin{pmatrix} 0_N \\ 0_N \\ 0_N \\ n_{\Tar} \end{pmatrix}, &
\textsf{<r>} &: \begin{pmatrix} 0_N \\ 0_N \\ n_{\Tar} \\ 0_N \end{pmatrix},
\text{CoT step } n_t : \begin{pmatrix} n_t \\ n_{t-1} \\ 0_N \\ 0_N \end{pmatrix},\nonumber
\end{align}
with $n_0=n_\Root$. Every edge-target token thus exposes its edge $(s_i\to t_i)$ in the last two blocks, and every CoT token exposes the transition $n_{t-1}\to n_t$ in the first two.

\paragraph{Second layer.}
The second layer uses two heads with linear attention activation, $\phi^{(2)}(x)=x$, and the weights
\begin{align}
\label{eq:second_layer_weight_kq_ov}
\begin{aligned}
    &\mW_{KQ}^{(2,1)} =
    \begin{pmatrix}
        c_1\mE & 0 & 0 & 0 \\
        0 & 0 & 0 & 0 \\
        c_1\mI & 0 & 0 & 0 \\
        0 & 0 & 0 & 0
    \end{pmatrix},\quad
    \mW_{KQ}^{(2,2)} =
    \begin{pmatrix}
        c_2\mI & 0 & 0 & 0 \\
        c_2\mI & 0 & 0 & 0 \\
        0 & 0 & 0 & 0 \\
        0 & 0 & 0 & 0
    \end{pmatrix},\\
    &\mW_V^{(2,1)} =
    \begin{pmatrix} -\mI \\ 0 \\ 0 \\ \mI \end{pmatrix}^{\!T}\!\!,\quad
    \mW_V^{(2,2)} =
    \begin{pmatrix} -\mI \\ \mI \\ 0 \\ 0 \end{pmatrix}^{\!T}\!\!,\quad
    \mW_O^{(2)} =
    \begin{pmatrix}
        0 & 0 \\
        0 & 0 \\
        \mI & 0 \\
        0 & \mI
    \end{pmatrix},\quad
    \mW_R=\begin{pmatrix}0 & 0 & \mI & \mI\end{pmatrix}^T ,
\end{aligned}
\end{align}
where $\mW_{KQ}^{(l,h)}=(\mW_K^{(l,h)})^T\mW_Q^{(l,h)}$ is the aggregated key-query matrix, all matrices are displayed restricted to the four node blocks and vanish on the constant and positional coordinates, $\mW_R$ has zero columns for the special tokens, and $c_1>c_2>0$ are constants. The leakage parameter of the first layer is chosen as
\begin{align}
\label{eq:delta_margin}
    \delta\leq\frac{c_1-c_2}{4c_1},
\end{align}
so that the first-layer copies, whose scalar factors lie in $[1-2\delta,1]$, cannot reorder the two scores below.
The two heads implement the two DFS actions. The \emph{forward head} matches, for the current node $n_t$, the edge tokens with source $n_t$ and writes their destinations into the readout block, while subtracting every node that already occurs in the CoT trace; an unvisited child thus receives score $\approx c_1$ and a visited child is canceled. The \emph{backtracking head} attends to the CoT transitions entering and leaving $n_t$; completed excursions cancel and the tree-edge arrival leaves score $\approx c_2$ on the stack parent. The backtracking candidate is therefore always available but is dominated whenever an unvisited child exists.

A CoT trace $\tau=(n_0,n_1,\ldots,n_t)$ with $n_0=n_{\Root}$ is a \emph{valid DFS trace} if every step follows the DFS rule of~\cref{sec:preliminaries}, and it is \emph{nonterminated} if DFS has a valid next move from it, i.e., $n_t\neq n_{\Tar}$ and $n_t$ either has an unvisited child or is not the root.

\begin{theorem}[Two-layer transformers can perform DFS with near-linear representation size]
\label{thm:two_layer_dfs}
Suppose the first layer is constructed as in~\cref{lem:first-layer-softmax-copy-relu} with a leakage parameter $\delta$ satisfying~\cref{eq:delta_margin}, and the second layer has weights~\cref{eq:second_layer_weight_kq_ov}. Then the resulting two-layer transformer has $O(1)$ heads per layer, embedding dimension $O(N)$, and $O(\log N)$-bit precision, hence representation size $O(N\log N)$, and for any graph $\gG=(\gV=[N],\gE)$ and any valid nonterminated DFS trace $\tau$ on $\gG$, every token in $\argmax_{v\in\gV} p_\theta(v;\tau_\Prompt,\tau)$ is a valid DFS move.
\end{theorem}

\begin{proof}[Proof sketch]
The representation-size bound is that of~\cref{lem:first-layer-softmax-copy-relu}, since the second-layer weights are constants. Fix a valid nonterminated trace with current node $n_t$. By~\cref{eq:hidden_state_after_first_layer}, the forward head assigns score $c_1\lambda$ with $\lambda\in[1-2\delta,1]$ to each edge-target token with source $n_t$ and score exactly $c_1$ to each CoT token, and its value map adds the destination of the former and subtracts the node of the latter; before the ReLU, an unvisited child of $n_t$ therefore has forward score at least $c_1(1-2\delta)$, while a visited child and a non-child have score at most $0$. The backtracking head attends to the CoT tokens that enter or leave $n_t$; as $n_t$ is on top of the stack, it has left only by forward moves, so each completed excursion to a child yields a departure and a return term that cancel up to the leakage (at most $2\delta c_2$ remains on that child), while the stack parent receives only the tree-edge arrival, between $c_2(1-2\delta)$ and $c_2$. After the ReLU and the readout, every unvisited child has logit at least $c_1(1-2\delta)$, the stack parent has logit in $[c_2(1-2\delta),c_2]$, and every other vertex has logit at most $2\delta c_2$; by~\cref{eq:delta_margin} and $\delta<1/4$ these three bands are ordered, so every maximizer is an unvisited child when one exists and the stack parent otherwise. Full details are in~\cref{app:proof_thm_two_layer}.
\end{proof}

\subsection{Multi-layer look-ahead construction and exponential depth advantage}
Additional layers can expand the bounded reachability information available before each DFS move: after an explicit initialization, each propagation layer doubles the radius represented on the existing edge records.

\begin{definition}[$L$-look-ahead DFS]
\label{def:L-lookahead-dfs}
Let $L\geq1$ be an integer and fix the outgoing-edge order by increasing destination label. Given a valid nonterminated DFS state with current node $u$, target $n_{\Tar}$, and visited set $S$, first consider the unvisited children $v$ with $(u\to v)\in\gE$ and $\operatorname{dist}_{\gG}(v,n_{\Tar})\leq L-1$. If this set is nonempty, move to its first vertex in the fixed order. Otherwise, apply the ordinary DFS rule: move to the first unvisited child if one exists, and otherwise backtrack to the parent in the current DFS stack.
\end{definition}

Here distance is measured in the original graph and a zero-length path is allowed; the rule tests a Boolean bounded-reachability predicate and does not require a qualifying child to minimize distance.

\begin{theorem}[Exponential look-ahead depth advantage]
\label{thm:exponential-depth-advantage}
For every integer $k\geq3$, there exists a $k$-layer transformer with unmasked attention, $O(1)$ heads per layer, embedding dimension $O(N)$, and $O_k(\log(N+1))$-bit fixed-point precision, whose token embeddings are augmented with the positional encoding of~\cref{lem:first-layer-softmax-copy-relu} together with a constant coordinate, a beginning-of-sequence indicator, and a period-three position indicator (\cref{app:multilayer-lookahead}), that implements $L$-look-ahead DFS with $L=1+2^{k-3}$.
Its representation size is $O_k(N\log(N+1))$, where the subscript indicates that the constant may depend on $k$.
\end{theorem}

\begin{proof}[Proof sketch]
The first layer creates persistent source and destination records on the edge-target positions, copies the goal, and prepares the predecessor information used by ordinary DFS. In the second layer, two linear-attention heads aggregate outgoing neighbors, one querying with the source identifier of each edge record and one with its destination identifier, which initializes source and child reachability memories whose supports are the sets reachable within one step from $u$ and $v$ for the record $(u\to v)$. Each of the next $k-3$ layers doubles both radii by querying source identifiers with the current memories; the memories are nonnegative counts and only their supports matter. The final layer combines the goal coordinate of the child memory, the outgoing adjacency of the current node, and the visited-node counts; two fresh ReLU blocks and a linear readout implement the eligibility test, followed by the fixed child order and the ordinary DFS fallback. The complete construction appears in~\cref{app:multilayer-lookahead}.
\end{proof}

Thus depth gives an exponential increase in the radius of the bounded-reachability test. This Boolean test alone does not imply shortest-path selection or an exponential improvement in worst-case search efficiency; in~\cref{app:multilayer-lookahead} we state the efficiency bound separately for a stronger policy that selects a child of minimum residual distance.

%% file: content/learning_dynamics.tex
\section{Training Dynamics}\label{sec:training_dynamics}

In this section we consider the training dynamics of the two-layer model constructed above. Since the predecessor-copy first layer is well studied~\citep{bietti2023birth,sanford2024one,zhu2025reasoning}, we focus on the second layer, whose two heads implement the forward and the backtracking move. The core questions are how each head learns to attend to the right tokens and how the competition between the two heads evolves during training. We therefore fix $\mW_O,\mW_V$ as in the construction and train the key-query matrices
\begin{align}
\mW_{KQ}^{(2,1)} = \begin{pmatrix} \mA_1 & 0 & 0 & 0 \\ 0 & 0 & 0 & 0 \\ \mA_2 & 0 & 0 & 0 \\ 0 & 0 & 0 & 0 \end{pmatrix}, \quad
\mW_{KQ}^{(2,2)} = \begin{pmatrix} \mB_1 & 0 & 0 & 0 \\ \mB_2 & 0 & 0 & 0 \\ 0 & 0 & 0 & 0 \\ 0 & 0 & 0 & 0 \end{pmatrix},
\end{align}
where $\mA_1,\mA_2,\mB_1,\mB_2\in\sR^{N\times N}$ are trainable. When decoding we use the dynamic readout $\mW_R(n_t) = \begin{pmatrix} 0 & 0 & \mD_{n_t} & \mD_{n_t}\end{pmatrix}^T$ with $\mD_{n_t}=\text{diag}(\mI_{n_t-1},0,\mI_{N-n_t})$, so that the logit of the current node $n_t$ is $0$ while $n_t$ remains a softmax candidate with weight $e^0=1$. Throughout, $p_\theta$ is the softmax over the $N$ vertex tokens (the special tokens are never admissible moves), the same readout is used for training and for the generalization loss below, and, since the loss is only piecewise smooth, gradient flow is understood in the Filippov sense ($\dot\theta\in-\alpha\,\partial\gL(\theta)$ with Clarke's generalized gradient; on a switching surface the velocity is a convex combination of the one-sided fields).

\paragraph{Training data.} All training data share a common graph skeleton $\gG_0 = ([N], \gE_0)$ with $N\geq 7$,
\begin{align}
    \gE_0 =\{(1\to i) : 2 \le i \le K+1\} \cup \{(N-1\to N-2),(N-2\to N-3),(N-3 \to 1)\},
\end{align}
where $2\leq K\leq N - 5$, so that the children $2,\ldots,K+1$ of node $1$ are distinct from $N-3,N-2,N-1,N$. The canonical target and root are $(n_\Tar,n_\Root) = (N,N-1)$, and the canonical CoT trajectory is
\begin{align}
    \begin{aligned}
        \tau \;=\; & \underbrace{ N-2 \to N-3 }_{\text{prefix } \tau_{\text{pre}}}\;
        \underbrace{\to 1 \to 2 \to 1\to 3\to 1\to\cdots \to K +1 \to 1 \to N-3 }_{\text{training trace } \tau_0}\\
        & \to N-2 \to N-1 .
    \end{aligned}
\end{align}
We train only on the $2K+2$ steps of $\tau_0$, which traverse all nodes reachable from $N-3$, visiting the children $2,\ldots,K+1$ of node $1$ in turn and backtracking after each; training on the full trace leads to the same mechanism, but $\tau_0$ keeps the gradient expressions clean and isolates the coupling between the two heads that drives the asymptotics.

\paragraph{Training loss.} We use the cross-entropy of the next token: for the canonical trajectory, $\ell_{\tau_\Prompt,\tau_\text{pre},\tau_0} = - \sum_{t=1}^{2K+2} \log p_{\theta}(\tau_0[t];\tau_\Prompt,\tau_\text{pre}, \tau_0[:t-1])$, and the dataset loss is the permutation average
\begin{align}\label{eq:permutation_loss}
    \gL_{\text{train}} = \sE_{\pi \sim \text{Unif}(S_\gV)} \ell_{\pi(\tau_\Prompt),\pi(\tau_\text{pre}),\pi(\tau_0)}.
\end{align}
$\gL_{\text{train}}$ cannot reach $0$, since different exploration branches conflict whenever $n_t$ has several unexplored children, and it only measures performance on the training skeleton.

\paragraph{Generalization loss.} At step $t$ of a CoT trace $\tau$ we score all admissible DFS moves at once, $-\log\sum_{v\in C_t(\tau)} p_{\theta}(v;\tau_\Prompt,\tau[:t-1])$, where $C_t$ is the set of unexplored children if nonempty and otherwise the singleton containing the stack parent of $n_t$; this loss can be driven to $0$. The generalization loss is the supremum over all directed graphs on $N$ vertices and all valid nonterminated DFS traces,
\begin{align}
    \gL_{\text{gen}} = \sup_{\gE,\tau}\ - \sum_{t=1}^{\text{len}(\tau)} \log\Bigl(\sum_{v\in C_t(\tau)} p_{\theta}(v;\tau_\Prompt,\tau[:t-1])\Bigr).
\end{align}

Since the training data is fully symmetric, the permutation-invariant parameters form an invariant set of the gradient flow.
\begin{proposition}\label{prop:symmetric_weight}
    Let $\gF$ denote the set of parameters of the form
    \begin{align}\label{eq:eight_scalar_symmetric}
            \mA_1 = a_1 \mI + b_1 \Eoff, \quad \mA_2 = a_2 \mI + b_2 \Eoff,\quad
 \mB_1 = g_1 \mI + d_1 \Eoff, \quad \mB_2 = g_2 \mI + d_2 \Eoff,
\end{align}
where $\Eoff = \mE - \mI$ has all off-diagonal entries equal to $1$ and zero diagonal. Then $\gF$ is invariant under the gradient flow of~\cref{eq:permutation_loss}: the generalized gradient of $\gL_{\text{train}}$ lies in $\gF$ at every point of $\gF$, and the flow restricted to $\gF$ is a solution of the full flow, unique whenever the loss is differentiable along it. In particular, matrices initialized in $\gF$ remain of the form~\cref{eq:eight_scalar_symmetric} for all $t\geq0$. Moreover, on $\gF$ the averaged loss equals the loss of any single permuted instance, $\gL_{\text{train}}(\theta)=\ell_{\pi(\tau_\Prompt),\pi(\tau_\text{pre}),\pi(\tau_0)}(\theta)$ for every $\pi\in S_\gV$.
\end{proposition}
\begin{proof}
    See~\cref{app:proof_prop_symmetric}.
\end{proof}

We initialize in $\gF$, with the off-diagonal entries of $\mA_2$ and $\mB_1$ strictly below their diagonal entries: for constants $0<c_0'<c_0$,
\begin{align}\label{eq:initialization}
    \mA_1=c_0\mE,\qquad \mA_2=c_0\mI+c_0'\Eoff,\qquad \mB_1=c_0\mI+c_0'\Eoff,\qquad \mB_2=c_0\mE,
\end{align}
that is, $a_1=b_1=a_2=g_1=g_2=d_2=c_0$ and $b_2=d_1=c_0'$. The strict separation $b_2(0)<b_1(0)$, $g_1(0)>d_1(0)$ places the initial point off the two ReLU switching surfaces $b_2=b_1$ and $g_1=d_1$, on the side that the dynamics then preserve (\cref{lemma:scalar_grad}). The all-constant initialization $c_0\mE$ lies on these surfaces, where the vector field is discontinuous, and its trajectory depends on the convention for $\relu'(0)$ and on $(N,K)$: with $\relu'(0)=0$ the backtracking head never activates in exact arithmetic, and when $N$ is large relative to $K$ the trajectory first leaves towards $b_2>b_1$ (\cref{add_exp}).

\begin{theorem}\label{thm:learning_dynamics}
Let $\mA_1,\mA_2,\mB_1,\mB_2$ be initialized as in~\cref{eq:initialization} with $0<c_0'<c_0\leq c_\ast N^{-2}$, where $c_\ast=c_\ast(K)>0$ is the constant of~\cref{lemma:scalar_grad}. Then, under gradient flow on~\cref{eq:permutation_loss} with learning rate $\alpha$, the scalar parameters of~\cref{eq:eight_scalar_symmetric} satisfy, as $t\to\infty$,
\begin{align}
    a_2 , b_1 = 2\log (\alpha t) + O(1),\qquad g_1 - d_1 = \log(\alpha t) + O(1), \qquad
        |g_1 - d_1 - g_2 + d_2 |\leq C,
\end{align}
where $C$ and the $O(1)$ terms depend only on $N$, $K$, $c_0$, and $c_0'$. As a consequence, there exist $t_\ast$ and $C_{N,K}$, depending only on $N,K,c_0,c_0'$, such that for all $t\geq t_\ast/\alpha$,
\begin{align}\label{eq:gen_rate}
    \gL_{\text{gen}} \leq  C_{N,K}\,(\alpha t)^{-\frac{N-1}{N}}.
\end{align}
\end{theorem}

The exponent $(N-1)/N$ in~\cref{eq:gen_rate} is sharp as $t\to\infty$. Forward moves and backtracking moves to a non-root parent have admissible-token loss $O(1/(\alpha t))$, with constants as large as $e^{\,a_2-b_1}$, so these terms dominate $\gL_{\text{gen}}$ over long horizons in practice (\cref{sec:experiment}). When the parent of the current node is the root, however, its logit is $g_1-d_1+d_2$ (\cref{tab:backward_head}), and $d_2=-\frac1N\log(\alpha t)+O(1)$ because $\tau_0$ never backtracks to the root, so $d_2$ is driven only by the conserved combination $g_2+(N-1)d_2$; such a step costs $(\alpha t)^{-(N-1)/N}$.

\begin{proof}[Proof sketch]
    By~\cref{prop:symmetric_weight}, training the matrices is equivalent to training the eight scalars of~\cref{eq:eight_scalar_symmetric}, whose gradients on the canonical data are explicit (\cref{lemma:scalar_grad}). The flow never moves $a_1$, and within an initial phase of length $T_0=\frac{N}{4(K+1)\alpha}$ it drives $b_2$ down to $0$ and $d_2$ to non-positive values; from then on $b_2\equiv0$, since its gradient vanishes on $\{b_2<0\}$ so that $b_2$ cannot become negative, and the trajectory stays in the region $\{b_1>0,\ b_2<b_1,\ a_2\geq b_1,\ g_1>d_1,\ g_2\geq d_2\}$. Writing $x_1=a_2$, $x_2=b_1$, $y_1=g_1-d_1$, $y_2=g_2-d_2$ and $u=x_1-x_2$, $v=y_1-y_2$, the remaining dynamics form a closed four-variable system (\cref{coro:parameter_aggre}) in which the ReLU enters only through the indicators $\ind_{u\geq0}$ and $\ind_{v\geq0}$; the vector field is therefore discontinuous on the switching surfaces $u=0$ and $v=0$, and the trajectory can slide along them (it does so along $v=0$, see~\cref{sec:experiment}). The main steps (\cref{thm:four_parameter_log_convergence}) are: (i) the sliding variable $v_+$ stays bounded, which we show with a Lyapunov function of the form $W=Zs^{p}$ built from the softmax normalizers; (ii) the gaps $u$ and $2y_1-x_1$ are bounded above by barrier arguments; (iii) with the gaps controlled, the leading terms of the equations for $X=x_1$ and $Y=y_1$ take the form $\dot X=a\,e^{Y-X}$, $\dot Y=b\,e^{-Y}-\tilde a\,e^{Y-X}$ with coefficients bounded above and below, and a logarithmic growth lemma for such pairs (\cref{lemma:log_growth}) shows that $e^{2Y-X}$ stays between two positive constants and yields $X=2\log(\alpha t)+O(1)$, $Y=\log(\alpha t)+O(1)$; the bounds on $u$ and $v$ transfer these rates to $x_2$ and $y_2$. Finally, since $\dot g_i=-(N-1)\dot d_i$, the individual parameters are recovered as $g_1=\frac{N-1}{N}y_1+O(1)$, $d_1=-\frac1N y_1+O(1)$, and likewise for $g_2,d_2$ (\cref{eq:gd_recovery}). Inserting these into the node-logit tables of~\cref{app:proof_learning_dynamics} for an arbitrary graph and valid trace bounds every admissible-token loss by $O(e^{-(x_2-y_1)}+e^{-y_1})=O(1/(\alpha t))$, except at a root parent, whose logit $g_1-d_1+d_2=\frac{N-1}{N}\log(\alpha t)+O(1)$ gives~\cref{eq:gen_rate}.
\end{proof}

%% file: content/experiment.tex
\section{Experiment}
\label{sec:experiment}

We validate the predictions of~\cref{sec:training_dynamics}: the effective second-layer parameters grow logarithmically at the predicted rates, the gap variables stay bounded, and the generalization loss decays at the predicted polynomial rates. The protocol, the test set, the invariant-subspace check, and the initialization ablation are detailed in~\cref{add_exp}.

\paragraph{Setup.}
We use $N=16$, $K=8$, the model of~\cref{sec:training_dynamics} with trainable $N\times N$ matrices $\mA_1,\mA_2,\mB_1,\mB_2$ on top of the fixed first layer, the initialization~\cref{eq:initialization} with $c_0=1/N^2$, $c_0'=c_0/2$, and float64 arithmetic. \emph{Full-matrix training} runs gradient descent with step size $\eta=0.05$ on a sampled version of~\cref{eq:permutation_loss} ($B=128$ random vertex permutations per step, $10^7$ steps, $t=5\times10^5$, $\alpha=1$). The \emph{reduced flow} integrates the eight-scalar system of~\cref{lemma:equivalence_parameter}, which is exactly the averaged gradient flow on $\gF$, up to $t=10^9$. Generalization is measured on $64{,}325$ DFS states from $300$ random directed graphs on $16$ vertices (with cycles and self-loops): $39{,}715$ forward steps, $20{,}388$ backtracking steps to a non-root parent, and $4{,}222$ backtracking steps to the root.

\begin{figure}[tbp!]
  \begin{center}
    \includegraphics[width=0.49\columnwidth]{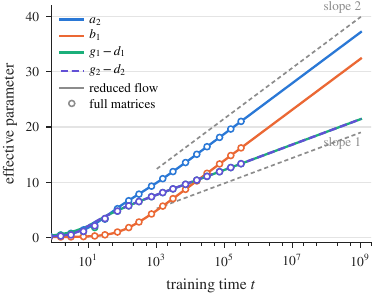}\hfill
    \includegraphics[width=0.49\columnwidth]{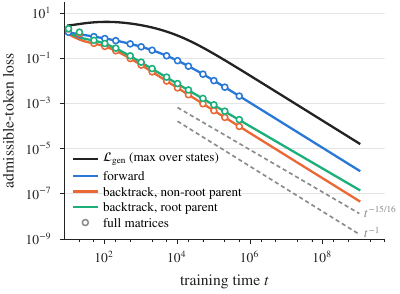}
    \caption{\textbf{Left:} effective parameters against $\log t$ for the reduced flow (lines) and full-matrix training (markers); fitted slopes $1.99$, $2.00$, $1.00$, $1.00$ against the predicted $2,2,1,1$. \textbf{Right:} admissible-token loss on the test set by step type, and its maximum $\gL_{\text{gen}}$. Backtracking to the root decays as $t^{-(N-1)/N}=t^{-0.9375}$; all other steps decay as $t^{-1}$.}
    \label{fig:weight_evolve}
  \end{center}
   \vspace{-18pt}
\end{figure}

\paragraph{Growth rates.}
\Cref{fig:weight_evolve} (left) compares the learned effective parameters with logarithmic fits on $t\in[10^3,10^9]$. For the reduced flow, $a_2$ and $b_1$ grow with slopes $1.992$ and $2.000$, and $g_1-d_1$, $g_2-d_2$ with slope $0.997$; full-matrix training gives $1.966$, $1.996$, $0.990$, $0.990$ on its shorter horizon. The individual parameters follow~\cref{eq:gd_recovery}: $g_1,g_2$ grow with slope $0.935$ (predicted $(N-1)/N=0.9375$) and $d_1,d_2$ with slope $-0.062$ (predicted $-1/N$), while $g_1+(N-1)d_1$ is conserved to $10^{-11}$. As~\cref{thm:four_parameter_log_convergence} predicts, $0\leq a_2-b_1\leq 5.1$ and $|g_1-d_1-g_2+d_2|\leq0.34$ along the whole trajectory, and $b_2,d_2$ become non-positive at $t\approx0.3$. The forward signal thus grows at twice the logarithmic rate of the backtracking signal, which is what lets the model prefer an unexplored child whenever one exists while keeping a lower-priority backtracking candidate. With $B=128$ sampled permutations per step, the full matrices leave the invariant subspace $\gF$ of~\cref{prop:symmetric_weight} during the initial phase, when the dynamics slide along the switching surfaces (relative distance up to $0.9$ for $\mB_1,\mB_2$), but return to within $6\times10^{-4}$ of it by $t=5\times10^5$, where their effective scalars agree with the reduced flow to within $0.03$; larger batches ($B\geq8192$) keep the distance below $0.03$ throughout (\cref{add_exp}).

\paragraph{Generalization.}
\Cref{fig:weight_evolve} (right) shows the admissible-token loss by step type. On $t\in[10^6,10^9]$ the fitted exponents are $-1.000$ for forward steps, $-1.000$ for backtracking to a non-root parent, and $-0.9375$ for backtracking to the root, exactly the $-(N-1)/N$ of~\cref{thm:learning_dynamics}; the mechanism of~\cref{eq:gd_recovery} is visible directly, since the logit of a root parent equals $g_1-d_1+d_2$ to within $0.005$. Repeating the reduced flow for other $N$ (with $K=3$ for $N=8$ and $K=4$ otherwise) gives fitted root-backtracking exponents
\begin{center}
\small
\begin{tabular}{lcccccc}
\toprule
$N$ & 8 & 10 & 12 & 16 & 24 & 32\\
\midrule
fitted & $-0.8750$ & $-0.9000$ & $-0.9167$ & $-0.9376$ & $-0.9585$ & $-0.9690$\\
$-(N-1)/N$ & $-0.8750$ & $-0.9000$ & $-0.9167$ & $-0.9375$ & $-0.9583$ & $-0.9688$\\
\bottomrule
\end{tabular}
\end{center}
all within $3\times10^{-4}$ of the prediction (\cref{fig:exponent_vs_N} in~\cref{add_exp} shows the local exponents). Over the whole observed range, however, the supremum $\gL_{\text{gen}}$ is attained by forward steps at a node that also has an edge back to its stack parent: there the parent collects $g_1-d_1$ from the backtracking head and $a_2-b_1$ from the forward head, so the loss is $\approx e^{-(b_1-(g_1-d_1))}=\Theta(e^{u}/t)$, a $t^{-1}$ rate with the large constant $e^{u}\approx e^{5}$; the bound~\cref{eq:gen_rate} holds, but the $t^{-(N-1)/N}$ term overtakes this one only far beyond any practical horizon. The same states are the last to be classified correctly, at $t\approx2\times10^4$, whereas all other test states are from $t\approx40$ on (\cref{add_exp}).

%% file: content/conclusion.tex
\section{Conclusion}

We gave a theoretical account of how transformers implement, scale, and learn depth-first search on directed graphs. We established a sharp depth threshold for efficient DFS: any one-layer transformer requires $\Omega(N^2)$ representation size (heads $\times$ width $\times$ precision) on dense graphs, while a two-layer construction suffices with $O(N\log N)$ representation size, where the first layer exposes predecessor information and the second combines competing forward and backtracking heads. Beyond this threshold, each additional layer doubles the radius of the bounded-reachability test available before every move, so the look-ahead horizon grows exponentially with depth. Finally, we proved that the two-layer mechanism emerges from gradient flow on a permutation-symmetric curriculum, with logarithmic growth rates and a polynomial generalization rate that our experiments reproduce quantitatively. Together, these results show that efficient DFS requires a genuine two-layer mechanism, that depth translates into exponentially larger look-ahead, and that such reasoning behavior can arise from gradient-based training on discrete chain-of-thought traces.

\paragraph{Limitations and future work.} Our learnability analysis relies on a permutation-symmetric training curriculum and a fixed first layer; extending the gradient-flow analysis to end-to-end training over broader graph distributions is a natural next step. Our look-ahead construction is exact and has a fixed horizon, whereas trained large models appear to perform approximate, heuristic-driven search; closing this gap likely requires combining our discrete-CoT analysis with tools from approximate reasoning and value estimation. Finally, while DFS captures branching, memory, and backtracking, real reasoning tasks involve richer structures such as soft constraints, probabilistic transitions, and hierarchical subgoals. We hope the framework developed here---identifying depth thresholds, quantifying the gain from each additional layer, and tracking gradient-flow dynamics on structured curricula---can serve as a template for analyzing these more complex reasoning algorithms.

%% file: acknowledgement.tex
\section{Acknowledgement}

This work was supported by the U.S. Army Research Laboratory and the U.S. Army Research Office
under Grant W911NF2010219, Office of Naval Research, and NSF. This work used Jetstream2 at
Indiana University through allocation CIS251312 from the Advanced Cyberinfrastructure Coordination Ecosystem: Services \& Support (ACCESS) program, which is supported by National Science
Foundation grants \#2138259, \#2138286, \#2138307, \#2137603, and \#2138296.

%% file: content/related_work.tex
\section{Related Work}

\paragraph{Expressive power of transformers and chain of thought.}
A line of work characterizes what transformers can and cannot compute as a function of depth, precision, and CoT length. \citet{hahn2020theoretical} gives early limitations of self-attention on hierarchical and periodic structure, and \citet{yun2020transformers} shows universal approximation in the seq-to-seq sense. Tighter circuit-complexity bounds place log-precision transformers inside uniform $\mathsf{TC}^0$~\citep{merrill2023parallelism,merrill2023logic,strobl2023average}, and \citet{merrill2025exact} extend these bounds with padding. On the positive side, \citet{liu2023transformers} show that low-depth transformers can simulate finite-state automata via Krohn--Rhodes shortcuts, and \citet{edelman2022inductive} characterize sparse-variable creation under bounded norm. CoT extends these limits: \citet{feng2023towards} prove that constant-size transformers \emph{with} CoT can solve dynamic programming and arithmetic that constant-depth transformers cannot, \citet{li2024chain} show $T$ CoT steps suffice to express any size-$T$ circuit at constant precision, and \citet{merrill2024expressive} give a tight characterization of decoder transformers' expressive power as the CoT length scales. On the learning side, \citet{huang2026transformers} analyze implicit CoT, where intermediate steps are removed during training so that the reasoning is absorbed into hidden states: a $\log_2 k$-layer transformer trained under a curriculum that removes these steps in geometric chunks learns $k$-parity on $n$ bits with $\mathrm{poly}(n)$ samples and $\log_2 k$ training stages, matching the sample efficiency of explicit CoT without generating thinking tokens at inference. Our results refine this landscape along the orthogonal axis of \emph{search-horizon-per-depth}: we show DFS look-ahead grows exponentially in depth at near-linear width.
 
\paragraph{Graph reasoning with transformers.}
The graph-reasoning testbed has emerged as a natural arena for both empirical and theoretical analysis. Empirically, NLGraph~\citep{wang2023nlgraph}, GraphQA~\citep{fatemi2024talk}, and the path-star analyses of~\citet{bachmann2024pitfalls,frydenlund2024mystery} expose systematic failures of LLMs on connectivity, shortest-path, and traversal tasks. \citet{saparov2024transformers} find that small transformers trained on DAG path-finding learn to maintain a per-vertex reachable set whose support grows exponentially with depth. \citet{saparov2023language,saparov2024testing} document analogous greedy-reasoning failures in deductive-proof settings. Theoretically, \citet{sanford2024understanding} build a representational hierarchy of nine graph algorithms and show log-depth is necessary and sufficient for connectivity; \citet{sanford2024transformers} relate depth to MPC rounds; and \citet{yehudai2025depth} establish depth--width tradeoffs for graph primitives via 2-cycle communication-complexity arguments. Our work differs by isolating the \emph{algorithmic} primitive (DFS, with explicit stack and backtracking) and proving a tight one-layer lower bound, a two-layer matching upper bound, and a depth--look-ahead hierarchy.
 
\paragraph{Provable training dynamics.}
A growing body of work analyzes gradient-based training of transformers on structured tasks. \citet{tian2023scan} dissect token composition in one-layer transformers; \citet{nichani2024transformers} prove a two-layer transformer learns Markov-chain causal structure under gradient descent; \citet{chen2024unveiling} and~\citet{olsson2022context} provide training and circuit-level accounts of induction heads; \citet{wang2024transformers} prove transformers (but not MLPs) provably learn sparse token selection; and~\citet{edelman2024evolution} analyze in-context Markov-chain learning. Another line explains what one-layer transformers infer beyond their training facts through the implicit bias of gradient descent: \citet{huang2025generalization} attribute out-of-context reasoning, which underlies both generalization and hallucination after fine-tuning, to a bias toward small nuclear norm of the combined output--value matrix, and \citet{ma2026breaking} prove that the reversal curse occurs on forward-only data but is broken when identity-bridge statements of the form ``$A \to A$'' are added to the training set. Closest to our setting, \citet{yang2025multihead} prove that a one-layer multi-head transformer learns path-finding on \emph{trees} via gradient descent with multi-head specialization. Trees, however, are search-trivial: there is at most one path between any pair of vertices, so neither branching nor backtracking arises. Our paper generalizes this line to arbitrary directed graphs by showing that two-layer transformers provably learn to execute DFS---the canonical algorithm that handles branching and revisitation---under gradient flow and generalize to unseen graphs. \citet{zhu2025emergence} study a complementary regime: they prove that two-layer transformers learn directed-graph reachability with \emph{continuous} CoT through the emergence of superposed reasoning traces. Whereas continuous CoT permits parallel exploration via embedding superposition, our discrete-CoT setting forces the model to produce a sequential search trace, and our analysis tracks how the canonical sequential-search algorithm (DFS) emerges in the parameters. Together, these two results paint a picture of how the same graph-search problem is solved differently in the continuous and discrete CoT regimes.
 
\paragraph{Search and planning in language models.}
Beyond the symbolic setting, several empirical works probe whether LLMs perform search internally. \citet{lehnert2024beyond} train Searchformer to imitate $\mathrm{A}^\ast$ search and improve upon it via search-dynamics bootstrapping; \citet{valmeekam2023planning,kambhampati2024llms} document failure modes of LLMs as planners; and \citet{stechly2023gpt4} analyze iterative-refinement limitations. \citet{saparov2023language} show that LLMs are ``greedy reasoners'' that struggle when proofs branch. Mechanistic-interpretability work~\citep{olsson2022context,nanda2023progress,brinkmann2024mechanistic} identifies internal circuits underlying simple reasoning and traversal behaviors, including a backward-chaining circuit on tree path-finding that motivated~\citet{yang2025multihead}. Our theoretical results provide a complementary explanation for \emph{why} transformers exhibit depth-dependent search horizons in these empirical studies: the $k$-layer DFS construction we exhibit has look-ahead horizon $1+2^{k-3}$, a depth-dependent horizon of the kind that empirical work observes, and our learnability result shows this construction is reachable by gradient-based training on appropriately curated CoT data.

%% file: content/appendix/proof_expressivity.tex
\section{Omitted proof in~\cref{sec:expressivity}}

\subsection{Proof of~\cref{thm:lower_bound}}
\label{app:proof_thm_lower_bound}

\begin{proof}
We reduce from the one-way communication problem $\INDEX$~\citep{kushilevitz1997communication}. Alice receives
\begin{align}
    f=(f_1,\dots,f_M)\in\{0,1\}^M,
\end{align}
and Bob receives an index $i^\star\in[M]$. Alice sends one message to Bob, who must output $f_{i^\star}$. Every deterministic one-way protocol for $\INDEX$ requires $\Omega(M)$ bits.

Assume a one-layer transformer is correct on every valid instance of the unique outgoing-child task whose graph has vertex set $[N]$ and exactly $M$ distinct edges, where $N\geq 3\lceil\sqrt{M}\rceil+1$. We construct a protocol for $\INDEX$.

\paragraph{A row--column encoding with distinct edges.}
Set $m=\lceil\sqrt{M}\rceil$ and assign record $i\in[M]$ to the grid cell
\begin{align}
    r(i)&=1+\left\lfloor\frac{i-1}{m}\right\rfloor,
    &c(i)&=1+((i-1)\bmod m).
\end{align}
The pairs $(r(i),c(i))$ are distinct and belong to $[m]\times[m]$.
Choose $3m+1$ pairwise distinct vertices in $[N]$, named
\[
    q,\quad a_1,\ldots,a_m,\quad
    b_{1,0},b_{1,1},\ldots,b_{m,0},b_{m,1}.
\]
These names denote a fixed public labeling of the vertices. All remaining vertices, if any, are isolated.

For each record $i\in[M]$, define
\begin{align}
    t_i&=b_{c(i),f_i},
    &s_i&=
    \begin{cases}
        q, & i=i^\star,\\
        a_{r(i)}, & i\neq i^\star.
    \end{cases}
\end{align}
Let
\[
    \gE=\{(s_i\to t_i):i\in[M]\},
\]
and list these edges in the public order $1,\ldots,M$.

We verify that $|\gE|=M$. The selected record is the only edge with source $q$, so it cannot coincide with any other edge. For two distinct unselected records $i,j$, if $r(i)\neq r(j)$, their sources are different. If $r(i)=r(j)$, then $c(i)\neq c(j)$, and their targets belong to the disjoint pairs $\{b_{c(i),0},b_{c(i),1}\}$ and $\{b_{c(j),0},b_{c(j),1}\}$. Thus their targets are different. Every edge is therefore distinct. Moreover, the source vertices and target vertices are disjoint, so the construction has no self-loops.

The final query node is the fixed vertex $q$. It has exactly one outgoing edge, whose target is
\begin{align}
    t_{i^\star}=b_{c(i^\star),f_{i^\star}}.
\end{align}
Since Bob knows $i^\star$ and the public vertex labeling, this output determines $f_{i^\star}$.

\paragraph{One-way simulation of the final position.}
The prompt has total length
\begin{align}
    T=3M+2.
\end{align}
Its final token is always $q$, at the same position, so its input embedding $\vx_T$, including its positional encoding, is known to both parties and does not depend on $i^\star$.
Partition the positions into
\begin{align}
    A=\{3,6,9,\dots,3M\},\qquad
    B=[T]\setminus A.
\end{align}
Alice controls the target-token positions $A$: each such token depends only on her input bit and its public column index. Bob knows all positions in $B$, comprising the source tokens, special tokens, and final query. In particular, the input embedding at every position in $A$ is independent of Bob's index.

Fix one attention head $h$, and write
\begin{align}
    A_h=(\mW_K^{(1,h)})^\top\mW_Q^{(1,h)},
    \qquad
    w_j^{(h)}=\phi(\vx_j^\top A_h\vx_T).
\end{align}

First consider unnormalized attention. The final column of head $h$ is
\begin{align}
    \Head_h^{(1)}(\mX)_{:,T}
    =\sum_{j=1}^T w_j^{(h)}\mW_V^{(1,h)}\vx_j.
\end{align}
Alice computes
\begin{align}
    S_A^{(h)}
    =\sum_{j\in A}w_j^{(h)}\mW_V^{(1,h)}\vx_j
    \in\mathbb{R}^d
\end{align}
and sends it to Bob. By the fixed-point convention in the definition of $p$-bit precision, $S_A^{(h)}$ is a coordinate-wise sum of at most $T$ numbers that are each $p$-bit fixed-point numbers, so it is exactly representable with $p+\lceil\log_2T\rceil=O(p)$ bits per coordinate, and the same holds for the partial sum over $B$ below; no additional rounding is introduced by the protocol. The vector lives in the value space of the head, whose dimension is at most $d$. Bob computes
\begin{align}
    S_B^{(h)}
    =\sum_{j\in B}w_j^{(h)}\mW_V^{(1,h)}\vx_j
\end{align}
and reconstructs the head output as $S_A^{(h)}+S_B^{(h)}$. The message has size $O(dp)$ bits for this head.

For normalized attention, the final column is
\begin{align}
    \Head_h^{(1)}(\mX)_{:,T}
    =\frac{S_A^{(h)}+S_B^{(h)}}{Z_A^{(h)}+Z_B^{(h)}},
\end{align}
where
\begin{align}
    Z_A^{(h)}=\sum_{j\in A}w_j^{(h)},
    \qquad
    Z_B^{(h)}=\sum_{j\in B}w_j^{(h)}.
\end{align}
Alice sends $S_A^{(h)}$ and $Z_A^{(h)}$, each exactly representable with $O(p)$ bits per coordinate by the same fixed-point argument, and Bob computes the corresponding quantities for $B$. The additional normalizing scalar leaves the communication at $O(dp)$ bits per head.

Repeating for all $H$ heads, Alice sends $O(Hdp)$ bits. Bob reconstructs every final-column head output, applies the output matrix $\mW_O^{(1)}$ and the residual update
\begin{align}
    \mX^{(2)}_{:,T}
    =\vx_T+\relu\!\left(\Atten^{(1)}(\mX)_{:,T}\right),
\end{align}
and computes the final logits $\mW_R^\top\mX^{(2)}_{:,T}$.
By correctness of the transformer, the predicted token is $b_{c(i^\star),f_{i^\star}}$, from which Bob recovers $f_{i^\star}$.
Thus the transformer induces a one-way protocol for $\INDEX$ using $O(Hdp)$ bits. The $\Omega(M)$ communication lower bound implies
\begin{align}
    Hdp=\Omega(M).
\end{align}
\end{proof}

\subsection{Proof of~\cref{cor:dense_graph_lower_bound}}
\label{app:proof_cor_dense}
\begin{proof}
We have $3\lceil\sqrt{M_N}\rceil+1=3m_N+1\leq N$, so~\cref{thm:lower_bound} applies. Since $M_N=\Theta(N^2)$, it gives $Hdp=\Omega(N^2)$.

For the DFS conclusion, use the same graph family from~\cref{app:proof_thm_lower_bound}, take $q$ as the root, and choose the fixed row vertex $a_1$ as the target. Insert the public target token $a_1$ and the public start-reasoning token $\textsf{<r>}$ immediately before the final root token $q$. The initial DFS state is nonterminated and has visited set $\{q\}$. Its unique child is distinct from $q$, so every valid first DFS move must output that child. The additional task tokens are public and the final query remains fixed, so the same communication reduction applies. The target is unreachable in this family, which is permitted by the DFS task.
\end{proof}

\subsection{Token embeddings of the two-layer construction}
\label{app:token_embeddings}

We use the following token embeddings, excluding a fixed positional-embedding component:
\begin{align}
\text{Edge source } s_i &: \begin{pmatrix} 0_N \\ 0_N \\ 0_N \\ s_i \end{pmatrix}, &
\text{Edge target } t_i &: \begin{pmatrix} 0_N \\ 0_N \\ 0_N \\ t_i \end{pmatrix}, &
\textsf{<s>,<e>,<r>} &: \begin{pmatrix} 0_N \\ 0_N \\ 0_N \\ 0_N \end{pmatrix}, \\
\text{Target } n_{\Tar} &: \begin{pmatrix} 0_N \\ 0_N \\ 0_N \\ n_{\Tar} \end{pmatrix}, &
\text{Root } n_{\Root} &: \begin{pmatrix}n_{\Root} \\ 0_N \\ 0_N \\ 0_N \end{pmatrix}, &
\text{CoT step } n_t &: \begin{pmatrix} n_t \\ 0_N \\ 0_N \\ 0_N \end{pmatrix}. &&
\end{align}
Here $s_i$, $t_i$, $n_{\Tar}$, and $n_t$ denote the corresponding $N$-dimensional one-hot node embeddings. After the first layer of~\cref{lem:first-layer-softmax-copy-relu}, up to the scalar factors in $[1-2\delta,1]$ on the copied one-hot vectors, the hidden states in $\mX^{(2)}$ are those displayed in~\cref{eq:hidden_state_after_first_layer}.

\paragraph{On the embedding dimension.}
The token-embedding dimension is $O(N)$, which is natural for exact DFS because the algorithm must maintain and query an arbitrary visited set $S\subseteq[N]$. There are $2^N$ possible visited sets, so any representation that preserves enough information for exact membership queries must distinguish $2^N$ states. If the visited-set representation has dimension $m$ and uses $p$ bits of precision per coordinate, then it can encode at most $2^{O(mp)}$ distinguishable states, which implies $mp=\Omega(N)$. With the $p=O(\log N)$ precision used in our softmax construction, this yields $m=\Omega(N/\log N)$. Thus, the choice $m=O(N)$ is within a logarithmic factor of the information-theoretic minimum for exact visited-set representation, while keeping the construction simple through one-hot node embeddings.

\subsection{Formal construction of the first layer}

\begin{lemma}[Predecessor-copy layer]
\label{lem:first-layer-softmax-copy-relu}
Let $N$ be the number of graph nodes.  Let $T_{\mathrm{cot}}$ be an upper bound on the length of the CoT trace and let $L_{\max}=3M+T_{\mathrm{cot}}+4$, which bounds the total length of the prompt followed by the CoT trace (one \textsf{<s>} token, $3M$ edge tokens, the three task tokens $n_{\Tar}$, \textsf{<r>}, $n_{\Root}$, and at most $T_{\mathrm{cot}}$ CoT tokens); assume $M\leq N^2$ and $T_{\mathrm{cot}}\leq \poly(N)$.  For each position $i\in[L_{\max}]$, define
\begin{equation}
    p_i=
    \begin{pmatrix}
        \cos(i\alpha)\\
        \sin(i\alpha)
    \end{pmatrix},
    \qquad
    \alpha:=\frac{\pi}{4L_{\max}}.
\end{equation}
Each token representation is
\begin{equation}
    \vx_i=(a_i,b_i,c_i,d_i,\rho_i,p_i)\in\sR^{4N+3},
    \qquad \rho_i=1,
\end{equation}
where $a_i,b_i,c_i,d_i\in\sR^N$.  The scalar coordinate $\rho_i$ is a constant technical coordinate.  Consider the first layer
\begin{align}
      \mX^{(2)} &= \mX^{(1)}+\relu\!\left(\Atten^{(1)}(\mX^{(1)})\right),\\
    \Atten^{(1)}(\mX^{(1)})
      &= \mW_O^{(1)}
    \begin{bmatrix}
        \Head_1^{(1)}(\mX^{(1)})\\
        \Head_2^{(1)}(\mX^{(1)})\\
        \Head_3^{(1)}(\mX^{(1)})
    \end{bmatrix},\\
    \Head_h^{(1)}(\mX^{(1)})
      &=\mW_V^{(1,h)}\mX^{(1)}
      \phi^{(1)}\!\left((\mX^{(1)})^\top(\mW_K^{(1,h)})^\top \mW_Q^{(1,h)}\mX^{(1)}\right),
\end{align}
where $\phi^{(1)}=\softmaxcol$ is column-wise softmax.  Then for any $\delta\in(0,1/8)$, there exist matrices
\[
    \{\mW_Q^{(1,h)},\mW_K^{(1,h)},\mW_V^{(1,h)}\}_{h=1}^3,
    \qquad \mW_O^{(1)},
\]
with embedding dimension $d=4N+3=O(N)$, such that for every position $i\geq 2$:
\begin{enumerate}
    \item if $a_{i-1}=u$ is a one-hot vector, then the second block added by the first layer is supported exactly on $\operatorname{supp}(u)$ and equals $\lambda u$ for some $\lambda\in[1-2\delta,1]$;
    \item if $a_{i-1}=0_N$, then the second block added by the first layer is exactly $0_N$;
    \item if $d_{i-1}=v$ is a one-hot vector, then the third block added by the first layer is supported exactly on $\operatorname{supp}(v)$ and equals $\lambda v$ for some $\lambda\in[1-2\delta,1]$;
    \item if $d_{i-1}=0_N$, then the third block added by the first layer is exactly $0_N$.
\end{enumerate}
Equivalently, for the tokens used in the DFS prompt,
\begin{align}
\mathrm{Edge}\ s_i =
\begin{pmatrix}0_N\\0_N\\0_N\\s_i\\1\\p\end{pmatrix},
&
\mathrm{Edge}\ t_i =
\begin{pmatrix}0_N\\0_N\\0_N\\t_i\\1\\p\end{pmatrix},
\textsf{<s>,<e>,<r>} =
\begin{pmatrix}0_N\\0_N\\0_N\\0_N\\1\\p\end{pmatrix},\\
\mathrm{Target} =
\begin{pmatrix}0_N\\0_N\\0_N\\n_\Tar\\1\\p\end{pmatrix},&
\mathrm{Root} =
\begin{pmatrix}n_\Root\\0_N\\0_N\\0_N\\1\\p\end{pmatrix},
\mathrm{CoT}\ t  =
\begin{pmatrix}n_t\\0_N\\0_N\\0_N\\1\\p\end{pmatrix},
\end{align}
the first layer gives
\begin{align}
\mathrm{Edge}\ s_i
&\longmapsto
\begin{pmatrix}0_N\\0_N\\0_N\\s_i\\1\\p\end{pmatrix},\quad
\mathrm{Root} \longmapsto
\begin{pmatrix}n_\Root\\0_N\\0_N\\0_N\\1\\p\end{pmatrix},\quad
\mathrm{CoT}\ t\longmapsto
\begin{pmatrix}n_t\\(1\pm O(\delta))n_{t-1}\\0_N\\0_N\\1\\p\end{pmatrix},\\
\mathrm{Edge}\ t_i
&\longmapsto
\begin{pmatrix}0_N\\0_N\\(1\pm O(\delta))s_i\\t_i\\1\\p\end{pmatrix},\quad
\mathrm{Target} \longmapsto
\begin{pmatrix}0_N\\0_N\\0_N\\n_\Tar\\1\\p\end{pmatrix},\quad
\textsf{<e>}\ \longmapsto
\begin{pmatrix}0_N\\0_N\\(1\pm O(\delta))t_i\\0_N\\1\\p\end{pmatrix},\\
\textsf{<r>}\ &\longmapsto
\begin{pmatrix}0_N\\0_N\\(1\pm O(\delta))n_\Tar\\0_N\\1\\p\end{pmatrix},
\textsf{<s>}\ \longmapsto
\begin{pmatrix}0_N\\0_N\\0_N\\0_N\\1\\p\end{pmatrix},.
\end{align}
Here every factor $(1\pm O(\delta))$ denotes a scalar in $[1-2\delta,1]$, and the supports of the copied one-hot vectors are exact after the ReLU filter.  The claim for \textsf{<s>} at position $1$, which has no predecessor, is included: that position attends to itself and all of its node blocks are zero.  If $L_{\max}=\poly(N)$ and $\delta\geq 1/\poly(N)$, then the construction uses $O(\log N)$-bit parameter magnitudes and precision.
\end{lemma}

\begin{proof}
We give an explicit construction.

\emph{Step 1: a softmax predecessor selector.}
Let
\begin{equation}
    R_{-1}:=
    \begin{pmatrix}
        \cos\alpha & \sin\alpha\\
        -\sin\alpha & \cos\alpha
    \end{pmatrix}.
\end{equation}
Then $R_{-1}p_i=p_{i-1}$ for all $i\geq2$.  For each head $h\in\{1,2,3\}$, choose $\mW_Q^{(1,h)}$ and $\mW_K^{(1,h)}$ to ignore the four node blocks and the scalar coordinate, and to act on the positional block as
\begin{equation}
    \mW_Q^{(1,h)}\vx_i=\beta R_{-1}p_i,
    \qquad
    \mW_K^{(1,h)}\vx_j=p_j.
\end{equation}
Thus the score from key position $j$ to query position $i$ is
\begin{equation}
    S_{j,i}=\beta\langle p_j,p_{i-1}\rangle
    =\beta\cos((j-i+1)\alpha).
\end{equation}
For fixed $i\geq2$, the unique maximizer is $j=i-1$.  Moreover, for $j\neq i-1$,
\begin{equation}
    S_{i-1,i}-S_{j,i}
    =\beta\left(1-\cos((j-i+1)\alpha)\right)
    \geq \beta(1-\cos\alpha).
\end{equation}
Since $\alpha=\pi/(4L_{\max})$, there is a universal constant $c>0$ such that
\begin{equation}
    1-\cos\alpha\geq \frac{c}{L_{\max}^2}.
\end{equation}
Choose
\begin{equation}
    \beta\geq C L_{\max}^2\log(L_{\max}/\delta)
\end{equation}
with a sufficiently large universal constant $C$.  Then the column-wise softmax attention matrix $P$ satisfies
\begin{equation}
    P_{i-1,i}\geq 1-\delta,
    \qquad
    \sum_{j\neq i-1}P_{j,i}\leq\delta.
    \label{eq:pred-softmax-bound}
\end{equation}

\emph{Step 2: value maps.}
All three heads use the same softmax predecessor matrix $P$.
The first head extracts the first node block:
\begin{equation}
    \mW_V^{(1,1)}\vx_j=a_j.
\end{equation}
The second head extracts the fourth node block:
\begin{equation}
    \mW_V^{(1,2)}\vx_j=d_j.
\end{equation}
The third head uses the constant scalar coordinate to produce a threshold vector:
\begin{equation}
    \mW_V^{(1,3)}\vx_j=\rho_j\ones_N=\ones_N.
\end{equation}
Because every column of $P$ sums to one, the third head outputs exactly $\ones_N$ at every position:
\begin{equation}
    \left(\Head_3^{(1)}(\mX^{(1)})\right)_i=\ones_N.
\end{equation}

\emph{Step 3: output map and ReLU filter.}
Let the output matrix place a thresholded copy of the first head into the second node block and a thresholded copy of the second head into the third node block:
\begin{equation}
    \mW_O^{(1)}
    \begin{bmatrix}u\\v\\r\end{bmatrix}
    =
    \begin{pmatrix}
        0_N\\
        2u-r\\
        2v-r\\
        0_N\\
        0\\
        0_2
    \end{pmatrix},
    \qquad u,v,r\in\sR^N.
\end{equation}
Hence the pre-ReLU attention output at position $i$ is
\begin{equation}
    z_i:=\Atten^{(1)}(\mX^{(1)})_i
    =
    \begin{pmatrix}
        0_N\\
        2\sum_j P_{j,i}a_j-\ones_N\\
        2\sum_j P_{j,i}d_j-\ones_N\\
        0_N\\
        0\\
        0_2
    \end{pmatrix}.
\end{equation}
Now suppose $a_{i-1}=e_k$ is one-hot.  By Eq.~\eqref{eq:pred-softmax-bound}, the $k$-th coordinate of $\sum_jP_{j,i}a_j$ lies in $[1-\delta,1]$: it is at least $P_{i-1,i}\geq1-\delta$, and it is at most one because the attention weights sum to one and every $a_j$ has entries in $\{0,1\}$. Every coordinate outside $k$ is at most $\delta$, because the total non-predecessor mass is at most $\delta$.  Hence
\begin{equation}
    \relu\!\left(2\sum_jP_{j,i}a_j-\ones_N\right)
    =\lambda e_k,\qquad \lambda\in[1-2\delta,1],
\end{equation}
with exact support $\{k\}$, because $2\delta-1<0$ kills every other coordinate.  If $a_{i-1}=0_N$, then every coordinate of $\sum_jP_{j,i}a_j$ is at most $\delta<1/2$, and therefore
\begin{equation}
    \relu\!\left(2\sum_jP_{j,i}a_j-\ones_N\right)=0_N.
\end{equation}
The same argument applied to the fourth node block proves the corresponding claims for $d_{i-1}$ and the third block.

\emph{Step 4: applying to the prompt.}
For an edge target token $\mathrm{Edge}\ t_i$, the previous token is $\mathrm{Edge}\ s_i$, whose fourth node block is $s_i$ and whose first node block is zero.  Thus the ReLU-filtered residual addition places $(1\pm O(\delta))s_i$ into the third block and places zero into the second block.  For the source token, the previous token is $\textsf{<e>}$, whose four node blocks are all zero, so the ReLU-filtered residual addition is zero.  For a CoT token $n_t$, the previous token has first node block $n_{t-1}$ and fourth node block zero, so the ReLU-filtered residual addition places $(1\pm O(\delta))n_{t-1}$ into the second block and places zero into the third block. Finally, at position $i=1$ the scores are $S_{j,1}=\beta\cos(j\alpha)$ for $j\geq1$, whose unique maximizer is $j=1$ with gap $S_{1,1}-S_{j,1}\geq\beta(\cos\alpha-\cos2\alpha)\geq\beta(1-\cos\alpha)$; hence the same concentration bound holds with $i-1$ replaced by $1$, and since $a_1=d_1=0_N$ for \textsf{<s>}, both ReLU-filtered additions at position $1$ are $0_N$.

Finally, $d=4N+3=O(N)$.  The nonzero parameters are $0,1,2$, the rotation entries $\cos\alpha,\sin\alpha$, and the scale $\beta$.  If $L_{\max}=\poly(N)$ and $\delta\geq1/\poly(N)$, then $\beta=\poly(N)$ and all parameters can be represented to inverse-polynomial accuracy with $O(\log N)$ bits.  The softmax concentration bound is stable under inverse-polynomial perturbations of the scores, so $O(\log N)$ precision suffices.
\end{proof}

\subsection{Proof of~\cref{thm:two_layer_dfs}}
\label{app:proof_thm_two_layer}
\begin{proof}
The representation-size bound follows from~\cref{lem:first-layer-softmax-copy-relu}: $O(1)$ heads, embedding dimension $4N+3$, and $O(\log N)$ precision, the second-layer weights being constants. The lemma applies with $L_{\max}=3N^2+2N+2=\poly(N)$ to every graph on $[N]$ and every valid DFS trace on it. Indeed $M\leq N^2$, and a valid DFS trace $(n_0,\ldots,n_t)$ has $t\leq 2N-2$: each forward move enters a previously unvisited vertex, so there are at most $N-1$ forward moves, and each backtracking step pops a vertex pushed by an earlier forward move, so there are at most as many backtracking steps as forward moves.

Fix a valid nonterminated DFS trace and let $n_t$ be the current node. By~\cref{lem:first-layer-softmax-copy-relu}, every edge-target token exposes its source in the third block and its destination in the fourth block, and every CoT token exposes the current node in the first block and the previous node in the second block, where each copied one-hot vector carries a scalar factor in $[1-2\delta,1]$ and all other coordinates of the copied block are exactly zero. The forward head assigns score $c_1\lambda$ with $\lambda\in[1-2\delta,1]$ to each edge-target token whose source is $n_t$, and score exactly $c_1$ to each CoT token, including the root; its value map adds the destination of an edge-target token and subtracts the node of a CoT token. Hence, before the ReLU, an unvisited child of $n_t$ has forward score at least $c_1(1-2\delta)$, a visited child has forward score at most $c_1(1-1)=0$, and a non-child has forward score at most $0$. The backtracking head attends to the CoT tokens that enter or leave $n_t$. Since $n_t$ is the top of the current DFS stack, it has never been popped. Hence every step that left $n_t$ was a forward move to a previously unvisited child, each such child was entered exactly once, and its excursion has been completed by the backtracking step that returned to $n_t$. In particular, no step went directly from $n_t$ to its stack parent, which was visited before $n_t$ and is therefore neither an unvisited child of $n_t$ nor a backtracking target from $n_t$ while $n_t$ is on the stack. Consequently every completed excursion from $n_t$ to a child and back contributes exactly one departure term and one return term, whose node coordinates cancel up to the leakage, leaving at most $2\delta c_2$ on that child, while the stack parent receives only the unique tree-edge arrival into $n_t$, which contributes at least $c_2(1-2\delta)$ and at most $c_2$ on the stack parent and a negative amount on $n_t$; at the root there is no tree-edge arrival. After the ReLU and the readout, every unvisited child has logit at least $c_1(1-2\delta)$, the stack parent has logit in $[c_2(1-2\delta),c_2]$, and every other vertex has logit at most $2\delta c_2$. By~\cref{eq:delta_margin}, $c_1(1-2\delta)>c_2$, and $\delta<1/4$ gives $c_2(1-2\delta)>2\delta c_2$. Therefore, if $n_t$ has an unvisited child, every maximizer is an unvisited child of $n_t$; otherwise $n_t$ is not the root, because the trace is nonterminated, and the unique maximizer is the stack parent. In both cases every maximizer of the next-token distribution is a valid DFS move.
\end{proof}

\subsection{Proof of the look-ahead construction and a conditional efficiency bound}
\label{app:multilayer-lookahead}

\begin{lemma}[$k$-layer transformer look-ahead]
\label{lem:k-layer-lookahead}
For every integer $k\geq3$, there exists a $k$-layer transformer with unmasked attention that implements the bounded-reachability policy in~\cref{def:L-lookahead-dfs} with horizon $L=1+2^{k-3}$. Its token embeddings consist of node one-hot blocks together with the positional encoding $p_i$ of~\cref{lem:first-layer-softmax-copy-relu}, a constant coordinate, a beginning-of-sequence indicator, and a period-three position indicator, as specified in the proof. The construction has $O(1)$ heads per layer, width $O(N)$, and fixed-point precision $O_k(\log(N+1))$.
\end{lemma}

\begin{proof}
We specify the hidden state after each layer and the attention heads producing its next update. Write $g=n_{\Tar}$, and let $e_w\in\mathbb{R}^N$ be the one-hot vector of vertex $w$. An \emph{edge record} is the target-token position in a triple $(u,v,\textsf{<e>})$. All attention sums below range over the entire current input prefix, of length $T$.

\paragraph{State notation and permitted updates.}
Let $h_i^{[\ell]}=\mX^{(\ell+1)}_{:,i}$ denote the state at position $i$ after $\ell$ layers, so $h_i^{[0]}$ is the input embedding. The architecture updates it by
\begin{equation}
\label{eq:lookahead-state-update}
    h_i^{[\ell+1]}
    =h_i^{[\ell]}+
      \relu\!\left(W_O^{[\ell+1]}
      [H_{1,i};\ldots;H_{H,i}]\right).
\end{equation}
For a linear-attention head, a row $(q_i,k_s,v_s)$ in the tables below means
\[
    H_i=\sum_{s=1}^T\langle q_i,k_s\rangle v_s.
\]
Each displayed query, key, and value is a linear function of the incoming state; smaller vectors are padded by zeros to a common head dimension. The output projection routes and linearly combines these head outputs in the specified coordinate blocks; the only post-attention nonlinearity is the single ReLU in~\cref{eq:lookahead-state-update}.

For exact logical copies, use a pair of physical blocks
\[
    \widehat Y=(Y^+,Y^-),\qquad Y=Y^+-Y^-.
\]
A query, key, value, or readout map can decode $Y$ by a linear subtraction. For a nonnegative increment $\Delta$, write
\[
    \widehat Y\oplus\Delta=(Y^++\Delta,Y^-).
\]
This is a residual update to the plus block, leaving the minus block unchanged. Also write
\[
    \Pi(z)=\bigl(\relu(z),\relu(z-\mathbf{1})\bigr).
\]
When written into two initially zero blocks, $\Pi(z)$ is realized by two rows of the same output projection followed by the layer's single ReLU. Neither notation introduces an extra layer.

Retain the input block
\[
    \eta_i=(a_i,d_i,\gamma_i,\rho_i,\chi_i,\zeta_i,p_i).
\]
Here $a_i$ is the node one-hot at root and CoT positions and zero elsewhere; $d_i$ is the node one-hot at edge endpoints and the task-target position and zero elsewhere; and $\gamma_i=e_g$ only at the task-target position. Furthermore, $\rho_i=1$, $\zeta_i$ is the BOS indicator, $\chi_i=\mathbf{1}\{i\equiv0\pmod3\}$, and $p_i$ is the two-dimensional positional encoding from~\cref{lem:first-layer-softmax-copy-relu}. The features use $O(N)$ coordinates and remain unchanged throughout. An edge record is characterized by $\chi_i=1$ and $\mathbf{1}^{\top}d_i=1$. Indeed, the edge-target token of the $i$-th edge sits at position $3i$; the task target sits at position $3M+2$ and has $\chi=0$; the token \textsf{<r>} sits at position $3M+3$ and has $\chi=1$ but $d=0$; and every root or CoT position, as well as every source and \textsf{<e>} position, has $d=0$ or $\chi=0$.

The complete physical state has the fixed coordinate layout
\begin{equation}
\label{eq:lookahead-state-layout}
    h_i^{[\ell]}=
    \bigl(\eta_i,\widehat b_i,\widehat U_i,\widehat V_i,
    \widehat G_i,\widehat S_i,\widehat C_i,
    D_i^{\rm f},D_i^{\rm b},Z_i^+,Z_i^-\bigr),
\end{equation}
where the blocks after $\eta_i$ are initially zero. The paired blocks store the predecessor, source identifier, destination identifier, goal, source reachability memory, and child reachability memory, respectively. The last four blocks hold the forward DFS score, backtracking score, and two final eligibility outputs. Each unpaired working block has $N$ coordinates; each paired block has $2N$ coordinates.

\paragraph{Layer 1: endpoint and goal copies.}
After the first layer, the exact decoded values are
\begin{equation}
\label{eq:lookahead-endpoint-records}
    (a_i,b_i,U_i,V_i,G_i)=
    \begin{cases}
        (0,0,e_u,e_v,e_g),&i\text{ records }(u\to v),\\
        (e_{n_{\rm Root}},0,0,0,e_g),&i\text{ is the root position},\\
        (e_{n_t},e_{n_{t-1}},0,0,e_g),&i\text{ is CoT step }t,\\
        (0,0,0,0,e_g),&\text{otherwise},
    \end{cases}
\end{equation}
where $n_0$ is the root. Define the immutable block
\[
    I_i=(\eta_i,\widehat b_i,\widehat U_i,\widehat V_i,\widehat G_i).
\]
Duplicate the source and destination pairs into the two mutable memories. Thus the full state after layer 1 is
\begin{equation}
\label{eq:lookahead-state-layer1}
    \boxed{\displaystyle
    h_i^{[1]}=(I_i,\widehat U_i,\widehat V_i,0,0,0,0).}
\end{equation}
In this display the second and third entries occupy the $\widehat S$ and $\widehat C$ coordinates of~\cref{eq:lookahead-state-layout}; they are separate copies of the pairs retained in $I_i$.

Here is an explicit realization of this state. The first layer uses the following three \emph{softmax} heads:
\[
\begin{array}{c|c|c|c}
    \text{head}&q_i&k_s&v_s\\ \hline
    \text{predecessor}&\beta R_{-1}p_i&p_s&(d_s,a_s,\rho_s)\\
    \text{self}&\beta p_i&p_s&(d_s,a_s,\chi_s,\rho_s)\\
    \text{goal}&\beta_g\rho_i&\mathbf{1}^{\top}\gamma_s&(\gamma_s,\rho_s)
\end{array}
\]
The predecessor selector is that of~\cref{lem:first-layer-softmax-copy-relu}; the self selector uses the identity rotation. Choose the parameters and rounding accuracy so that their outputs satisfy
\[
    z_i^{\rm pre}\approx d_{i-1},\quad
    h_i^{\rm pre}\approx a_{i-1},\quad
    z_i^{\rm self}\approx d_i,\quad
    h_i^{\rm self}\approx a_i,\quad
    c_i\approx\chi_i,\quad z_i^{\rm goal}\approx e_g,
\]
with effective error at most $\delta<1/64$ on every used coordinate and scalar sum. The predecessor approximations are needed only for $i>1$. Each head's constant-value output is one. Set $n_i=\mathbf{1}^{\top}z_i^{\rm self}$ and $t_i=\mathbf{1}^{\top}h_i^{\rm self}$. The output projection forms
\begin{align*}
    \omega_i(z)&=4z+4(c_i+n_i)\mathbf{1}-9\mathbf{1},\\
    r_i&=4h_i^{\rm pre}+4t_i\mathbf{1}-5\mathbf{1},\\
    g_i^{\rm copy}&=4z_i^{\rm goal}-\mathbf{1},
\end{align*}
and writes
\[
    \widehat U_i=\Pi(\omega_i(z_i^{\rm pre})),\qquad
    \widehat V_i=\Pi(\omega_i(z_i^{\rm self})),\qquad
    \widehat b_i=\Pi(r_i),\qquad
    \widehat G_i=\Pi(g_i^{\rm copy}).
\]
The same outputs for $\widehat U_i,\widehat V_i$ are routed to the two mutable memories. At a desired one-hot coordinate the argument of $\Pi$ is close to $3$; at any other coordinate it is at most $-1+12\delta<0$. The difference of the two physical outputs is therefore exactly one or zero, respectively. For endpoint copies, the three tests are the endpoint coordinate, the record-position feature, and the presence of a node in $d_i$. This excludes source tokens, the task target, special tokens, and CoT tokens from the edge memories. All offsets are linear maps of the constant-value output, and all pairs are produced in parallel by one ReLU. The finite-precision justification is given below.

\paragraph{Layer 2: one-step memories and the DFS scores.}
All subsequent attention is unnormalized linear attention. Use four heads, with $\tau_s=\mathbf{1}^{\top}a_s$:
\begin{equation}
\label{eq:lookahead-layer2-heads}
\begin{array}{c|c|c|c|c}
    \text{head}&q_i&k_s&v_s&\text{output block}\\ \hline
    \text{source neighbors}&U_i&U_s&V_s&S_i^+\\
    \text{child neighbors}&V_i&U_s&V_s&C_i^+\\
    \text{DFS forward}&a_i&U_s+\tau_s\mathbf{1}&2(V_s-a_s)&D_i^{\rm f}\\
    \text{DFS backtrack}&a_i&a_s+b_s&b_s-a_s&D_i^{\rm b}
\end{array}
\end{equation}
Every entry is a linear map of $h^{[1]}$. Route each head to the indicated block and set all other rows of the output projection to zero. Consequently,
\begin{align}
    \widehat S_i^{(0)}
        &=\widehat U_i\oplus
          \sum_s\langle U_i,U_s\rangle V_s,
        &\widehat C_i^{(0)}
        &=\widehat V_i\oplus
          \sum_s\langle V_i,U_s\rangle V_s,
        \label{eq:lookahead-layer2-memory-update}\\
    D_i^{\rm f}
        &=\relu\!\left(
          2\sum_s\langle a_i,U_s+\tau_s\mathbf{1}\rangle(V_s-a_s)
          \right),\label{eq:lookahead-forward-block}\\
    D_i^{\rm b}
        &=\relu\!\left(
          \sum_s\langle a_i,a_s+b_s\rangle(b_s-a_s)
          \right).\label{eq:lookahead-backtrack-block}
\end{align}
The memory increments are nonnegative, so the ReLU leaves them unchanged. The score blocks start at zero and receive the ReLU of their respective head outputs. The full hidden state is now
\begin{equation}
\label{eq:lookahead-state-layer2}
    \boxed{\displaystyle
    h_i^{[2]}=
    (I_i,\widehat S_i^{(0)},\widehat C_i^{(0)},
     D_i^{\rm f},D_i^{\rm b},0,0).}
\end{equation}
For the record $e=(u,v)$, its decoded memories are explicitly
\begin{align}
\label{eq:lookahead-initialization-source}
    S_e^{(0)}&=e_u+\sum_{(u\to w)\in\gE}e_w,\\
\label{eq:lookahead-initialization-child}
    C_e^{(0)}&=e_v+\sum_{(v\to w)\in\gE}e_w.
\end{align}
The endpoint terms come from the residual connection, and the neighbor sums come from the first two attention heads. At nonrecord positions both memories are zero.

At the final position $i=T$, let $x$ be the current node. The last two heads are the ordinary DFS heads in~\cref{eq:second_layer_weight_kq_ov} with $c_1=2,c_2=1$, now written as explicit projections into fresh blocks. The forward head subtracts two times the occurrence count of each visited node from two times the adjacency vector of $x$. In the backtracking head, completed excursions cancel, leaving the stack-parent coordinate positive. Therefore, on a valid nonterminated state,
\begin{equation}
\label{eq:lookahead-dfs-block-meaning}
    D_{T,v}^{\rm f}=2\mathbf{1}\{v\text{ is an unvisited child of }x\},
    \qquad
    D_{T,v}^{\rm b}=\mathbf{1}\{v\text{ is the stack parent of }x\}.
\end{equation}
The second indicator is zero at the root. Since attention is unmasked, this interpretation is asserted at the final query position; the scores at earlier positions also use the whole current prefix. All later layers preserve both score blocks.

\paragraph{Layers $3,\ldots,k-1$: doubling the stored radius.}
Put $J=k-3$. For $j=0,\ldots,J-1$, layer $j+3$ takes the state in~\cref{eq:lookahead-state-layer2} with memories indexed by $j$ and uses two heads:
\[
\begin{array}{c|c|c|c|c}
    \text{head}&q_i&k_s&v_s&\text{output block}\\ \hline
    \text{source propagation}&S_i^{(j)}&U_s&S_s^{(j)}&S_i^+\\
    \text{child propagation}&C_i^{(j)}&U_s&S_s^{(j)}&C_i^+
\end{array}
\]
Thus the physical and decoded updates are, respectively,
\begin{align}
    \widehat S_i^{(j+1)}
        &=\widehat S_i^{(j)}\oplus
          \sum_s\langle S_i^{(j)},U_s\rangle S_s^{(j)},\\
    \widehat C_i^{(j+1)}
        &=\widehat C_i^{(j)}\oplus
          \sum_s\langle C_i^{(j)},U_s\rangle S_s^{(j)},\\[0.3em]
\label{eq:lookahead-doubling}
    S_e^{(j+1)}
        &=S_e^{(j)}+
          \sum_{f=(a,b)\in\gE}S_e^{(j)}[a]S_f^{(j)},\\
    C_e^{(j+1)}
        &=C_e^{(j)}+
          \sum_{f=(a,b)\in\gE}C_e^{(j)}[a]S_f^{(j)}.
\end{align}
Every update is a nonnegative head output added to a plus block. No block is overwritten and no Boolean threshold is applied. After $j$ propagation layers, including $j=0$ for the state after layer 2, the full state is
\begin{equation}
\label{eq:lookahead-state-propagation}
    \boxed{\displaystyle
    h_i^{[j+2]}=
    (I_i,\widehat S_i^{(j)},\widehat C_i^{(j)},
     D_i^{\rm f},D_i^{\rm b},0,0),\qquad 0\leq j\leq J.}
\end{equation}
To specify its meaning, define
\[
    \mathcal R_r(w)=\{z:\operatorname{dist}_{\gG}(w,z)\leq r\}.
\]
The decoded memories have nonnegative integer coordinates and satisfy
\begin{equation}
\label{eq:lookahead-support-invariant}
    \operatorname{supp}_+(S_e^{(j)})=\mathcal R_{2^j}(u),
    \qquad
    \operatorname{supp}_+(C_e^{(j)})=\mathcal R_{2^j}(v),
    \qquad e=(u,v).
\end{equation}
The base case follows directly from~\cref{eq:lookahead-initialization-source,eq:lookahead-initialization-child}. Self-loops may give an entry of two, which does not affect its positive support. For the induction step, write $r=2^j$. Every summand in~\cref{eq:lookahead-doubling} concatenates two paths of length at most $r$. Conversely, paths of length at most $r$ remain through the residual. Split a path of length $r<\ell\leq2r$ after $r$ edges, at vertex $a$. Its remaining suffix is nonempty, so $a$ has an outgoing edge and a corresponding source record $f$. Both the coefficient at $a$ and the appropriate coordinate of $S_f^{(j)}$ are positive, representing the whole path. The sum is over edge records, and multiple records with the same source only add positive contributions. This proves the invariant for both memories, including the zero-length path at a sink.

In particular, the input to the final layer is $h_i^{[k-1]}=h_i^{[J+2]}$, with each edge's child memory already representing radius $r=2^J$.

\paragraph{Layer $k$: eligibility blocks and readout.}
Four parallel linear-attention heads read $h^{[k-1]}$:
\begin{equation}
\label{eq:lookahead-final-heads}
\begin{array}{c|c|c|c|c}
    \text{head}&q_i&k_s&v_s&\text{head output}\\ \hline
    \text{goal reachability}&G_i&C_s^{(J)}&V_s&F_i\\
    \text{outgoing adjacency}&a_i&U_s&V_s&A_i\\
    \text{visited counts}&\rho_i&\rho_s&a_s&m\\
    \text{constant}&\rho_i&\zeta_s&\rho_s\mathbf{1}&\mathbf{1}
\end{array}
\end{equation}
At $i=T$, these outputs are
\[
    F=\sum_{e=(u,v)\in\gE}C_e^{(J)}[g]e_v,\qquad
    A=\sum_{e=(u,v)\in\gE}\langle e_x,e_u\rangle e_v,
    \qquad m=\sum_s a_s.
\]
The simple-graph assumption gives $A_v\in\{0,1\}$, with $A_v=1$ exactly for an outgoing child. Also $m_v=0$ exactly for an unvisited vertex. For any outgoing child $v$, the support invariant gives
\[
    F_v>0\quad\Longleftrightarrow\quad
    \operatorname{dist}_{\gG}(v,g)\leq r.
\]
This includes $v=g$, even if $g$ is a sink.

Choose an integer $B\geq1$ bounding every coordinate of $F$, as specified below. The output projection writes only to the two fresh eligibility blocks, with preactivations $z_i$ and $z_i-\mathbf{1}$, where
\[
    z_i=F_i+B A_i-2B m-B\mathbf{1}.
\]
Consequently, the complete state after the final layer is
\begin{equation}
\label{eq:lookahead-state-final}
    \boxed{\displaystyle
    h_i^{[k]}=
    (I_i,\widehat S_i^{(J)},\widehat C_i^{(J)},
     D_i^{\rm f},D_i^{\rm b},\relu(z_i),\relu(z_i-\mathbf{1})).}
\end{equation}
At the query position define, by a linear readout,
\begin{equation}
\label{eq:lookahead-eligibility}
    E_v=Z_{T,v}^+-Z_{T,v}^-
       =\relu(z_{T,v})-\relu(z_{T,v}-1).
\end{equation}
All entries defining $z_{T,v}$ are integers. An unvisited outgoing child with $F_v>0$ has $z_{T,v}=F_v\geq1$, hence $E_v=1$. A nonchild has $z_{T,v}\leq F_v-B\leq0$; a visited child has $z_{T,v}\leq F_v+B-2B-B\leq F_v-2B\leq -B<0$; and a child with $F_v=0$ and $m_v=0$ has $z_{T,v}=0$. Thus $E_v$ is exactly the eligibility indicator.

Let $Q_N=2^{\lceil\log_2N\rceil}$. The final linear readout assigns vertex $v$ the logit
\begin{equation}
\label{eq:lookahead-final-logits}
    \ell_v=4(Z_{T,v}^+-Z_{T,v}^-)
           +D_{T,v}^{\rm f}+D_{T,v}^{\rm b}
           +\frac{N-v}{4Q_N}\rho_T.
\end{equation}
Assign each special token logit $-\rho_T$. The dyadic last term for vertex tokens lies in $[0,1/4)$ and breaks ties by increasing label. Eligible children are preferred; if none exists, the forward DFS scores select an unvisited child, or the backtracking score selects the stack parent. This implements~\cref{def:L-lookahead-dfs} with $L=r+1=1+2^{k-3}$.

\paragraph{Precision and resources.}
Use a common fixed-point grid with step $2^{-b}$, where $b=O(\log(N+1))$, and sufficient range to avoid overflow. First-layer head outputs are rounded to this grid before the integer-coefficient output projection. The two preactivations in each pair have the same nonconstant terms and biases differing by exactly one. The gate margins keep both active or both inactive as specified, so their physical outputs decode to an exact one or zero. The positional-selector and goal-selector parameters need only $O(\log(N+1))$ bits because the input length is polynomial in $N$. After this layer, decoded queries, keys, values, and memory increments are integers. An integer increment to the plus block preserves the exact decoded value.

After layer 2 the decoded memory entries are at most $P_0=2$. Since $M\leq N^2$, a uniform bound for the propagation is
\[
    P_{j+1}=(N^2+1)P_j^2,\qquad
    P_j=2^{2^j}(N^2+1)^{2^j-1}.
\]
Indeed, each coordinate increases by at most $M P_j^2$ in addition to its old value $P_j$. Every destination has at most $N$ incoming edges, so take $B=N P_J$. Valid DFS traces have $O(N)$ node occurrences, and all remaining intermediate values have magnitude at most a polynomial factor in $N$ times $B$. Hence
\[
    p=O\!\left(2^{k-3}\log(N+1)\right)
      =O_k(\log(N+1))
\]
bits suffice, including fractional precision and the physical pairs' bounded initial offsets.

The fixed layout~\cref{eq:lookahead-state-layout} uses $O(N)$ coordinates; the same two mutable memories are reused at every propagation layer. There are three heads in layer 1, four in layer 2, two in each propagation layer, and four in the final layer. Zero heads may be added if a common head count is required. The layer count is
\[
    \underbrace{1}_{\text{copies}}
    +\underbrace{1}_{\text{neighbor aggregation and DFS}}
    +\underbrace{(k-3)}_{\text{doublings}}
    +\underbrace{1}_{\text{eligibility}}
    =k.
\]
For $k=3$ the propagation range is empty and the final layer reads the one-step child memories in $h^{[2]}$.
\end{proof}

\begin{proof}[Proof of~\cref{thm:exponential-depth-advantage}]
Apply~\cref{lem:k-layer-lookahead}. Its constant head count, $O(N)$ width, and $O_k(\log(N+1))$ precision give representation size $O_k(N\log(N+1))$ under the definition of~\cref{sec:expressivity}.
\end{proof}

\paragraph{Distinguishing bounded reachability from distance minimization.}
The preceding construction tests whether each candidate child reaches the goal within one specified radius. It does not compare the distances of qualifying children. For example, with edges $u\to v$, $u\to g$, and $v\to g$, horizon $L=2$, and $v$ ordered before $g$, both children qualify and the policy takes two moves although the target is one edge from $u$. Therefore the depth--horizon theorem alone does not imply the search-efficiency formula below. We state that formula for a stronger policy with an explicit distance-minimization requirement.

\begin{definition}[Distance-minimizing $L$-look-ahead DFS]
\label{def:distance-minimizing-lookahead}
At a nonterminated DFS state with current node $u$, goal $g$, and visited set $S$, let $d_S(v,g)$ be the length of a shortest path from an unvisited child $v$ to $g$ all of whose vertices lie outside $S$, or $\infty$ if no such path exists. If some unvisited child satisfies $d_S(v,g)\leq L-1$, select a child minimizing $d_S(v,g)$, breaking ties by the fixed child order. Otherwise apply the ordinary DFS rule.
\end{definition}

For $n\geq2$, define the worst-case efficiency of a fixed policy by
\[
    \operatorname{eff}
    =\min_{\gG,n_s,g}
      \frac{\operatorname{dist}_{\gG}(n_s,g)}
           {\text{number of CoT moves before reaching }g},
\]
where the minimum is over instances with exactly $n$ vertices reachable from $n_s$ and a distinct reachable target $g$.

\begin{lemma}[Efficiency with distance-minimizing look-ahead]
\label{lem:lookahead-efficiency}
For every integer $L\geq1$, the distance-minimizing policy in~\cref{def:distance-minimizing-lookahead} has worst-case efficiency
\begin{equation}
    \operatorname{eff}=
    \begin{cases}
        \displaystyle\frac{L+1}{2n-L-3},&n\geq L+2,\\[1em]
        1,&n<L+2.
    \end{cases}
\end{equation}
This statement is a property of that stronger policy and does not assert that the Boolean construction implements it.
\end{lemma}

\begin{proof}
Let $m=\operatorname{dist}_{\gG}(n_s,g)$. If $m\leq L$, the suffix after the root of a shortest source-to-goal path lies outside the initial visited set. The policy selects a child on a shortest residual path. The suffix of that path remains available after the move, and its minimum remaining length decreases by one. Induction shows that the target is reached in exactly $m$ moves, giving efficiency one.

Suppose instead that $m\geq L+1$. Let $D$ be the length of the final DFS-stack path to the target, and let $T$ be the total number of moves. Every discovered tree edge outside that final path is traversed at most twice, while each of the $D$ final-path edges is traversed once before termination. Since $D\geq m$,
\[
    T\leq2(n-1)-D\leq2n-m-2.
\]
The function $m/(2n-m-2)$ is increasing on the relevant range, so
\[
    \frac{m}{T}\geq\frac{m}{2n-m-2}
       \geq\frac{L+1}{2n-L-3}.
\]
If $n<L+2$, every reachable target has distance at most $n-1\leq L$, so the first case applies to every instance.

For tightness when $n\geq L+2$, take a target path of length $L+1$ from the root and a distracting chain containing the remaining $n-L-2$ vertices, ordered before the target branch. The target is outside the horizon at the root and is unreachable from the distracting chain. The policy explores and backtracks along that chain before following the target path, taking
\[
    2(n-L-2)+(L+1)=2n-L-3
\]
moves. The shortest-path distance is $L+1$, attaining the bound. When $n=L+2$, the distracting chain is empty and is omitted.
\end{proof}

\begin{corollary}[Conditional depth--efficiency implication]
\label{cor:k-layer-efficiency}
For every integer $k\geq3$, if a transformer implements the \emph{distance-minimizing} policy of~\cref{def:distance-minimizing-lookahead} with $L=1+2^{k-3}$, then its worst-case efficiency is
\begin{equation}
    \operatorname{eff}=
    \begin{cases}
        \displaystyle\frac{2^{k-3}+2}{2n-2^{k-3}-4},
          &n\geq2^{k-3}+3,\\[1em]
        1,&n<2^{k-3}+3.
    \end{cases}
\end{equation}
The bounded-reachability transformer of~\cref{thm:exponential-depth-advantage} is not asserted to satisfy this additional premise.
\end{corollary}

\begin{proof}
Substitute $L=1+2^{k-3}$ into~\cref{lem:lookahead-efficiency}.
\end{proof}

%% file: content/appendix/proof_learning_dynamics.tex
\section{Proof of~\cref{thm:learning_dynamics}}
\label{app:proof_learning_dynamics}

\subsection{Permutation symmetry and reduction to scalar parameters}
\label{app:proof_prop_symmetric}

\begin{proof}[Proof of~\cref{prop:symmetric_weight}]
Write $\theta=(\mA_1,\mA_2,\mB_1,\mB_2)$ and, for $\pi\in S_\gV$, let $\mP_\pi\in\sR^{N\times N}$ be its permutation matrix, $\mP_\pi\ve_v=\ve_{\pi(v)}$. Define the action
$\pi\cdot\theta:=(\mP_\pi\mA_1\mP_\pi^T,\mP_\pi\mA_2\mP_\pi^T,\mP_\pi\mB_1\mP_\pi^T,\mP_\pi\mB_2\mP_\pi^T)$; it is linear and orthogonal for the Frobenius inner product.

\emph{Step 1: equivariance of the network.} Relabeling the vertices of a prompt and a trace by $\pi$ replaces every node one-hot $\ve_v$ by $\mP_\pi\ve_v$ and leaves the special tokens, the positions, and the length unchanged, so the input embeddings become $\bm{\Pi}\mX$ with $\bm{\Pi}=\operatorname{diag}(\mP_\pi,\mP_\pi,\mP_\pi,\mP_\pi)$ on the four node blocks and the identity on the constant and positional coordinates. The fixed first layer (\cref{lem:first-layer-softmax-copy-relu}) computes its attention scores from the positional coordinates only, its value maps copy node blocks or the constant vector $\ones_N=\mP_\pi\ones_N$, and the ReLU and the residual act entrywise; hence its output transforms as $\mX^{(2)}\mapsto\bm{\Pi}\mX^{(2)}$ (exactly, also with leakage $\delta>0$). In the second layer, writing $\vx_j=(\va_j,\vb_j,\vc_j,\vd_j)$ for the four node blocks,
\[
    (\bm{\Pi}\vx_j)^T\mW_{KQ}^{(2,1)}[\pi\cdot\theta]\,\bm{\Pi}\vx_i
    =\va_j^T\mP_\pi^T\mP_\pi\mA_1\mP_\pi^T\mP_\pi\va_i+\vc_j^T\mP_\pi^T\mP_\pi\mA_2\mP_\pi^T\mP_\pi\va_i
    =\vx_j^T\mW_{KQ}^{(2,1)}[\theta]\,\vx_i ,
\]
and likewise for the second head. The value maps satisfy $\mW_V^{(2,h)}\bm{\Pi}\vx_j=\mP_\pi\mW_V^{(2,h)}\vx_j$, the output matrix $\mW_O^{(2)}$ only routes blocks, $\relu(\mP_\pi\vy)=\mP_\pi\relu(\vy)$, and the dynamic readout satisfies $\operatorname{diag}(\mI_{\pi(n)-1},0,\mI_{N-\pi(n)})=\mP_\pi\operatorname{diag}(\mI_{n-1},0,\mI_{N-n})\mP_\pi^T$. Hence the vertex logits satisfy $\TF(\bm{\Pi}\mX;\pi\cdot\theta)=\mP_\pi\TF(\mX;\theta)$ and $p_{\pi\cdot\theta}(\pi(v);\bm{\Pi}\mX)=p_\theta(v;\mX)$. Consequently, for every prompt, prefix, and trace,
\begin{equation}\label{eq:loss_equivariance}
    \ell_{\pi(\tau_\Prompt),\pi(\tau_{\rm pre}),\pi(\tau_0)}(\pi\cdot\theta)=\ell_{\tau_\Prompt,\tau_{\rm pre},\tau_0}(\theta).
\end{equation}

\emph{Step 2: invariance of the averaged loss.} Fix $\sigma\in S_\gV$. Applying~\cref{eq:loss_equivariance} with $\pi=\sigma$ to the instance relabeled by $\sigma^{-1}\pi'$ gives $\ell_{\pi'(\cdot)}(\sigma\cdot\theta)=\ell_{\sigma^{-1}\pi'(\cdot)}(\theta)$; averaging over $\pi'\in S_\gV$ and using that $\pi'\mapsto\sigma^{-1}\pi'$ is a bijection yields $\gL_{\rm train}(\sigma\cdot\theta)=\gL_{\rm train}(\theta)$ for all $\theta$ and $\sigma$.

\emph{Step 3: the fixed subspace is invariant.} A matrix commuting with all permutation matrices has constant diagonal and constant off-diagonal entries, because $S_N$ acts transitively on $\{(i,i)\}$ and on $\{(i,j):i\neq j\}$. Hence the fixed-point set $\{\theta:\sigma\cdot\theta=\theta\ \forall\sigma\}$ is exactly $\gF$, and $c\mE=c\mI+c\Eoff\in\gF$. Wherever $\gL_{\rm train}$ is differentiable, differentiating $\gL_{\rm train}\circ\sigma=\gL_{\rm train}$ and using that the action is orthogonal gives $\nabla\gL_{\rm train}(\sigma\cdot\theta)=\sigma\cdot\nabla\gL_{\rm train}(\theta)$; in particular $\nabla\gL_{\rm train}(\theta)\in\gF$ for $\theta\in\gF$, so the vector field $-\alpha\nabla\gL_{\rm train}$ is tangent to $\gF$. At a nonsmooth point the same chain rule holds for the Clarke generalized gradient, $\partial\gL_{\rm train}(\sigma\cdot\theta)=\sigma\cdot\partial\gL_{\rm train}(\theta)$, so for $\theta\in\gF$ the convex set $\partial\gL_{\rm train}(\theta)$ is $S_\gV$-invariant and therefore contains the group average of each of its elements; since the group average is the orthogonal projection $\Pi_\gF$ onto $\gF$, we get $\Pi_\gF\,\partial\gL_{\rm train}(\theta)\subset\partial\gL_{\rm train}(\theta)$. Together with the chain-rule inclusion $\partial(\gL_{\rm train}|_\gF)(\theta)\subset\Pi_\gF\,\partial\gL_{\rm train}(\theta)$ for the linear embedding of $\gF$, every solution of the gradient-flow inclusion restricted to $\gF$ is a solution of the full inclusion, and whenever $\gL_{\rm train}$ is differentiable along the solution it is the unique one.

\emph{Step 4: reduction to one instance.} For $\theta\in\gF$, \cref{eq:loss_equivariance} gives $\ell_{\pi(\cdot)}(\theta)=\ell_{(\cdot)}(\pi^{-1}\cdot\theta)=\ell_{(\cdot)}(\theta)$, hence $\gL_{\rm train}(\theta)=\ell_{\tau_\Prompt,\tau_{\rm pre},\tau_0}(\theta)$ on $\gF$.
\end{proof}

Since the weight matrices remain permutation invariant, we can transform the training on the whole matrices to training on the eight scalar parameters. Throughout the appendix,
\[
    \ell(a_1,b_1,a_2,b_2,g_1,d_1,g_2,d_2):=\ell_{\tau_\Prompt,\tau_\text{pre},\tau_0}\bigl(a_1\mI+b_1\Eoff,\ a_2\mI+b_2\Eoff,\ g_1\mI+d_1\Eoff,\ g_2\mI+d_2\Eoff\bigr)
\]
denotes the loss of the canonical instance viewed as a function of the eight scalars, and $\partial\ell/\partial a_i$ etc.\ denote its ordinary partial derivatives, with the convention $\relu'(x)=\ind_{x\geq0}$ at the kinks.
\begin{lemma}\label{lemma:equivalence_parameter}
    Let $(\mA_1,\mA_2,\mB_1,\mB_2)(t)$ be a solution of the gradient flow $\dot\theta=-\alpha\nabla_\theta\gL_{\text{train}}$ started in $\gF$, so that $\mA_1=a_1\mI+b_1\Eoff$, \ldots, $\mB_2=g_2\mI+d_2\Eoff$ by~\cref{prop:symmetric_weight}. Then the scalars solve
    \begin{align}\label{eq:scalar_flow}
        &\dot a_i=-\frac{\alpha}{N}\frac{\partial \ell}{\partial a_i},\qquad
        \dot g_i=-\frac{\alpha}{N}\frac{\partial \ell}{\partial g_i},\\
        &\dot b_i=-\frac{\alpha}{N(N-1)}\frac{\partial \ell}{\partial b_i},\qquad
        \dot d_i=-\frac{\alpha}{N(N-1)}\frac{\partial \ell}{\partial d_i},\qquad i\in\{1,2\},
    \end{align}
    that is, the scalars follow the gradient flow of the single-instance loss $\ell$ with the learning rate normalization
    \begin{align}
        \alpha_{a} = \alpha_{g} = \frac{\alpha}{N},\qquad\alpha_{b} = \alpha_{d} = \frac{\alpha}{N(N-1)}.
    \end{align}
\end{lemma}
\begin{proof}
    By Step 4 of the proof of~\cref{prop:symmetric_weight}, $\gL_{\text{train}}=\ell$ on $\gF$, and by~\cref{eq:loss_equivariance}, $\gL_{\text{train}}(\theta)=\frac{1}{N!}\sum_{\pi}\ell_{\tau_\Prompt,\tau_\text{pre},\tau_0}(\pi\cdot\theta)$ for every $\theta$. Differentiating at $\theta\in\gF$, where $\pi\cdot\theta=\theta$,
    \[
        \nabla_{\mA_1}\gL_{\text{train}}(\theta)=\frac{1}{N!}\sum_{\pi}\mP_\pi^T\,\mG\,\mP_\pi,\qquad \mG:=\nabla_{\mA_1}\ell_{\tau_\Prompt,\tau_\text{pre},\tau_0}(\theta),
    \]
    and for any matrix $\mG$ the conjugation average equals $\frac{\Tr\mG}{N}\mI+\frac{\sum_{k\neq l}G_{kl}}{N(N-1)}\Eoff$. By the chain rule, $\Tr\mG=\sum_k\partial\ell/\partial(\mA_1)_{kk}=\partial\ell/\partial a_1$ and $\sum_{k\neq l}G_{kl}=\partial\ell/\partial b_1$. Hence
    \[
        \nabla_{\mA_1}\gL_{\text{train}}=\frac{1}{N}\frac{\partial\ell}{\partial a_1}\,\mI+\frac{1}{N(N-1)}\frac{\partial\ell}{\partial b_1}\,\Eoff,
    \]
    and likewise for $\mA_2,\mB_1,\mB_2$. Comparing the coefficients of $\mI$ and $\Eoff$ in $\dot{\mA}_1=-\alpha\nabla_{\mA_1}\gL_{\text{train}}$ gives~\cref{eq:scalar_flow}. At a kink of the ReLU the same identity holds for every fixed permutation-equivariant selection of the derivative, in particular for $\relu'(0)=1$.
\end{proof}

To calculate the gradient with respect to each parameter, we need the formulation of the logits $\TF(\mX,\theta)$. Since $\TF(\mX,\theta)$ is the sum of the attention head values after ReLU, we first compute the node logits of each head separately, for a general directed graph and a general valid DFS prefix $(n_0=n_\Root,n_1,\ldots,n_t)$.
We summarize the node logits in~\cref{tab:forward_head} and~\cref{tab:backward_head}, separately.

\begin{table}[pbht]
    \centering
    \begin{center}
\begin{tabular}{ll}
\toprule
{Node type ($v\neq n_t$)} &  {Node logits} \\
\midrule
Unvisited child of $n_t$ ($m_v=0$) & $a_2 - b_2 + b_2 \cdot d_v^{\text{in,total}}$ \\
Visited child of $n_t$ ($m_v\geq1$) &  $a_2 - b_2 + b_2 \cdot d_v^{\text{in,total}} - b_1 m_v$ \\
Non-child &  $b_2 \cdot d_v^{\text{in,total}} - b_1 m_v$ \\
\bottomrule
\end{tabular}
\end{center}
    \caption{Node logits of the forward head (pre-ReLU). A child of $n_t$ is a vertex $v$ with $(n_t\to v)\in\gE$. $m_v$ is the number of occurrences of $v$ in $(n_\Root,n_1,\ldots,n_t)$, i.e., in the CoT trace including the root token. $d_v^{\text{in,total}}$ is the in-degree of $v$ in the prompt's edge list (self-loops included). The coordinate $v=n_t$ is masked by the dynamic readout.}
    \label{tab:forward_head}
\end{table}

\begin{table}[tp!b]
    \centering
\begin{center}\small
\renewcommand{\arraystretch}{1.4}
\begin{tabular}{lll}
\toprule
{Node type} &  If $v = n_\Root$?  & {Node logits} \\
\midrule
Parent of $n_t$ & No & $g_1 - d_1$ \\
Parent of $n_t$  & Yes &  $g_1 - d_1 + d_2$ \\
Tree child of $n_t$ (entered from $n_t$)  & No  & $g_1 - g_2 - d_1 + d_2$ \\
$n_\Root$, neither parent nor $n_t$ & Yes  & $d_2$ \\
Any other $v\neq n_t$ & No & $0$ \\
$n_t$  & No   & $k(g_2 - g_1 + d_1 - d_2) - g_1 - d_2$ \\
$n_t$  & Yes   & $k(g_2 - g_1 + d_1 - d_2) - g_1$ \\
\bottomrule
\end{tabular}
\end{center}
    \caption{Node logits of the backtracking head (pre-ReLU). $k$ is the number of children of $n_t$ that the DFS has entered from $n_t$; since the search is currently at $n_t$, each of these excursions has been completed. ``Any other $v$'' covers unvisited children, unvisited non-children, and visited vertices that are neither the parent, the root, nor a tree child of $n_t$. A visited out-neighbor of $n_t$ that was reached by another route (the parent, the root, an ancestor, or a vertex finished in another branch) is not a tree child and falls into the parent, root, or ``any other'' row.}
    \label{tab:backward_head}
\end{table}

\begin{proof}[Derivation of the tables]
Let the query be the last token, whose first block is $n_t$. Since $\mA_1 n_t=a_1 n_t+b_1(\bone-n_t)$, and likewise for $\mA_2,\mB_1,\mB_2$, the only keys with nonzero score and nonzero value are, for the forward head, the edge-target tokens $(0,0,s_i,t_i)$, with score $a_2$ if $s_i=n_t$ and $b_2$ otherwise and value $+t_i$, and the root and CoT tokens $(n_s,\cdot,0,0)$, $0\le s\le t$, with score $a_1$ if $n_s=n_t$ and $b_1$ otherwise and value $-n_s$. Hence, for $v\neq n_t$,
\[
    (\mathrm{head}_1)_v=(a_2-b_2)\ind[(n_t\to v)\in\gE]+b_2 d_v^{\mathrm{in,total}}-b_1 m_v ,
\]
which is \cref{tab:forward_head}. For the backtracking head only the root token, with score $g_1\ind[n_\Root=n_t]+d_1\ind[n_\Root\ne n_t]$ and value $-n_\Root$, and the CoT tokens $s\ge1$, with score $\sigma_s=d_1+d_2+(g_1-d_1)\ind[n_s=n_t]+(g_2-d_2)\ind[n_{s-1}=n_t]$ and value $n_{s-1}-n_s$, contribute. For $v\ne n_t$ the $(d_1+d_2)$-part telescopes to $(d_1+d_2)\ind[v=n_\Root]$, so
\[
    (\mathrm{head}_2)_v=(g_1-d_1)A_{v\to n_t}-(g_2-d_2)A_{n_t\to v}+d_2\ind[v=n_\Root],
\]
where $A_{x\to y}$ is the number of CoT steps $x\to y$. In a valid DFS prefix, $n_t$ is entered once from its stack parent if $n_t\ne n_\Root$, and otherwise only by the $k$ returns from the children it has entered, each of which it left exactly once; it has not yet left towards its parent. Substituting gives the rows for $v\ne n_t$, every remaining vertex having value $0$. For $v=n_t$, each arrival contributes $-(g_1+d_2)$, each departure $+(g_2+d_1)$, and the root token $-g_1$ if $n_t=n_\Root$, which gives the last two rows.
\end{proof}

By~\cref{lemma:equivalence_parameter}, we only need to analyze the gradient of the single-instance loss $\ell$ on the canonical data $\gE_0,\tau_0$. Recall the initialization~\cref{eq:initialization}: $a_1=b_1=a_2=g_1=g_2=d_2=c_0$ and $b_2=d_1=c_0'$ with $0<c_0'<c_0$. Throughout we write
\begin{align}
    u:=a_2-b_1,\qquad v:=g_1-d_1-g_2+d_2,\qquad E:=e^{u_++v_+},
\end{align}
and we use the following region of parameter space:
\begin{align}\label{eq:invariant_region}
    \gR:=\bigl\{\,b_1>0,\ \ b_2<b_1,\ \ a_2>0,\ \ g_1-d_1> 0\,\bigr\}.
\end{align}
The region is open on purpose: on the faces $a_2=0$ and $g_1=d_1$ the ReLUs $(a_2)_+$ and $(g_1-d_1)_+$ are at their kinks, where a Filippov solution may select any value in $[0,1]$ for their derivative, so the gradient formulas below hold on $\gR$ but not on these faces.
\begin{lemma}[Scalar parameter gradient on canonical data]\label{lemma:scalar_grad}
There is a constant $c_\ast=c_\ast(K)>0$ such that the following holds for the flow~\cref{eq:scalar_flow} started from~\cref{eq:initialization} with $0<c_0'<c_0\leq c_\ast N^{-2}$.
\begin{enumerate}
    \item \emph{(Invariant region.)} The trajectory stays in $\gR$ for all $t\geq0$. More precisely, $a_1\equiv c_0$; $b_1$ is non-decreasing and $b_2$ is non-increasing with $0\leq b_2\leq c_0'$, so $b_1\geq c_0>0$ and $b_2<b_1$; $u=a_2-b_1\geq 0$, hence $a_2\geq b_1\geq c_0>0$; $g_1-d_1>0$; $g_2-d_2$ is non-decreasing with $g_2-d_2\geq0$; and $d_2$ is non-increasing.
    \item \emph{(Initial phase.)} With $T_0:=\frac{N}{4(K+1)\alpha}$, we have $b_2(t)=0$ and $d_2(t)\leq 0$ for all $t\geq T_0$.
    \item \emph{(Gradient after the initial phase.)} For $t\geq T_0$,
\allowdisplaybreaks
        \begin{align}
    & \frac{\partial  \ell}{\partial a_1} = 0,\\
    &\begin{aligned}
                \frac{\partial  \ell}{\partial a_2} = - \frac{e^{ g_1-d_1}  + N-2}{D_1} -\sum_{i = 1}^K & \frac{e^{ g_1 - d_1}  +N-K - 1 +  (i-1)E \ind_{u < 0}}{D_{2,i}}
        +\frac{K E\ind_{u\geq 0}}{D_4},
    \end{aligned}\\
    &\begin{aligned}
        \frac{\partial  \ell}{\partial b_1} =
         - \sum_{i=1}^K \frac{(i-1)E\ind_{u\geq 0}}{D_{2,i}}
        - \frac{KE\ind_{u\geq 0} }{D_4},
    \end{aligned}\\
    &\begin{aligned}
        b_2\equiv 0\ \ \text{(so the selected value of }\partial\ell/\partial b_2\text{ along the flow is }0\text{)},\qquad (b_2)_+=(b_2-b_1)_+=0,
    \end{aligned}\\
    &\begin{aligned}
         \frac{\partial  \ell}{\partial g_1} = \frac{e^{g_1-d_1} }{D_1}+ \sum_{i=1}^K \frac{(i-1)E\ind_{v \geq 0} + e^{ g_1 - d_1} }{D_{2,i}}
          - \frac{K(N-1) }{D_{3}}
          -  \frac{N - K - 1 + KE\ind_{v< 0} }{D_4},
    \end{aligned}\\
    &\begin{aligned}
        \frac{\partial  \ell}{\partial g_2} &= - \sum_{i=1}^K \frac{(i-1)E\ind_{v\geq 0}}{D_{2,i}} - \frac{KE \ind_{v\geq 0}}{D_4},
    \end{aligned}\\
    &\begin{aligned}
        \frac{\partial  \ell}{\partial d_1 } &= - \frac{\partial  \ell}{\partial g_1},\qquad
        \frac{\partial  \ell}{\partial d_2 } = - \frac{\partial  \ell}{\partial g_2}\ \ \text{wherever } d_2<0,
    \end{aligned}
    \end{align}
    where the denominators are
    \begin{align}
    D_1 =& e^{a_2} + e^{ g_1-d_1}  + N-2,\\
    D_{2,i} =& (K-i+1)e^{a_2} + (i-1)E +e^{ g_1 - d_1}  +N-K - 1,\\
    D_{3} = &e^{g_1 - d_1} + N-1,\\
    D_4 =& KE + e^{ g_1 - d_1} + N - K - 1.
    \end{align}
\end{enumerate}
\end{lemma}
Note that the relation $\partial\ell/\partial d_1=-\partial\ell/\partial g_1$ holds for all $t\geq0$, because by~\cref{tab:backward_head} every unmasked logit depends on $(g_1,d_1)$ only through $g_1-d_1$; the relation for $(g_2,d_2)$ holds wherever $d_2<0$; at times $t\geq T_0$ with $d_2=0$ it can fail, but $g_2+(N-1)d_2$ is still conserved (Step 5 of the proof).
\begin{proof}
    For the cross-entropy loss $-\log(p_\theta(v;\mX))$, the gradient with respect to a parameter $\theta$ is
    \begin{align}\label{eq:ce_grad}
        \frac{\partial (-\log(p_\theta(v;\mX)))}{\partial \theta} = - \frac{\partial\TF(\mX,\theta)_v}{\partial \theta} + \sum_{v'\in \gV} p_\theta(v';\mX) \frac{\partial\TF(\mX,\theta)_{v'}}{\partial \theta},
    \end{align}
    and the gradient of $\ell$ is the sum of~\cref{eq:ce_grad} over the $2K+2$ steps of $\tau_0$.

   \emph{Step 1: node logits along the canonical trace.} In the canonical trace, DFS starts from node $N-3$ and moves forward to $1$. After exploring all $K$ child nodes of $1$, it then returns to $N-3$. Therefore, there are four types of steps in the trajectory. We next compute the node logits for these four steps from~\cref{tab:forward_head,tab:backward_head}; all nodes $2,\ldots,K+1$ have in-degree one, and the nodes $N-2,N-3,1$ have in-degree one as well. Recall that the logit of the current node $n_t$ is masked to $0$, and that the remaining $N-K-4$ vertices are isolated and unvisited, so their logits are $0$. In the following, we use $p(v)$ as an abbreviation of $p_\theta(v;\mX)$.

    \paragraph{Case 1: $N-3\to 1$.}
    The logits $\TF(\mX,\theta)$ of all nodes are
    \begin{itemize}
        \item $v=1$: $\TF(\mX,\theta)_v = (a_2)_+$
        \item $v=N-2$ (parent of $N-3$): $\TF(\mX,\theta)_v = (b_2-b_1)_+ + (g_1-d_1)_+$
        \item $v=N-1$ (root): $\TF(\mX,\theta)_v = (-b_1)_+ + (d_2)_+$
        \item $2\leq v\leq K+1$: $\TF(\mX,\theta)_v = (b_2)_+ $
        \item Other: $\TF(\mX,\theta)_v = 0 $
    \end{itemize}

    \paragraph{Case 2: $1\to i$ for $2\leq i \leq K+1$.}
    The logits $\TF(\mX,\theta)$ of all nodes are
    \begin{itemize}
        \item $i\leq v\leq K+1$ (unvisited children): $\TF(\mX,\theta)_v = (a_2)_+$.
        \item $2\leq v\leq i -1$ (tree children, each visited once): $\TF(\mX,\theta)_v = (a_2 -  b_1)_+ + (g_1 - d_1 - g_2 + d_2)_+$.
        \item $v=N-3$ (parent) : $\TF(\mX,\theta)_v = (b_2 - b_1)_+ + (g_1 - d_1)_+$.
        \item  $v=N-2$: $\TF(\mX,\theta)_v = (b_2-b_1)_+$.
        \item $v=N-1$: $\TF(\mX,\theta)_v = ( - b_1)_+ + (d_2)_+$
        \item Other: $\TF(\mX,\theta)_v =0$
    \end{itemize}

    \paragraph{Case 3: $i \to 1$ for $2\leq i \leq K+1$.} At this step node $1$ has occurred $i-1$ times in the trace. The logits $\TF(\mX,\theta)$ of all nodes are
    \begin{itemize}
        \item $v=1$ (parent): $\TF(\mX,\theta)_v = (b_2-(i-1)b_1)_+ + (g_1 - d_1 )_+$.
        \item $v=N-2$ or $N-3$: $\TF(\mX,\theta)_v = (b_2 -  b_1)_+ $
        \item $v=N-1$: $\TF(\mX,\theta)_v = ( - b_1)_+ + (d_2)_+$
        \item $2\leq v\leq i-1$: $\TF(\mX,\theta)_v = (b_2 - b_1)_+ $.
        \item  $i+1\leq v\leq K+1$: $\TF(\mX,\theta)_v = (b_2)_+$.
        \item Other: $\TF(\mX,\theta)_v =0$
    \end{itemize}

    \paragraph{Case 4: $1\to N-3$.}
    The logits $\TF(\mX,\theta)$ of all nodes are
    \begin{itemize}
        \item $v=N-3$ (parent): $\TF(\mX,\theta)_v = (b_2-b_1)_+ + (g_1 - d_1)_+$
        \item $v=N-2$: $\TF(\mX,\theta)_v = (b_2-b_1)_+ $
        \item $v=N-1$: $\TF(\mX,\theta)_v = (-b_1)_+ + (d_2)_+$
        \item $2\leq v\leq K+1$ (tree children): $\TF(\mX,\theta)_v = (a_2 - b_1)_+ + (g_1 - d_1 - g_2 + d_2)_+$
        \item Other: $\TF(\mX,\theta)_v = 0 $
    \end{itemize}
    In the gradient formulas below we index the $K$ steps of Case 2 and of Case 3 by $i\in[K]$, i.e., we shift the child label by one, so that at the $i$-th such step there are $K-i+1$ unvisited and $i-1$ visited children of node $1$.

    \emph{Step 2: the gradient on $\gR$.} On the region $\gR$ of~\cref{eq:invariant_region} we have $b_2-b_1<0$, $b_2-(i-1)b_1\leq b_2-b_1<0$ and $-b_1<0$, so the terms $(b_2-b_1)_+$, $(b_2-(i-1)b_1)_+$, $(-b_1)_+$ and their derivatives vanish identically, while $(a_2)_+=a_2$ and $(g_1-d_1)_+=g_1-d_1$ with derivative $1$, because $a_2>0$ and $g_1-d_1>0$ on the open region $\gR$. The remaining kinks $u=0$, $v=0$, $b_2=0$ and $d_2=0$ can lie in $\gR$; there the indicators $\ind_{u\geq0},\ind_{v\geq0},\ind_{b_2\geq0},\ind_{d_2\geq0}$ below stand for the values in $[0,1]$ selected by the Filippov solution (the weights of the convex combination of the one-sided fields). Applying~\cref{eq:ce_grad} to the four cases gives, on $\gR$,
\allowdisplaybreaks
    \begin{align}
    & \frac{\partial  \ell}{\partial a_1} = 0,\label{eq:grad_R_a1}\\
    &\begin{aligned}
                \frac{\partial  \ell}{\partial a_2} &= -1+ \frac{e^{a_2}}{D_1}
        + \sum_{i = 1}^K \Bigl(-1 + \frac{(K-i+1)e^{a_2} + (i-1)E \ind_{u \geq 0}}{D_{2,i}}\Bigr)
        + \frac{K E\ind_{u\geq 0}}{D_4},
    \end{aligned}\label{eq:grad_R_a2}\\
    &\begin{aligned}
        \frac{\partial  \ell}{\partial b_1} &= - \sum_{i=1}^K \frac{(i-1)E\ind_{u\geq 0}}{D_{2,i}} - \frac{KE\ind_{u\geq 0}}{D_4},
    \end{aligned}\label{eq:grad_R_b1}\\
    &\begin{aligned}
        \frac{\partial  \ell}{\partial b_2} &= e^{(b_2)_+}\ind_{b_2\geq 0}\Bigl[\frac{K}{D_1}+\sum_{i=1}^K\frac{K-i}{D_{3,i}}\Bigr],
    \end{aligned}\label{eq:grad_R_b2}\\
    &\begin{aligned}
         \frac{\partial  \ell}{\partial g_1} &= \frac{e^{g_1-d_1}}{D_1}
         + \sum_{i=1}^K \frac{(i-1)E\ind_{v \geq 0} + e^{g_1 - d_1}}{D_{2,i}}
         + \sum_{i=1}^K \Bigl(-1 + \frac{e^{g_1 - d_1}}{D_{3,i}}\Bigr)
         -1 +  \frac{KE\ind_{v\geq 0} + e^{g_1 - d_1}}{D_4},
    \end{aligned}\label{eq:grad_R_g1}\\
    &\begin{aligned}
        \frac{\partial  \ell}{\partial g_2} &= - \sum_{i=1}^K \frac{(i-1)E\ind_{v\geq 0}}{D_{2,i}} - \frac{KE \ind_{v\geq 0}}{D_4},
    \end{aligned}\label{eq:grad_R_g2}\\
    &\begin{aligned}
        \frac{\partial  \ell}{\partial d_1 } &= - \frac{\partial  \ell}{\partial g_1},
    \end{aligned}\label{eq:grad_R_d1}\\
    &\begin{aligned}
        \frac{\partial  \ell}{\partial d_2 } &= - \frac{\partial  \ell}{\partial g_2} + e^{(d_2)_+}\ind_{d_2 \geq 0} \left[\frac{ 1 }{D_1} + \sum_{i=1}^K \frac{1}{D_{2,i}} + \sum_{i=1}^K \frac{1}{D_{3,i}} + \frac{1}{D_4}\right],
    \end{aligned}\label{eq:grad_R_d2}
    \end{align}
    where the denominators are the softmax normalizers of the four cases,
    \begin{align}
        &\begin{aligned}
            D_1 = e^{a_2} + e^{g_1-d_1} +  e^{(d_2)_+}
              + K e^{(b_2)_+} + N-K-3,
        \end{aligned}\\
        &\begin{aligned}
             D_{2,i} = (K-i+1)e^{a_2} + (i-1)E + e^{g_1 - d_1} + 1+ e^{(d_2)_+} + N-K - 3,
        \end{aligned}
      \\
    &\begin{aligned}
        D_{3,i} = e^{g_1 - d_1} +2 + e^{(d_2)_+} + (i-1)
        + (K-i) e^{(b_2)_+ } + N-K-3,
    \end{aligned}\\
      & \begin{aligned}
          D_4 = KE + e^{g_1 - d_1} + 1
          + e^{(d_2)_+} + N - K - 3.
      \end{aligned}
    \end{align}
    Each denominator has exactly $N$ terms, one per vertex token, including the masked current node with weight $e^0=1$. In particular $D_1,D_{2,i},D_{3,i},D_4\geq N$ on $\gR$, and each of the $2K+2$ steps contributes a quantity in $[-1,1]$ to every partial derivative, so that
    \begin{align}\label{eq:grad_bound}
        \Bigl|\frac{\partial\ell}{\partial\theta}\Bigr|\leq 2K+2\qquad\text{on }\gR,\ \text{for every scalar }\theta .
    \end{align}
    The identity~\cref{eq:grad_R_d1} holds on all of parameter space, since $g_1,d_1$ enter the unmasked logits only through $g_1-d_1$; the extra term in~\cref{eq:grad_R_d2} comes from the logit $(d_2)_+$ of the root in every step.

    \emph{Step 3: $\gR$ is invariant.} The initial point lies in $\gR$: $b_1=c_0>0$, $b_2=c_0'<c_0=b_1$, $a_2=c_0>0$, $g_1-d_1=c_0-c_0'>0$. As long as the trajectory is in $\gR$, \cref{eq:grad_R_a1,eq:grad_R_b1,eq:grad_R_b2,eq:grad_R_g2} and~\cref{eq:scalar_flow} give $\dot a_1=0$, $\dot b_1\geq 0$, $\dot b_2\leq 0$ and $\dot g_2\geq0$; moreover $\dot d_2=-\frac{\alpha}{N(N-1)}\frac{\partial\ell}{\partial d_2}\leq 0$ because both terms of~\cref{eq:grad_R_d2} are nonnegative, and
    \begin{align}\label{eq:y2_dot}
        \frac{\dd}{\dd t}(g_2-d_2)=-\frac{\alpha}{N-1}\frac{\partial\ell}{\partial g_2}+\frac{\alpha}{N(N-1)}\,e^{(d_2)_+}\ind_{d_2\geq0}\Bigl[\frac1{D_1}+\sum_i\frac1{D_{2,i}}+\sum_i\frac1{D_{3,i}}+\frac1{D_4}\Bigr]\geq 0,
    \end{align}
    with strict inequality whenever $v\geq 0$, because then $-\partial\ell/\partial g_2\geq KE/D_4>0$. Hence $b_1\geq c_0$, $b_2-b_1$ is non-increasing and stays negative, and $g_2-d_2\geq 0$ is non-decreasing. Moreover $b_2$ never becomes negative: on $\{b_2<0\}\cap\gR$ the field~\cref{eq:grad_R_b2} vanishes, so if $b_2(t_3)<0$ and $t'=\sup\{t\leq t_3: b_2(t)\geq 0\}$, then $b_2$ is constant on $(t',t_3]$ and $b_2(t_3)=b_2(t')=0$ by continuity, a contradiction; hence $0\leq b_2\leq c_0'$. Since $\gR$ is open and these inequalities keep $b_1>0$ and $b_2<b_1$, a continuous trajectory can leave $\gR$ only through one of the two faces $\{a_2=0\}$ or $\{g_1-d_1=0\}$. We show that neither face is reached. The arguments use the field only on $\gR$, in a neighborhood of the faces, and never on the faces themselves, where the selections for $(a_2)_+$ and $(g_1-d_1)_+$ are free.

    \emph{The face $a_2=0$.} On the open set $\{u<0\}\cap\gR$ we have $\ind_{u\geq0}=0$, so by~\cref{eq:grad_R_a2,eq:grad_R_b1}
    \[
        \frac{\partial\ell}{\partial a_2}=-\Bigl(1-\frac{e^{a_2}}{D_1}\Bigr)-\sum_{i=1}^K\Bigl(1-\frac{(K-i+1)e^{a_2}}{D_{2,i}}\Bigr)<0,\qquad\frac{\partial\ell}{\partial b_1}=0,
    \]
    because $D_1>e^{a_2}$ and $D_{2,i}>(K-i+1)e^{a_2}$. Hence $u'=\dot a_2-\dot b_1>0$ on $\{u<0\}\cap\gR$; since $u(0)=0$, the trajectory never enters $\{u<0\}$, so $u\geq0$ and $a_2\geq b_1\geq c_0>0$ for all $t$. In particular the face $a_2=0$ is never reached.

    \emph{The face $g_1-d_1=0$.} Since $v(0)=g_1(0)-d_1(0)-(g_2-d_2)(0)=c_0-c_0'>0$, by continuity $v>0$ on some interval $[0,\delta)$, where~\cref{eq:y2_dot} gives $\frac{\dd}{\dd t}(g_2-d_2)\geq\frac{\alpha}{N-1}\frac{KE}{D_4}>0$; as $g_2-d_2$ is non-decreasing, $(g_2-d_2)(t)>0$ for all $t>0$. Fix $a\in(0,\delta)$ so small that $g_1-d_1>0$ on $[0,a]$, and let $\epsilon_0:=\min\{(g_2-d_2)(a),\log 2\}>0$. For $t\geq a$, at every state of $\gR$ with $g_1-d_1<\epsilon_0$ we have $v=(g_1-d_1)-(g_2-d_2)<\epsilon_0-(g_2-d_2)(a)\leq 0$, so $\ind_{v\geq 0}=0$ in a neighborhood of the state, and with $e^{g_1-d_1}\leq 2$ and all denominators at least $N$, \cref{eq:grad_R_g1} gives
    \begin{align}
        \frac{\partial\ell}{\partial g_1}
        \leq \frac{2}{N}+\frac{2K}{N}-K\Bigl(1-\frac2N\Bigr)-\Bigl(1-\frac2N\Bigr)
        =(K+1)\Bigl(\frac{4}{N}-1\Bigr)<0 .
    \end{align}
    Hence $\frac{\dd}{\dd t}(g_1-d_1)=-\frac{\alpha}{N}\frac{\partial\ell}{\partial g_1}+\frac{\alpha}{N(N-1)}\frac{\partial\ell}{\partial d_1}=-\frac{\alpha}{N-1}\frac{\partial\ell}{\partial g_1}>0$ on the whole set $\{g_1-d_1<\epsilon_0\}\cap\gR$ (for $t\geq a$), not only at a single time, and the one-sided barrier argument gives $g_1-d_1\geq\min\{(g_1-d_1)(a),\epsilon_0\}>0$ for all $t\geq a$. Thus $g_1-d_1>0$ for all $t\geq0$. This proves item 1.

    \emph{Step 4: the initial phase.} By~\cref{eq:grad_bound} and~\cref{eq:scalar_flow}, every scalar moves at speed at most $(2K+2)\alpha/N$, so on $[0,T_0]$ with $T_0=\frac{N}{4(K+1)\alpha}$ every parameter changes by at most $\frac12$; since $0<c_0'<c_0\leq c_\ast N^{-2}<\frac14$ (by the definition of $c_\ast$ below), all parameters lie in $[-\frac12,1]$ on $[0,T_0]$. Consequently every exponent appearing in the denominators is bounded by an absolute constant, and there exist absolute constants $C_1,C_2\geq0$ such that
    \begin{align}
        \max_{i} \{D_1,D_{2,i},D_{3,i},D_4\} \leq N+ C_1K+C_2\qquad\text{on }[0,T_0].
    \end{align}
    While $b_2>0$ (so that $\ind_{b_2\geq0}=1$), \cref{eq:grad_R_b2} gives
    \begin{align}
        \frac{\partial  \ell}{\partial b_2} \geq \frac{K + \frac{K(K-1)}{2}}{N+ C_1K+C_2}\geq\frac{K}{N+ C_1K+C_2},
        \qquad\text{so}\qquad
        \dot b_2\leq-\frac{\alpha}{N(N-1)}\,\frac{K}{N+C_1K+C_2},
    \end{align}
    and while $d_2>0$, \cref{eq:grad_R_d2} gives
    \begin{align}
         \frac{\partial  \ell}{\partial d_2} \geq\frac{2K+2}{N+ C_1K+C_2} \geq \frac{K}{N+ C_1K+C_2},
         \qquad\text{so}\qquad
        \dot d_2\leq-\frac{\alpha}{N(N-1)}\,\frac{K}{N+C_1K+C_2}.
    \end{align}
    Over the time interval $[0,T_0]$ each of $b_2,d_2$ therefore decreases, as long as it is positive, by at least
    \begin{align}
        \frac{\alpha}{N(N-1)}\, \frac{K}{N+ C_1K+C_2}\, \frac{N}{4(K+1)\alpha }
        &=  \frac{K}{4(K+1)(N-1)(N+C_1K+C_2)}\nonumber\\
        &\geq \frac{K\,N^{-2}}{4(K+1)(1+C_1K+C_2)}=:c_\ast N^{-2},
    \end{align}
    where we used $(N-1)(N+C_1K+C_2)\leq N^2(1+C_1K+C_2)$. Since $b_2(0)=c_0'<c_0\leq c_\ast N^{-2}$ and $d_2(0)=c_0\leq c_\ast N^{-2}$, both $b_2$ and $d_2$ reach $0$ before time $T_0$. Since $b_2$ is non-increasing and $b_2\geq0$ (Step 3), $b_2=0$ for all $t\geq T_0$; since $d_2$ is non-increasing, $d_2\leq0$ for all $t\geq T_0$. This proves item 2.

    \emph{Step 5: simplification.} For $t\geq T_0$ we have $b_2=0$, $e^{(b_2)_+}=e^{(d_2)_+}=1$, and $\ind_{d_2\geq 0}=0$ wherever $d_2<0$. The term with $\ind_{d_2\geq0}$ in~\cref{eq:grad_R_d2} enters only $\dot d_2$ and does not affect the reduced system: the set $\{t\geq T_0: d_2(t)=0\}$ is an interval, as $d_2\leq0$ is non-increasing; on its interior $\dot d_2=0$, so both nonnegative terms of $\partial\ell/\partial d_2$ in~\cref{eq:grad_R_d2} vanish, and after it $d_2<0$. In both cases $\dot g_2+(N-1)\dot d_2=-\frac{\alpha}{N}\bigl(\frac{\partial\ell}{\partial g_2}+\frac{\partial\ell}{\partial d_2}\bigr)=0$ a.e., so $g_2+(N-1)d_2$ is constant on $[T_0,\infty)$ and $\frac{\dd}{\dd t}(g_2-d_2)=\frac{N}{N-1}\dot g_2=-\frac{\alpha}{N-1}\frac{\partial\ell}{\partial g_2}$ for a.e.\ $t\geq T_0$. Substituting into~\cref{eq:grad_R_a2,eq:grad_R_b1,eq:grad_R_g1,eq:grad_R_g2,eq:grad_R_d1,eq:grad_R_d2} and using
    \begin{align*}
        -1+\frac{e^{a_2}}{D_1}&=-\frac{e^{g_1-d_1}+N-2}{D_1},\\
        -1+\frac{(K-i+1)e^{a_2}+(i-1)E\ind_{u\geq0}}{D_{2,i}}&=-\frac{e^{g_1-d_1}+N-K-1}{D_{2,i}}-\frac{(i-1)E\ind_{u<0}}{D_{2,i}},\\
        \sum_{i=1}^K\Bigl(-1+\frac{e^{g_1-d_1}}{D_{3,i}}\Bigr)&=-\frac{K(N-1)}{D_3},\\
        -1+\frac{KE\ind_{v\geq0}+e^{g_1-d_1}}{D_4}&=-\frac{N-K-1+KE\ind_{v<0}}{D_4},
    \end{align*}
    where now $D_{3,i}=e^{g_1-d_1}+N-1=:D_3$ no longer depends on $i$, gives the expressions in item 3.
\end{proof}

By~\cref{lemma:scalar_grad}, we can aggregate the parameters and deal with a more compact system as follows.
\begin{corollary}[Parameter aggregation]\label{coro:parameter_aggre}
    Let $x_1 = a_2,x_2 = b_1,y_1 = g_1 - d_1,y_2 = g_2-d_2$ and $u= x_1-x_2,v=y_1-y_2$. Under the conditions of~\cref{lemma:scalar_grad}, for $t\geq T_0$ the quadruple $(x_1,x_2,y_1,y_2)$ solves the closed four-variable system
    \begin{align}
        \dx_1 &= \frac{\alpha}{N} \Bigl[ \frac{e^{ y_1}  + N-2}{D_1} +\sum_{i = 1}^K  \frac{e^{ y_1}  +N-K - 1 +  (i-1)e^{u_+ + v_+} \ind_{u < 0}}{D_{2,i}}- \frac{K e^{u_+ + v_+}\ind_{u\geq 0}}{D_4}\Bigr]\\
        \dx_2 &=  \frac{\alpha}{N(N-1)} \Bigl[\sum_{i=1}^K \frac{(i-1)e^{u_+ + v_+}\ind_{u\geq 0}}{D_{2,i}}
        + \frac{Ke^{u_+ + v_+}\ind_{u\geq 0} }{D_4}\Bigr]\\
        \dy_1 &= \frac{\alpha}{N-1}\Bigl[\frac{K(N-1)}{D_{3}}+  \frac{N - K - 1 + Ke^{u_+ + v_+}\ind_{v< 0} }{D_4}
          -\frac{e^{y_1} }{D_1}- \sum_{i=1}^K \frac{(i-1)e^{u_+ + v_+}\ind_{v \geq 0} + e^{ y_1} }{D_{2,i}}
          \Bigr]\\
          \dy_2 & = \frac{\alpha}{N-1}\Bigl[\sum_{i=1}^K \frac{(i-1)e^{u_+ + v_+}\ind_{v\geq 0}}{D_{2,i}} + \frac{Ke^{u_+ + v_+} \ind_{v\geq 0}}{D_4}\Bigr],
    \end{align}
    where the denominators are
    \begin{align}
    D_1 =& e^{x_1} + e^{ y_1}  + N-2\\
    D_{2,i} =& (K-i+1)e^{x_1} + (i-1)e^{u_+ + v_+} +e^{ y_1}  +N-K - 1\\
    D_{3} = &e^{y_1} + N-1\\
    D_4 =& Ke^{u_+ + v_+} + e^{ y_1} + N - K - 1 .
    \end{align}
    The remaining four scalars are recovered affinely: $a_1\equiv c_0$; $b_2\equiv 0$ for $t\geq T_0$; and since $\partial\ell/\partial d_i=-\partial\ell/\partial g_i$ while the learning rates are $\alpha/N$ and $\alpha/(N(N-1))$, we have $\dot g_i=-(N-1)\dot d_i$, so that $g_1+(N-1)d_1\equiv c_0+(N-1)c_0'$ for all $t\geq0$ and $g_2+(N-1)d_2$ is constant for $t\geq T_0$; hence
    \begin{align}\label{eq:gd_recovery}
        g_1&=\frac{N-1}{N}\,y_1+\frac{c_0+(N-1)c_0'}{N},& d_1&=-\frac{1}{N}\,y_1+\frac{c_0+(N-1)c_0'}{N},\\
        g_2&=\frac{N-1}{N}\,y_2+O(1),& d_2&=-\frac{1}{N}\,y_2+O(1),
    \end{align}
    where the $O(1)$ terms are constants depending on $N,K,c_0,c_0'$. At time $T_0$ the state satisfies $u\geq0$, $x_2>0$, $y_1>0$ and $y_2\geq0$, and along the system $y_2$ is non-decreasing.
\end{corollary}
\begin{proof}
    By~\cref{eq:scalar_flow}, $\dx_1=-\frac{\alpha}{N}\frac{\partial\ell}{\partial a_2}$, $\dx_2=-\frac{\alpha}{N(N-1)}\frac{\partial\ell}{\partial b_1}$, and, using $\partial\ell/\partial d_1=-\partial\ell/\partial g_1$ and, for $i=2$, Step 5 of the proof of~\cref{lemma:scalar_grad},
    \[
        \dy_i=\dot g_i-\dot d_i=-\frac{\alpha}{N}\frac{\partial\ell}{\partial g_i}-\frac{\alpha}{N(N-1)}\frac{\partial\ell}{\partial g_i}=-\frac{\alpha}{N-1}\frac{\partial\ell}{\partial g_i}.
    \]
    Substituting item 3 of~\cref{lemma:scalar_grad} gives the displayed system; it depends on the eight scalars only through $x_1,x_2,y_1,y_2$. The recovery formulas follow from $\dot g_i=-(N-1)\dot d_i$ and $g_i-d_i=y_i$ (for $i=2$, $g_2+(N-1)d_2$ is constant on $[T_0,\infty)$ by Step 5, and its value differs from the initial value $Nc_0$ by at most $1$, the amount accumulated on $[0,T_0]$). The properties of the state at $T_0$ are item 1 of~\cref{lemma:scalar_grad}, and $\dy_2\geq0$ is read off the system.
\end{proof}

Before analyzing the system in~\cref{coro:parameter_aggre}, we need the following lemma to characterize the main growth rate. It is applied to Filippov solutions, so the differential equations are only required to hold almost everywhere, and the coefficient bounds only from some time $t_0$ on.
\begin{lemma}[Logarithm growth lemma]\label{lemma:log_growth}
    Let $t_0\geq1$ and let $X,Y:[t_0,\infty)\to\sR$ be locally absolutely continuous and satisfy, for a.e.\ $t\geq t_0$,
    \begin{align}
        X'&=a(t)e^{Y-X},\\
        Y'&=b(t)e^{-Y}-\widetilde a(t)e^{Y-X}+r(t),
    \end{align}
    with measurable coefficients such that, for a.e.\ $t\geq t_0$,
    \begin{align}
        &0<a_-\le a(t)\le a_+<\infty,\qquad
        0<\widetilde a_-\le \widetilde a(t)\le \widetilde a_+<\infty,\\
        &0<\beta_-\leq \beta(t):=b(t)+e^{Y(t)}r(t)\le\beta_+<\infty .
    \end{align}
    Set $\Theta:=e^{2Y-X}$ and
    \[
        \Theta_-:=\frac{2\beta_-}{a_++2\widetilde a_+},\qquad \Theta_+:=\frac{2\beta_+}{a_-+2\widetilde a_-},\qquad c:=\min\{\Theta(t_0),\Theta_-\},\qquad C:=\max\{\Theta(t_0),\Theta_+\}.
    \]
    Then for all $t\geq t_0$,
    \begin{align}
        c\le e^{2Y(t)-X(t)}\le C,\qquad
        e^{X(t_0)/2}+\tfrac{a_-\sqrt c}{2}(t-t_0)\;\le\; e^{X(t)/2}\;\le\; e^{X(t_0)/2}+\tfrac{a_+\sqrt C}{2}(t-t_0).
    \end{align}
    In particular $X,Y\to\infty$ and
    \begin{align}
         X(t)=2\log t+O(1),
     \qquad
     Y(t)=\log t+O(1),
    \end{align}
    where the $O(1)$ terms depend only on $t_0$, $X(t_0)$, $Y(t_0)$ and the coefficient bounds.
\end{lemma}

\begin{proof}
    Set $P=e^X$ and $S=e^Y$, so $\Theta=S^2/P>0$ and, a.e.\ on $[t_0,\infty)$,
    \begin{align}
     P'=aS,
     \qquad
     S'=\beta-\widetilde a\,\frac{S^2}{P},
     \qquad
     (\log\Theta)'=2Y'-X'=\frac1S\bigl\{2\beta-(a+2\widetilde a)\Theta\bigr\}.
    \end{align}
    Since $S>0$, we have $\Theta'<0$ a.e.\ on $\{\Theta>\Theta_+\}$ and $\Theta'>0$ a.e.\ on $\{\Theta<\Theta_-\}$: $\Theta$ is pushed downward when it is large and upward when it is small. If $\Theta(t_1)>C$ for some $t_1>t_0$, let $t_2=\sup\{t\in[t_0,t_1]:\Theta(t)\le C\}$; then $\Theta(t_2)=C$, $\Theta>C\ge\Theta_+$ on $(t_2,t_1]$, and $\Theta(t_1)=\Theta(t_2)+\int_{t_2}^{t_1}\Theta'\le C$, a contradiction. Hence $\Theta\le C$, and symmetrically $\Theta\ge c>0$. Next, $(\sqrt P)'=P'/(2\sqrt P)=\tfrac12 a\sqrt\Theta\in[\tfrac12a_-\sqrt c,\tfrac12a_+\sqrt C]$ a.e., and integrating from $t_0$ gives the two-sided bound on $e^{X/2}=\sqrt P$. Finally $X=2\log\sqrt P$ and $Y=\tfrac12\log\Theta+\log\sqrt P$, and $\sqrt{P(t)}/t\in[\tfrac14 a_-\sqrt c,\ \sqrt{P(t_0)}+\tfrac12 a_+\sqrt C]$ for $t\ge 2t_0$, which proves the claim.
\end{proof}

With the dynamical system in~\cref{coro:parameter_aggre} and the logarithm growth lemma~\cref{lemma:log_growth}, we establish the convergence results in the following theorem.
\begin{theorem}\label{thm:four_parameter_log_convergence}
    Consider the four-dimensional discontinuous dynamical system of~\cref{coro:parameter_aggre}, with Filippov solutions on the switching surfaces $u=0$ and $v=0$, along the trajectory of~\cref{lemma:scalar_grad} for $t\geq T_0$; by item 1 of that lemma, $u\geq0$, $x_2>0$, $y_1>0$ and $y_2\geq0$ hold along it. Then the following holds, where ``bounded'' and $O(1)$ refer to constants that do not depend on $t$ but may depend on $N$, $K$ and the state at time $T_0$:
    \begin{enumerate}
        \item The positive part $v_+ = (y_1 - y_2)_+$ is bounded for $t\geq T_0$.
        \item $x_1,x_2,y_1,y_2 \to \infty$ as $t\to \infty$.
        \item The differences $2y_1 - x_1$ and $x_1 - x_2$ are upper bounded for $t\geq T_0$.
        \item $x_1,x_2 = 2\log(\alpha t)+O(1)$ and $y_1,y_2=\log(\alpha t)+O(1)$, where the $O(1)$ terms do not depend on $\alpha$.
    \end{enumerate}
\end{theorem}

\begin{proof}
    Throughout the proof we shift time so that the starting time $T_0$ becomes $0$; thus $u(0)\geq0$, $x_2(0)> 0$, $y_1(0)>0$, $y_2(0)\geq0$, and, by item 1 of~\cref{lemma:scalar_grad}, $u\geq0$, $x_2>0$, $y_1>0$, $y_2\geq0$ for all $t\geq0$ (we nevertheless re-derive $u\geq0$ below, since only the four-variable system is used from now on). The right-hand side of the system is bounded (every fraction below is at most $1$), so Filippov solutions are Lipschitz; all barrier arguments below use only that the relevant quantity is absolutely continuous and that its derivative has a definite sign on an open region, hence they hold for every Filippov solution. We also note that the $x$-equations do not contain $\ind_{v\geq0}$ and the $y$-equations do not contain $\ind_{u\geq0}$, so $u'$ is single-valued on the open set $\{u<0\}$ and $v'$ is single-valued on $\{v>0\}$ and on $\{v<0\}$, whatever happens on the other switching surface.
    Define
    \begin{align}
         G=&\frac{e^{y_1}+N-2}{D_1},
         \qquad
         H=\frac{e^{y_1}}{D_1}, \qquad E=e^{u_+ + v_+}\\
         B_i=&\frac{e^{y_1}}{D_{2,i}},
         \qquad \qquad
         C_i=\frac{e^{y_1}+N-K-1}{D_{2,i}},
         \qquad \qquad
         A_i=\frac{(i-1)E}{D_{2,i}},\\
         Q=&\frac{E}{D_4},
         \qquad \qquad
         R=\frac{N-K-1}{D_4},
         \qquad \qquad
         M=\frac{N-1}{D_3}.
    \end{align}
Also set
\begin{align}
 \rho=\frac{N}{N-1}.
\end{align}
Since multiplying the vector field by the positive constant $\alpha$ only rescales time,
we keep $\alpha$ in the formulas but note that the solution with learning rate $\alpha$ at time $t$ coincides with the solution for $\alpha=1$ at time $\alpha t$; hence all constants below are independent of $\alpha$ once the growth is expressed in terms of $\alpha t$.

When $u<0$, all $\1_{u\ge0}$ terms vanish and all $\1_{u<0}$ terms are active.  Hence
\begin{align}
        x_2'=0,
\end{align}
whereas
\begin{align}
 x_1'=\frac{\alpha}{N}\left[
 G+\sum_{i=1}^K(C_i+A_i)
 \right]>0.
\end{align}
Therefore $u'=x_1'-x_2'>0$ whenever $u<0$.  Since $u(0)\geq0$, the Filippov forward
barrier gives
\begin{align}
        {u(t)=x_1(t)-x_2(t)\ge0.}
\end{align}
Indeed, suppose $u(t_1)<0$ for some $t_1>0$ and let $t_0=\sup\{t<t_1:\,u(t)\ge 0\}$. Then $u(t_0)=0$ and $u<0$ on $(t_0,t_1]$; since $x_1',x_2'$ do not involve $\ind_{v\ge0}$, on the open set $\{u<0\}$ every Filippov solution satisfies $u'=x_1'>0$ a.e., so $u(t_1)>u(t_0)=0$, a contradiction. Consequently,
\begin{align}
        u_+=u,\quad \forall t\geq 0 .
\end{align}
The same one-sided barrier argument is used repeatedly below without further comment.

We first prove that $v_+$ is upper-bounded.
\paragraph{Upper bound of $v_+$.} 
In order to prove
\begin{align}
        {v_+(t)=(y_1(t)-y_2(t))_+\le C.}
\end{align}
It is enough to consider the open region $v>0$, where $E=e^{u+v}$ and
\begin{align}
\begin{aligned}
 v'
 &=\frac{\alpha}{N-1}\left[
 K M+R-H-\sum_{i=1}^K B_i
 -2\sum_{i=1}^K A_i-KQ
 \right]\le \frac{\alpha}{N-1}(KM+R-KQ).
\end{aligned}
\end{align}

To get a negative gradient $v'$, we split and discuss two cases by fixing a constant $L \geq 1$.

Case 1: $e^{y_2-u}\le L$.

 Since $y_1=y_2+v$ and $E=e^{u+v}$,
\begin{align}
 KQ=\frac{Ke^{u+v}}{Ke^{u+v}+e^{y_1}+N-K-1}
 =\frac{K}{K+e^{y_2-u}+(N-K-1)e^{-u-v}},
\end{align}
so $KQ$ has a positive lower bound, \emph{i.e.}, $KQ\ge c_L:=K/(L+N-1)>0$ (here we used $u+v\geq0$).  Meanwhile, for some constant $C=C_{N,K}$, we have
\begin{align}
        KM+R\le Ce^{-y_1}+Ce^{-u-v} = Ce^{-y_2 - v}+Ce^{-u-v} \le 2Ce^{-v},
\end{align}
where the last step uses $y_2\geq0$ and $u\geq 0$.
Thus $v'<0$ once $v$ is larger than a constant depending on $L$.

Case 2: $e^{y_2-u}>L.$ 

Put
\begin{align}
        S:=e^{y_2-u}.
\end{align}
Then we have $S>L$.  Since $v>0$ and $u\ge0$, we have
\begin{align}
        E=e^{u+v}\ge e^v,
        \qquad
        e^{y_1}=e^{y_2+v}=SE .
\end{align}
Consequently
\begin{align}
        D_4=K E+e^{y_1}+N-K-1
            =E\left(K+S+(N-K-1)E^{-1}\right).
\end{align}
The negative term coming from $D_4$ therefore satisfies, once $L\ge1$,
\begin{align}
        KQ
        =\frac{K E}{D_4}
        =\frac{K}{K+S+(N-K-1)E^{-1}}
        \ge \frac{K}{(2N)S}=:\frac{c_Q}{S} ,
\end{align}
where we used $S>L\ge1$, $K\le N-1$, and $E^{-1}\le1$, so
$K+S+(N-K-1)E^{-1}\le 2NS$.

On the other hand, the two positive terms in the upper bound for $v'$ are much smaller:
\begin{align}
        KM
        =K\frac{N-1}{e^{y_1}+N-1}
        \le K\frac{N-1}{SE},
\end{align}
and, since $D_4\ge e^{y_1}=SE$,
\begin{align}
        R=\frac{N-K-1}{D_4}\le \frac{N-K-1}{SE}.
\end{align}

Thus
\begin{align}
        KM+R\le \frac{C_{N,K}}{SE}
              \le \frac{C_{N,K}e^{-v}}{S}.
\end{align}

Choose $v_1=v_1(N,K)$ so large that
\begin{align}
        C_{N,K}e^{-v_1}\le \frac{c_Q}{2}.
\end{align}

Then for every $v\ge v_1$ and every point in the case $S>L$,
\begin{align}
        KM+R\le \frac{c_Q}{2S}\le \frac12 KQ.
\end{align}

Returning to the differential inequality, we have
\begin{align}
        v'
        \le \frac{\alpha}{N-1}(KM+R-KQ)
        \le -\frac{\alpha}{2(N-1)}KQ<0 .
\end{align}
This proves that in the whole region $e^{y_2-u}>L$, the positive $y$-gap is pushed
strictly downward whenever $v\ge v_1$.

Combining the two cases, there exists $C_v$ such that
\begin{align}
        v>C_v\quad\Longrightarrow\quad v'<0.
\end{align}
Since $\{v>C_v\}$ lies strictly inside the open region $\{v>0\}$, where $v'$ is single-valued, the one-sided barrier argument gives
\begin{align}
        v_+ = (y_1-y_2)_+\le C = \max\{ v(0), C_v\}, \quad \forall t\geq 0.
\end{align}

\paragraph{Prove $x_1,x_2,y_1,y_2 \to \infty$ as $t\to \infty$.}  First observe that, for every Filippov selection (the $\ind_{u\geq0}$-terms cancel in the combination $x_1+(N-1)x_2$),
\begin{align}
    (x_1 + (N-1)x_2)' &= \frac{\alpha}{N}\Bigl[G+\sum_{i=1}^K(C_i+A_i)\Bigr]\geq \frac{\alpha}{N} G \\
    &\geq  \frac{\alpha}{N(N-2)}\frac{e^{y_1}}{e^{x_1} + e^{y_1}  +1} \geq \frac{\alpha}{N(N-2)}\frac{1}{e^{x_1+(N-1)x_2} + 2},
\end{align}
where we used $e^{x_1}+e^{y_1}+N-2\leq (N-2)(e^{x_1}+e^{y_1}+1)$, $e^{y_1}\geq 1$ and $x_2\geq0$. Writing $w=x_1+(N-1)x_2$, this gives $(e^{w}+2w)'\geq c$, hence $w\geq \log(ct)-C$ for large $t$, i.e., $x_1 + (N-1)x_2  \gtrsim \log(t)$. Since $x_1\geq x_2$, we conclude that $\lim_{t\to\infty} x_1 = +\infty$. Therefore $\lim_{t\to\infty} D_1 = +\infty,\lim_{t\to\infty} D_{2,i} = +\infty$, and consequently $H\to0$ and $B_i\to0$ as long as $y_1$ stays bounded.

Next we show that $y_1$ is unbounded. Note that $y_1'$ contains the negative term $-\ind_{v\geq0}\sum_iA_i$, which need not be small at this stage, because $A_i\to0$ only once $x_2\to\infty$; we therefore consider $y_1+y_2$, whose derivative does not depend on $\ind_{v\geq0}$ and hence is the same for every Filippov selection on $v=0$:
\begin{align}
 (y_1+y_2)'=\frac{\alpha}{N-1}\Bigl[KM+R-H-\sum_{i=1}^K B_i+KQ\Bigr]
 \ \ge\ \frac{\alpha}{N-1}\Bigl[KM-H-\sum_{i=1}^K B_i\Bigr].
\end{align}
Suppose $y_1\le B$ for all $t$. Since $y_2'=0$ on $\{v<0\}$, $y_2$ can increase only while $y_2\le y_1\le B$; hence $y_2(t)\le\max\{y_2(0),B\}$ for all $t$ (if $y_2(t_1)>\max\{y_2(0),B\}$, let $t_0<t_1$ be the last time with $y_2\le\max\{y_2(0),B\}$; on $(t_0,t_1]$ we have $y_2>B\ge y_1$, i.e.\ $v<0$, so $y_2'=0$ a.e.\ there, a contradiction). Thus $y_1+y_2$ is bounded. On the other hand $M\ge (N-1)/(e^B+N-1)=:c_B>0$ depends only on $B$, while $H+\sum_i B_i\le e^{B}\bigl(1/D_1+\sum_i 1/D_{2,i}\bigr)\to 0$ because $x_1\to\infty$. Hence $(y_1+y_2)'\ge \frac{\alpha}{N-1}\frac{Kc_B}{2}>0$ for all large $t$, so $y_1+y_2\to\infty$, a contradiction. Therefore $\limsup_t y_1=+\infty$. Since $y_2\ge y_1-v_+\ge y_1-C$, also $\limsup_t y_2=+\infty$, and since $y_2$ is non-decreasing, $\lim_t y_2=+\infty$.

For $x_2$, suppose it is bounded. Since $x_1\to\infty$, $u>0$ for all large $t$, and on $\{u>0\}$ we have $x_2'=\frac{\alpha}{N(N-1)}[\sum_iA_i+KQ]$ while $y_2'=\eta\frac{\alpha}{N-1}[\sum_iA_i+KQ]\le N x_2'$ for every Filippov weight $\eta\in[0,1]$ of $\ind_{v\geq0}$. Integrating, $x_2(t)\geq x_2(T)+(y_2(t)-y_2(T))/N\to+\infty$, which contradicts the boundedness assumption. Therefore, $\lim_t x_2(t)\to +\infty$.

For $y_1$, $\forall B >0$, we prove that there exists a time $t_0$ such that $y_1\geq B$ for all $t\geq t_0$. On the region $\{v<0,\ y_1\le B\}$ there is no $A_i$-term in $y_1'$, and
\begin{align}
    y_1'=\frac{\alpha}{N-1}\Bigl[KM+R-H-\sum_iB_i+KQ\Bigr]\geq \frac{\alpha}{N-1}\Bigl[Kc_B-e^B\Bigl(\frac1{D_1}+\sum_i\frac1{D_{2,i}}\Bigr)\Bigr]\geq c>0
\end{align}
for all $t\geq t_1$, where $t_1$ is chosen, using $x_1\to\infty$, so that the bracket is at least $Kc_B/2$. Since $\lim_t y_2=+\infty$, pick $t_2\ge t_1$ with $y_2(t_2)>B$. For $t\ge t_2$, whenever $y_1(t)\le B$ we have $v(t)\le B-y_2(t_2)<0$ because $y_2$ is non-decreasing, so the state lies in $\{v<0,\,y_1\le B\}$ and $y_1'\ge c>0$ there. Hence $y_1\ge B$ for all $t\ge t_0:=t_2+(B-y_1(t_2))_+/c$, by the barrier argument. Thus $\lim_t y_1(t)\to +\infty$.

\paragraph{Upper bound of $2y_1 - x_1$.}

To analyze the behavior of $2y_1 - x_1$, we define new variables
\begin{align}
        s=e^u=e^{x_1-x_2},
        \qquad
        Z=e^{2y_1-x_1}.
\end{align}
The desired gap estimates are
\begin{align}
        s\le C,
        \qquad
        Z\le C.
\end{align}
Indeed, $s\le C$ is equivalent to $x_1-x_2\le C$, and $Z\le C$ is equivalent to
$2y_1-x_1\le C$.

To prove the upper bound, we use a Lyapunov argument. Define the Lyapunov product
\begin{align}
        W=Zs^p.
\end{align}

Choose
\begin{align}
       {\;3-\frac1N<p<1+\frac{2N}{N-1}.\;}
\end{align}

The interval is nonempty because
\begin{align}
        1+\frac{2N}{N-1}-\left(3-\frac1N\right)
        =\frac{2}{N-1}+\frac{1}{N}=\frac{3N-1}{N(N-1)}>0.
\end{align}

We discuss the regions where $u>0$ or $u=0$ separately.

\paragraph{Region 1: $u>0$.}

We have $\1_{u\ge0}=1$ and $\1_{u<0}=0$.  Thus
\begin{align}
 x_1'&=\frac{\alpha}{N}\left[G+\sum_{i=1}^K C_i-KQ\right],\\
 x_2'&=\frac{\alpha}{N(N-1)}\left[\sum_{i=1}^K A_i+KQ\right].
\end{align}

Therefore
\begin{align}
\begin{aligned}
 \frac{s'}s
 &=u'=\frac{\alpha}{N}\left[
 G+\sum_{i=1}^K C_i
 -\frac1{N-1}\sum_{i=1}^K A_i
 -\frac{KN}{N-1}Q
 \right].
\end{aligned}
\end{align}

Since $v$ may not be positive and can hit the sliding boundary, we introduce the Filippov parameter $0\le\eta\le1$, where $\eta$
is the convex value of $\1_{v\ge0}$.  Then every admissible velocity on $u>0$ satisfies
\begin{align}
 y_1'=\frac{\alpha}{N-1}\left[
 KM+R-H-\sum_{i=1}^K B_i
 -\eta\sum_{i=1}^K A_i
 +(1-\eta)KQ
 \right].
\end{align}

Consequently
\begin{align}
\begin{aligned}
 \frac{Z'}Z
 &=2y_1'-x_1'\\
 &=\frac{\alpha}{N}\Bigg[
 2\rho KM+2\rho R-2\rho H-G
 -\sum_{i=1}^K\{2\rho B_i+C_i\} 
 -2\rho\eta\sum_{i=1}^K A_i
 +K\{1+2\rho(1-\eta)\}Q
 \Bigg],
\end{aligned}
\end{align}
where $\rho=N/(N-1)$.  Therefore
\begin{align}
\begin{aligned}\label{eq:Lyapunov_grad}
 \frac{W'}W
 =  \frac{Z'}Z + p\frac{s'}{s} =&\frac{\alpha}{N}\Bigg[
 2\rho KM+2\rho R+\bigl((p-1)G-2\rho H\bigr)
 +\sum_{i=1}^K\bigl(pC_i-C_i-2\rho B_i\bigr) \\&
 -\left(2\rho\eta+\frac{p}{N-1}\right)\sum_{i=1}^K A_i
 +K\left(1+2\rho(1-\eta)-\frac{pN}{N-1}\right)Q
 \Bigg].
\end{aligned}
\end{align}

The coefficient of $Q$ is largest when $\eta=0$, and the lower bound
$p>3-1/N$ is exactly
\begin{align}
        1+2\rho-\frac{pN}{N-1}<0.
\end{align}

The coefficient of $\sum_i A_i$ is strictly negative.  The indexed $D_{2,i}$ contribution
can be written as
\begin{align}
 pC_i-C_i-2\rho B_i
 =\frac{(p-1)(N-K-1)+(p-1-2\rho)e^{y_1}}{D_{2,i}}.
\end{align}
Since $p<1+2\rho$, the coefficient of $e^{y_1}$ is strictly negative.  The
same calculation for $D_1$ gives
\begin{align}
 (p-1)G-2\rho H
 =\frac{(p-1)(N-2)+(p-1-2\rho)e^{y_1}}{D_1},
\end{align}
which has the same structure.

We now estimate each term in~\cref{eq:Lyapunov_grad}.

Define
\begin{align}
        a=e^{x_1},\qquad b=e^{y_1},\qquad s=e^u .
\end{align}

Since $v_+$ is bounded, there are constants $0<c_E<C_E<\infty$,
depending only on the previously obtained $v_+$ bound, such that
\begin{align}
        c_Es\le E=se^{v_+}\le C_Es .
\end{align}
Put
\begin{align}
        \delta_y:=1+2\rho-p>0,
        \qquad
        \delta_Q:=p\rho-(1+2\rho)>0,
        \qquad
        c_A:=\frac{p}{N-1}>0 .
\end{align}
The two strict inequalities follow exactly from the choice
$3-1/N<p<1+2N/(N-1)$.

Starting from the exact formula for $W'/W$, we separate the terms with a fixed negative
coefficient from the terms with a constant numerator.  For the $D_1$-term,
\begin{align}
 (p-1)G-2\rho H
 =\frac{(p-1)(N-2)-\delta_y b}{D_1}
 \le -\delta_y H+C_{N,K,p}\frac1{D_1}.
\end{align}
For every $i$,
\begin{align}
 pC_i-C_i-2\rho B_i
 =\frac{(p-1)(N-K-1)-\delta_y b}{D_{2,i}}
 \le -\delta_y B_i+C_{N,K,p}\frac1{D_{2,i}}.
\end{align}
Moreover the coefficient of $\sum_i A_i$ is at most $-c_A$, and the coefficient of
$KQ$ is maximal at $\eta=0$, where it equals $-K\delta_Q$.  Therefore, after enlarging
constants, every Filippov velocity in the region $u>0$ satisfies
\begin{align}
\begin{aligned}\label{eq:app_4.1}
 \frac{N}{\alpha}\frac{W'}W
 &\le \mathcal P-\mathcal N,\\
 \mathcal P&:= C\left(M+R+\frac1{D_1}+\sum_{i=1}^K\frac1{D_{2,i}}\right),\\
 \mathcal N&:= c\left(H+\sum_{i=1}^K B_i+\sum_{i=1}^K A_i+Q\right),
\end{aligned}
\end{align}
where $\mathcal P$ collects the terms with constant numerators and $\mathcal N$ the terms with a fixed negative coefficient (the letter $N$ is reserved for the number of vertices). Thus it is enough to show that $\mathcal N$ dominates $\mathcal P$ whenever
\begin{align}
        W=Zs^p=\frac{b^2s^p}{a}
\end{align}
is large.

We record the elementary estimates used below; recall $a=e^{x_1}\geq1$, $b=e^{y_1}\geq1$, $s=e^u\geq1$ and $E\asymp s$. Since $K-i+1\geq1$ in every $D_{2,i}$ and $N-K-1\geq 4$,
\begin{align}
        M\le \frac{N-1}{b+1},
        \qquad
        R\le \frac{N-K-1}{b+1},
        \qquad
        \frac1{D_1}+\sum_i\frac1{D_{2,i}}
        \le \frac{K+1}{b+1},
        \qquad\text{hence}\qquad
        \mathcal P\le\frac{C}{b+1}.
\label{eq:app_4.5}
\end{align}
For the negative part,
\begin{align}
        Q=\frac{E}{KE+b+N-K-1}
        \asymp \frac{s}{s+b+1},
\\
        H=\frac{b}{a+b+N-2}\asymp \frac{b}{a+b+1},
        \qquad
        B_1=\frac{b}{Ka+b+N-K-1}\asymp \frac{b}{a+b+1},
\label{eq:app_4.2}
\end{align}
so that, dropping the nonnegative terms $A_i$ and $B_i$, $i\geq 2$,
\begin{align}
        \mathcal N\ge c\left(\frac{b}{a+b+1}+\frac{s}{s+b+1}\right).
\label{eq:app_4.4}
\end{align}

We now show that for $b\geq1$, $s\geq1$ and $W=b^2s^p/a$,
\begin{align}
        \frac{b}{a+b+1}+\frac{s}{s+b+1}\ \ge\ \frac14\min\left\{1,\frac{W^{1/(p+1)}}{b+1}\right\}.
\label{eq:app_4.6}
\end{align}
Indeed, if $s\geq b+1$ the second term is at least $\frac12$; if $a\leq b+1$ the first term is at least $b/(2(b+1))\geq\frac14$. Otherwise $a>b+1\geq s$, so $b/(a+b+1)\geq b/(3a)=W/(3bs^p)$ and $s/(s+b+1)\geq s/(3(b+1))$, and by the weighted AM--GM inequality, for all $s>0$,
\begin{align}
        \frac{W}{bs^p}+\frac{s}{b+1}
        \ \ge\ \Bigl(\frac{W}{b}\Bigr)^{\frac1{p+1}}(b+1)^{-\frac{p}{p+1}}
        \ \ge\ \frac{W^{1/(p+1)}}{b+1},
\label{eq:app_4.7}
\end{align}
where we used $b\leq b+1$. This proves~\cref{eq:app_4.6}.

Combining~\cref{eq:app_4.1,eq:app_4.5,eq:app_4.4,eq:app_4.6}, every Filippov velocity in the region $u>0$ satisfies
\begin{align}
        \frac{N}{\alpha}\frac{W'}{W}\ \le\ \frac{C}{b+1}-c'\min\left\{1,\frac{W^{1/(p+1)}}{b+1}\right\}.
\label{eq:app_4.8}
\end{align}
Since $\lim_t y_1=+\infty$, we can pick a time $t_0$ such that $b=e^{y_1}\geq B_0:=2C/c'$ for all $t\geq t_0$. Then, for $t\geq t_0$, the right-hand side of~\cref{eq:app_4.8} is negative whenever $W\geq C_W:=(2C/c')^{p+1}$: if the minimum equals $1$ this follows from $b\geq B_0$, and otherwise $c'W^{1/(p+1)}/(b+1)\geq 2C/(b+1)$.

We conclude that for $t\geq t_0$, in the region $u>0$, $W'<0$ whenever $W\geq C_W$; in particular a trajectory cannot cross the level surface $W=C_W$ outward while $u>0$.

\paragraph{Region $u=0$.}
We now analyze the $u=0$ region. The gradient is
\begin{align}
        \frac{\dd}{\dd t}\log W
        =2y_1'-x_1'+p(x_1'-x_2')
        =2y_1'+(p-1)x_1'-p x_2'
\end{align}
for every admissible Filippov boundary velocity.

At $u=0$ we have
\begin{align}
        s=1,
        \qquad
        W=Z=\frac{e^{2y_1}}{e^{x_1}},
        \qquad
        E=e^{v_+}.
\end{align}

Since $v_+$ is bounded, there are constants
$0<c_E<C_E<\infty$ such that
\begin{align}
        c_E\le E\le C_E
\end{align}
on the whole boundary $u=0$.  Introduce the Filippov parameter
$0\le\theta\le1$ for the $u$-discontinuity, so that $\theta$ is the convex
value of $\1_{u\ge0}$ and $1-\theta$ is the convex value of $\1_{u<0}$.
As before, $0\le\eta\le1$ denotes the convex value of $\1_{v\ge0}$.
Then every admissible boundary velocity satisfies
\begin{align}
\begin{aligned}
 x_1'
 &=\frac{\alpha}{N}\left[
 G+\sum_{i=1}^K C_i+(1-\theta)\sum_{i=1}^K A_i-\theta KQ
 \right],\\
 x_2'
 &=\frac{\alpha\theta}{N(N-1)}\left[
 \sum_{i=1}^K A_i+KQ
 \right],\\
 y_1'
 &=\frac{\alpha}{N-1}\left[
 KM+R-H-\sum_{i=1}^K B_i
 -\eta\sum_{i=1}^K A_i+(1-\eta)KQ
 \right].
\end{aligned}
\end{align}
Therefore, with $\rho=N/(N-1)$,
\begin{align}
\begin{aligned}
 \frac{N}{\alpha}\frac{\dd}{\dd t}\log W
 &=2\rho(KM+R)
   +\bigl((p-1)G-2\rho H\bigr)
   +\sum_{i=1}^K\bigl((p-1)C_i-2\rho B_i\bigr)\\
 &\quad
   +\left((p-1)(1-\theta)-2\rho\eta-\frac{p\theta}{N-1}\right)
       \sum_{i=1}^K A_i\\
 &\quad
   +K\left(2\rho(1-\eta)-(p-1)\theta-\frac{p\theta}{N-1}\right)Q .
\end{aligned}
\label{eq:app_4.14}
\end{align}
The two denominator terms involving $D_1$ and $D_{2,i}$ have the same favorable
structure as in the interior calculation.  Namely, with
\begin{align}
        a=e^{x_1},\qquad b=e^{y_1},\qquad
        \delta_y=1+2\rho-p>0,
\end{align}
we have
\begin{align}
 (p-1)G-2\rho H
 =\frac{(p-1)(N-2)-\delta_y b}{D_1}
 \le -\delta_y H+C\frac1{D_1},
\label{eq:app_4.15}
\end{align}
and, for every $i$,
\begin{align}
 (p-1)C_i-2\rho B_i
 =\frac{(p-1)(N-K-1)-\delta_y b}{D_{2,i}}
 \le -\delta_y B_i+C\frac1{D_{2,i}}.
\label{eq:app_4.16}
\end{align}
The coefficients of the $A_i$ and $Q$ terms in \cref{eq:app_4.14} are uniformly bounded above,
independently of $\theta$ and $\eta$.  Hence \cref{eq:app_4.14,eq:app_4.15,eq:app_4.16} imply
\begin{align}
\begin{aligned}
 \frac{N}{\alpha}\frac{\dd}{\dd t}\log W
 &\le
 C\left(M+R+\frac1{D_1}+\sum_{i=1}^K\frac1{D_{2,i}}
       +\sum_{i=1}^K A_i+Q\right)
 -c\left(H+\sum_{i=1}^K B_i\right).
\end{aligned}
\label{eq:app_4.17}
\end{align}
This estimate is valid for every Filippov choice of $\theta$ and $\eta$ on $u=0$.

We now show that the negative term in \cref{eq:app_4.17} dominates whenever $W=Z$ is large.  Since $y_1(t)\to\infty$, choose $t_0$ so large that
\begin{align}
        b=e^{y_1(t)}\ge B_0
\end{align}
for all $t\ge t_0$, where $B_0$ will be fixed below.  On $u=0$ and for $t\ge t_0$,
$E$ is bounded and $b$ is large, so
\begin{align}
        M\le \frac{C}{b},
        \qquad
        R\le \frac{C}{b},
        \qquad
        Q\le \frac{C}{b}.
\label{eq:app_4.18}
\end{align}
Moreover, since every $D_{2,i}$ contains the term $(K-i+1)a$ and
$K-i+1\ge1$, while $E$ is bounded,
\begin{align}
        \sum_{i=1}^K A_i\le \frac{C}{a+b+1},
\label{eq:app_4.19}
\end{align}
and
\begin{align}
        \frac1{D_1}+\sum_{i=1}^K\frac1{D_{2,i}}
        \le \frac{C}{a+b+1}.
\label{eq:app_4.20}
\end{align}
Finally,
\begin{align}
        H+\sum_{i=1}^K B_i
        \ge c\frac{b}{a+b+1}.
\label{eq:app_4.21}
\end{align}
Combining \cref{eq:app_4.17}--\cref{eq:app_4.21}, every admissible boundary velocity satisfies
\begin{align}
        \frac{N}{\alpha}\frac{\dd}{\dd t}\log W
        \le \frac{C}{b}+\frac{C}{a+b+1}
             -c\frac{b}{a+b+1}.
\label{eq:app_4.22}
\end{align}
It remains only to check the two possible relative sizes of $a$ and $b$.

\smallskip
\noindent\textbf{Boundary case 1: $a\le L(b+1)$.}
Choose $L>1$ fixed.  Then
\begin{align}
        \frac{b}{a+b+1}\ge c_L>0.
\end{align}
Thus the negative term in \cref{eq:app_4.22} is bounded below by a positive constant, whereas the
positive terms are $O(b^{-1})$.  After increasing $B_0$ if necessary, \cref{eq:app_4.22} gives
\begin{align}
        \frac{\dd}{\dd t}\log W<0
\end{align}
throughout this boundary subcase.

\smallskip
\noindent\textbf{Boundary case 2: $a>L(b+1)$.}
Here
\begin{align}
        \frac{b}{a+b+1}\asymp \frac{b}{a}.
\end{align}
Since $W=Z=b^2/a$ on $u=0$, this becomes
\begin{align}
        \frac{b}{a}=\frac{W}{b}.
\end{align}
Also $(a+b+1)^{-1}\le C/a\le C/b$.  Hence \cref{eq:app_4.22} yields
\begin{align}
        \frac{N}{\alpha}\frac{\dd}{\dd t}\log W
        \le \frac{C}{b}-c\frac{W}{b}.
\label{eq:app_4.23}
\end{align}
Therefore the boundary derivative is strictly negative in this subcase whenever
$W\ge C_\partial$, where $C_\partial$ depends only on the fixed parameters and on the bound for $v_+$.

The two boundary cases show that, for $t\ge t_0$, no Filippov velocity at $u=0$ can
point outward through a sufficiently large level surface of $W$.  More precisely, after
choosing $C_\partial$ and $B_0$ large enough,
\begin{align}
W\ge C_\partial
        \quad\Longrightarrow\quad
        \frac{\dd}{\dd t}\log W<0
\end{align}
for every admissible boundary velocity.  This is the required boundary version of the
Lyapunov estimate.

Combining the interior and boundary estimates: for a.e.\ $t\geq t_0$, $W(t)\geq C_\ast:=\max\{C_W,C_\partial\}$ implies $(\log W)'(t)<0$, whether $u(t)>0$ or $u(t)=0$. Since $W$ is absolutely continuous (the vector field is bounded, so $x_i,y_i$ are Lipschitz), the one-sided barrier argument gives $W(t)\leq\max\{W(t_0),C_\ast\}$ for all $t\geq t_0$, and $W$ is bounded on the compact interval $[0,t_0]$ by continuity. Hence
\begin{align}
        {W(t)=Z(t)s(t)^p\le C}\qquad\forall t\geq0 .
\end{align}
Since $s\ge1$, we immediately obtain
\begin{align}
        {Z(t)\le C.}
\end{align}
Equivalently,
\begin{align}
        {2y_1(t)-x_1(t)\le C_Z}\qquad\forall t\geq 0 .
\end{align}

\paragraph{Upper bound of $x_1 -x_2$.}

We now prove that the upper barrier for
\begin{align}
        s=e^{u}=e^{x_1-x_2}
\end{align}
is finite.  We already know $u\ge0$, hence $s\ge1$.  It remains to exclude
escape of $s$ to infinity.

It suffices to consider the region $u>0$.  There $E=e^{u+v_+}=s e^{v_+}$, and the
formulae for $x_1'$ and $x_2'$ give
\begin{align}
\begin{aligned}
 u'
 &=x_1'-x_2' \\
 &=\frac{\alpha}{N}\left[
 G+\sum_{i=1}^K C_i
 -\frac1{N-1}\sum_{i=1}^K A_i
 -\frac{KN}{N-1}Q
 \right]
 \leq \frac{\alpha}{N}\Bigl[G+\sum_{i=1}^K C_i-Q\Bigr],
\end{aligned}
\label{eq:app_5.1}
\end{align}
where we dropped $A_i\geq0$ and used $KN/(N-1)\geq1$. Since $y_1\geq0$ and $x_2\geq0$, we have $b\geq1$ and $s\leq a$, so for the positive terms $G$ and $C_i$,
\begin{align}
        G\le C\frac{b+1}{a+b+1}\leq C \frac{b}{a},
        \qquad
        C_i\le C\frac{b+1}{a+b+1}
        \le C\frac{b}{a}.
\end{align}
For $Q$, since $E\leq e^{C_v}s$ we have the lower bound
\begin{align}
    Q \geq C' \frac{s}{s+b+1}.
\end{align}
Therefore $u'<0$ as soon as
\begin{align}
    (K+1) C\frac{b}{a} < C' \frac{s}{s+b+1},
    \qquad\text{i.e.}\qquad
    (K+1)C\bigl(1+e^{y_1 - u} + e^{-u}\bigr) < C' e^{x_1-y_1}.
    \label{eq:u_sufficient}
\end{align}
We claim that~\cref{eq:u_sufficient} holds whenever
\begin{align}
   u\ \geq\ 2 y_1 - x_1 + C_2,\qquad C_2:=2\log\bigl(3(K+1)C/C'\bigr).
   \label{eq:u_log_condition}
\end{align}
Indeed, since $x_2\geq0$ we have $x_1\geq u$, so~\cref{eq:u_log_condition} implies $x_1\geq 2y_1-x_1+C_2$, i.e.\ $x_1-y_1\geq C_2/2$; hence $1+e^{-u}\leq 2\leq 2e^{-C_2/2}e^{x_1-y_1}$ and $e^{y_1-u}\leq e^{-C_2}e^{x_1-y_1}\leq e^{-C_2/2}e^{x_1-y_1}$, so the left-hand side of~\cref{eq:u_sufficient} is at most $3(K+1)Ce^{-C_2/2}e^{x_1-y_1}=C'e^{x_1-y_1}$, with strict inequality after enlarging $C_2$ slightly. (Note that the two extra terms $1+e^{-u}$ cannot be dropped: without the input $x_1\geq u$ the implication fails.)

Therefore, we argue that
\begin{align}
    u(t) \leq U(t):= \max \Bigl\{u(0),\ \sup_{0\leq\tau\leq t}\bigl(2y_1(\tau) - x_1(\tau)\bigr) + C_2,\ \epsilon \Bigr\},
\end{align}
where $\epsilon$ is a small positive constant. To prove this, suppose the opposite and consider the first violation time
\begin{align}
    t_0 = \inf \{ t:u(t)>U(t)\}.
\end{align}
Since $u$ and $U$ are continuous, $u(t_0) = U(t_0)$, and $U$ is non-decreasing. Since $u(t_0)\geq\epsilon>0$, the state lies in the open region $\{u>0\}$, where $u$ is continuously differentiable (the $x$-equations contain no $v$-indicator and $E=e^{u+v_+}$ is continuous), and $u(t_0)\geq 2y_1(t_0)-x_1(t_0)+C_2$ gives $u'(t_0)<0$ by~\cref{eq:u_sufficient,eq:u_log_condition}. Hence $u(t)<u(t_0)=U(t_0)\leq U(t)$ on a right neighborhood of $t_0$, contradicting the definition of $t_0$.

Further, since $2y_1 - x_1\leq C_Z$ for all $t\geq0$, we conclude that $u$ is also upper-bounded, \emph{i.e.},
\begin{align}
    0\leq x_1 -x_2 \leq C_u:=\max\{u(0),C_Z+C_2,\epsilon\},\quad \forall t\geq0 .
\end{align}

\paragraph{Logarithm growth rate}
The averaged $x$-variable for the system is
\begin{align}
        X=\frac{x_1+(N-1)x_2}{N},
        \qquad
        Y=y_1.
\end{align}
Since $0\leq u=x_1-x_2\leq C_u$,
\begin{align}
        X=x_1+O(1)=x_2+O(1),\qquad\text{more precisely}\qquad X=x_1-\tfrac{N-1}{N}u .
\end{align}
Also $u$ and $v_+$ are bounded, so
\begin{align}
        1\leq E=e^{u+v_+}\leq C_E .
\end{align}
The estimate $Z=e^{2Y-x_1}\le C$ gives
\begin{align}
        e^{2Y}\le Ce^{x_1}\asymp Ce^X,
\end{align}
and in particular, since $Y\to\infty$,
\begin{align}\label{eq:b_over_a}
        e^{y_1-x_1}=Ze^{-y_1}\le Ce^{-Y}\to0,\qquad
        e^{-X}\le e^{\frac{N-1}{N}C_u}e^{-x_1}\le Ce^{-2Y}\le Ce^{-Y}.
\end{align}
Consequently $D_1=e^{x_1}(1+o(1))$ and $D_{2,i}=(K-i+1)e^{x_1}(1+o(1))$; this is where the bound $2y_1-x_1\leq C$ enters.

Writing $\theta\in[0,1]$ for the Filippov weight of $\ind_{u\geq0}$ (so $\theta=1$ on $\{u>0\}$),
\begin{align}
\begin{aligned}
 (x_1+(N-1)x_2)'
 &=\frac{\alpha}{N}\left[
 G+\sum_{i=1}^K C_i+(1-\theta)\sum_{i=1}^K A_i-\theta KQ
 +\theta\sum_{i=1}^K A_i+\theta KQ
 \right]\\
 &=\frac{\alpha}{N}\left[G+\sum_{i=1}^K(C_i+A_i)\right],
\end{aligned}
\end{align}
independently of $\theta$; the same identity therefore holds for the Filippov boundary field at $u=0$.  Hence
\begin{align}
        X'=\frac{\alpha}{N^2}\left[G+\sum_{i=1}^K(C_i+A_i)\right].
\end{align}
Because $E\leq C_E$ and $D_{2,i}\geq e^{x_1}$, the $A_i$ terms are $O(e^{-x_1})=O(e^{-X})$. Moreover
\begin{align}
        G+\sum_{i=1}^K C_i
        =e^{y_1}\Bigl(\frac{1}{D_1}+\sum_{i=1}^K\frac{1}{D_{2,i}}\Bigr)+O(e^{-x_1})
        =a_0(t)e^{Y-X}+O(e^{-X}),
\end{align}
where, by~\cref{eq:b_over_a}, $a_0(t)=e^{X-x_1}e^{x_1}\bigl(1/D_1+\sum_i1/D_{2,i}\bigr)$ is bounded above and below by positive constants once $Y$ is large: $e^{X-x_1}=e^{-\frac{N-1}{N}u}\in[e^{-\frac{N-1}{N}C_u},1]$ and $e^{x_1}(1/D_1+\sum_i1/D_{2,i})\to 1+\sum_{i=1}^K\frac{1}{K-i+1}$.  The
$O(e^{-X})$ term is absorbed into the coefficient because $e^{-X}\leq e^{Y-X}$ for $Y\geq0$.  Thus, for all large $t$,
\begin{align}
       {X'=a(t)e^{Y-X},}
\end{align}
with
\begin{align}
        0<a_-\le a(t)\le a_+<\infty.
\end{align}

Next, writing $\eta\in[0,1]$ for the Filippov weight of $\ind_{v\geq0}$ (so $\eta=1$ on $\{v>0\}$ and $\eta=0$ on $\{v<0\}$), the $Y=y_1$ equation is
\begin{align}
 Y'=\frac{\alpha}{N-1}\left[
 KM+R-H-\sum_{i=1}^K B_i
 -\eta\sum_{i=1}^K A_i+
 (1-\eta)KQ
 \right].
\end{align}
Here
\begin{align}
        KM=K\frac{N-1}{e^Y+N-1}=:b_0(t)e^{-Y},
\end{align}
with $b_0(t)\in[K(N-1)/2,K(N-1)]$ once $e^Y\geq N-1$.  Also
\begin{align}
        H+\sum_{i=1}^K B_i=e^{Y}\Bigl(\frac1{D_1}+\sum_{i=1}^K\frac1{D_{2,i}}\Bigr)=\widetilde a_0(t)e^{Y-X},
\end{align}
where $\widetilde a_0(t)=e^{X}\bigl(1/D_1+\sum_i1/D_{2,i}\bigr)$ is bounded above and below by positive constants for large $t$, by the same computation as for $a_0$.  The
remaining terms satisfy
\begin{align}
        &0\leq R\le (N-K-1)e^{-Y},
        \qquad
        0\leq (1-\eta)KQ\le KC_Ee^{-Y},
\\
        &0\leq \eta\sum_i A_i\le Ce^{-x_1}\leq Ce^{-2Y},
\end{align}
using $D_4\geq e^{Y}$, $E\leq C_E$, $D_{2,i}\geq e^{x_1}$ and~\cref{eq:b_over_a}. Therefore
\begin{align}
        {Y'=b(t)e^{-Y}-\widetilde a(t)e^{Y-X}+r(t),}
        \qquad
        r(t):=\frac{\alpha}{N-1}\Bigl[R+(1-\eta)KQ-\eta\sum_iA_i\Bigr],
\end{align}
where $b(t)=\frac{\alpha}{N-1}b_0(t)$, $\widetilde a(t)=\frac{\alpha}{N-1}\widetilde a_0(t)$ satisfy
\begin{align}
        0<b_-\le b(t)\le b_+<\infty,
        \qquad
        0<\widetilde a_-\le\widetilde a(t)\le\widetilde a_+<\infty,
\end{align}
and
\begin{align}
        |r(t)|\le Ce^{-Y}.
\end{align}
We now check that the effective input $\beta(t)=b(t)+e^Yr(t)$ is eventually bounded above and below by positive constants. We have
\[
 e^{Y}r=\frac{\alpha}{N-1}\Bigl[\frac{(N-K-1)e^{Y}}{D_4}+(1-\eta)\frac{KEe^{Y}}{D_4}-\eta\,e^{Y}\sum_iA_i\Bigr].
\]
The first two terms lie in $[0,\,N-K-1+KC_E]$, since $D_4\ge e^{Y}$ and $1\le E\le C_E$; they are of the same order as $b$, but they are nonnegative. For the last term, $e^{Y}\sum_iA_i\le Ce^{Y-x_1}=CZe^{-Y}\le Ce^{-Y}\to0$ by the bound $Z\le C$. Since $b(t)\geq \frac{\alpha}{N-1}\frac{K(N-1)}{2}=\frac{\alpha K}{2}$ once $e^{Y}\ge N-1$, we obtain
\begin{align}
        \frac{\alpha K}{4}\le\beta(t)\le \alpha K+\frac{\alpha}{N-1}\bigl(N-K-1+KC_E\bigr)
\end{align}
for all $t\geq t_0$, where $t_0$ is chosen, using $Y\to\infty$, so that all the above estimates hold for $t\geq t_0$.

All hypotheses of~\cref{lemma:log_growth} are therefore satisfied on $[t_0,\infty)$ (the equations hold a.e.\ and for every Filippov selection, since all bounds are uniform in $\theta,\eta\in[0,1]$). Applying the lemma to $\widetilde X(t)=X(t+t_0)$, $\widetilde Y(t)=Y(t+t_0)$ and using $\log(t+t_0)=\log t+O(1)$ for $t\geq1$ gives
\begin{align}
        X(t)=2\log t+O(1),
        \qquad
        Y(t)=\log t+O(1).
\end{align}
Since $X=x_1+O(1)=x_2+O(1)$ and $Y=y_1$, we obtain
\begin{align}
        x_1(t),x_2(t)=2\log t+O(1),
        \qquad
        y_1(t)=\log t+O(1).
\end{align}
Concerning the dependence on $\alpha$: if $(x_1,x_2,y_1,y_2)$ solves the system with learning rate $\alpha$, then $t\mapsto(x_1,x_2,y_1,y_2)(t/\alpha)$ solves the system with $\alpha=1$ from the same initial state. Hence $(x_1,x_2,y_1,y_2)(t)=(\hat x_1,\hat x_2,\hat y_1,\hat y_2)(\alpha t)$, where the hats denote the solution for $\alpha=1$, and all the $O(1)$ terms above can be taken independent of $\alpha$ when the growth is expressed in terms of $\alpha t$:
\begin{align}
        x_1(t),x_2(t)=2\log (\alpha t)+O(1),
        \qquad
        y_1(t)=\log (\alpha t)+O(1).
\end{align}

Since $y_1-y_2\le C$, hence
\begin{align}
        y_2(t)\ge y_1(t)-C=\log (\alpha t)+O(1).
\end{align}
For the matching upper bound on $y_2$, we use an elementary barrier.  Let
\begin{align}
        m(t)=\max_{0\le r\le t}y_1(r),
        \qquad
        h(t)=(y_2(t)-m(t))_+.
\end{align}
Both $y_2$ and $m$ are Lipschitz (the vector field is bounded), hence so is $h$. At almost every time at which $h(t)>0$, we have $y_2(t)>m(t)\ge y_1(t)$, hence $v<0$ strictly, so the state is in the open region $\{v<0\}$ where $y_2'=0$ for every Filippov selection, while $m'(t)\ge0$.  Therefore $h'(t)\le0$
for almost every such time, and consequently $h$ is non-increasing (if $h(t_2)>h(t_1)$ for some $t_1<t_2$, let $t_\ast=\sup\{t\in[t_1,t_2]:h(t)\le h(t_1)\}$; then $h>h(t_1)\geq0$ on $(t_\ast,t_2]$ and $h(t_2)=h(t_\ast)+\int_{t_\ast}^{t_2}h'\le h(t_1)$, a contradiction). Hence $y_2(t)\le m(t)+h(0)$, where $h(0)=(y_2(0)-y_1(0))_+$ is a constant (it vanishes for the trajectory of~\cref{lemma:scalar_grad}, along which $v>0$ at the initial time). Thus
\begin{align}
        y_2(t)\le\max_{0\le r\le t}y_1(r)+h(0)
        \le \log(\alpha t)+O(1)\qquad\text{for }t\geq t_0,
\end{align}
where we used that $y_1(r)\leq\log(\alpha r)+C\leq\log(\alpha t)+C$ for $t_0\leq r\leq t$ and that $y_1$ is bounded on $[0,t_0]$. Therefore
\begin{align}
        {y_2(t)=\log (\alpha t)+O(1).}
\end{align}
This completes the proof.
\end{proof}

By~\cref{thm:four_parameter_log_convergence}, we can prove~\cref{thm:learning_dynamics}.

\begin{proof}[Proof of~\cref{thm:learning_dynamics}]
By~\cref{lemma:scalar_grad,coro:parameter_aggre,thm:four_parameter_log_convergence}, for $t\geq T_0$ we have
\begin{align}
    &a_1\equiv c_0,\qquad (b_2)_+=(b_2-b_1)_+=0,\qquad d_2\leq 0,\\
    &a_2,b_1 =2\log(\alpha t) + O(1),\qquad
    g_1 - d_1 =\log(\alpha t) + O(1),\qquad
    g_2 - d_2 = \log(\alpha t) + O(1),
\end{align}
and $0\leq u=a_2-b_1\leq C_u$, $v_+=(g_1-d_1-g_2+d_2)_+\leq C_v$ for constants $C_u,C_v$ given by items 1 and 3 of~\cref{thm:four_parameter_log_convergence}. The two-sided bound $| g_1 - d_1 -g_2 + d_2 | \leq C$ follows from the last two growth rates for $t\geq t_0$ and from continuity on $[0,t_0]$; this proves the first part of the theorem. By~\cref{eq:gd_recovery} the individual parameters satisfy
\begin{align}
    g_1,g_2 = \frac{N-1}{N}  \log(\alpha t) + O(1),\qquad
    d_1,d_2 = - \frac{1}{N}  \log(\alpha t) + O(1);
\end{align}
in particular $d_2\to-\infty$. All constants below depend only on $N,K,c_0,c_0'$.

\emph{Logits on a general graph.} Consider any directed graph on $N$ vertices, any valid nonterminated DFS trace, and the current node $n_t$ with stack parent $p$ (if $n_t\neq n_\Root$). We read the post-ReLU logits off~\cref{tab:forward_head,tab:backward_head}, using $b_2=0$, $(d_2)_+=0$ (so $b_2 d_v^{\text{in,total}}=0$ and $(b_2-b_1 m_v)_+=0$), $(g_1-d_1)_+=g_1-d_1$ and the dynamic readout, which sets the logit of $n_t$ to $0$ while keeping $n_t$ in the softmax with weight $1$:
\begin{itemize}
    \item an unvisited child $v$: logit $(a_2+b_2(d_v^{\text{in,total}}-1))_+=a_2$, since $b_2=0$ for $t\geq T_0$;
    \item a visited child of $n_t$ that is not the parent: logit at most $(a_2-b_1)_+ + v_+\leq u_++v_+\leq C_u+C_v$;
    \item the parent $p$: logit $(g_1-d_1+d_2\ind[p=n_\Root])_+ + \ind[(n_t\to p)\in\gE]\,(a_2-b_1m_p)_+$, which lies between $(g_1-d_1+d_2\ind[p=n_\Root])_+$ and $g_1-d_1+u_+$;
    \item every other vertex, including the root when it is not the parent and $n_t$ itself: logit $0$.
\end{itemize}
Let $m$ be the number of unvisited children, $m_1$ the number of visited children other than the parent, and $m_2\leq N$ the number of remaining vertices. The softmax normalizer $\mathcal Z$ therefore satisfies
\begin{align}\label{eq:gen_normalizer}
    \mathcal Z\ \leq\ m\,e^{a_2}+m_1 e^{C_u+C_v}+e^{g_1-d_1+C_u}+m_2
    \ \leq\ m\,e^{a_2}+ C e^{g_1-d_1}+C .
\end{align}

\emph{Case 1: $n_t$ has unvisited children} ($m\geq1$). The admissible set $C_t$ consists of the unvisited children, each with logit $L_v=a_2$, and the normalizer is $\mathcal Z=\sum_{v\in C_t}e^{L_v}+\mathcal Z'$ with $\mathcal Z'\leq Ce^{g_1-d_1}+C$ by~\cref{eq:gen_normalizer}. Hence
\begin{align}
    - \log\Bigl(\sum_{v\in C_t} p_{\theta}(v)\Bigr) &=\log\Bigl(1+\frac{\mathcal Z'}{\sum_{v\in C_t}e^{L_v}}\Bigr)
    \leq \log\Bigl(1+\frac{Ce^{g_1 - d_1} + C}{m\,e^{a_2}}\Bigr)\\
    &\leq 2Ce^{g_1-d_1-a_2}
    = O\bigl((\alpha t)^{-1}\bigr),
\end{align}
where we used $g_1-d_1>0$ (so $C\leq Ce^{g_1-d_1}$) and $a_2-(g_1-d_1)=\log(\alpha t)+O(1)$.

\emph{Case 2: no unvisited child and $p\neq n_\Root$.} The admissible set is $\{p\}$. Writing $L_p$ for the parent's logit, $L_p\geq g_1-d_1$, and the normalizer is $e^{L_p}+\mathcal Z''$ with $\mathcal Z''\leq m_1e^{C_u+C_v}+m_2\leq C$ (the parent's own term, including its forward contribution when $(n_t\to p)\in\gE$, is in the numerator as well). Hence
\begin{align}
    - \log p_{\theta}(p)=\log\Bigl(1+\frac{\mathcal Z''}{e^{L_p}}\Bigr)\leq \log\bigl(1+Ce^{-(g_1-d_1)}\bigr)=O\bigl((\alpha t)^{-1}\bigr).
\end{align}

\emph{Case 3: no unvisited child and $p= n_\Root$.} Now the parent's logit is $L_p\geq g_1-d_1+d_2$, and the same computation gives
\begin{align}
    - \log p_{\theta}(p)\leq \log\bigl(1+Ce^{-(g_1-d_1+d_2)}\bigr)
    = O\Bigl(e^{-\frac{N-1}{N}\log(\alpha t)}\Bigr)=O\bigl((\alpha t)^{-\frac{N-1}{N}}\bigr),
\end{align}
since $g_1-d_1+d_2=\frac{N-1}{N}\log(\alpha t)+O(1)$. This rate cannot be improved: the masked current node contributes $1$ to the normalizer, so $-\log p_\theta(p)\geq\log(1+e^{-L_p})$ with $L_p=g_1-d_1+d_2$ exactly when $(n_t\to n_\Root)\notin\gE$, and the graph with the single edge $(n_\Root\to a)$ and the trace $n_\Root,a,n_\Root$ realizes this case.

A valid DFS trace on $N$ vertices has at most $2(N-1)$ steps, and in every step one of the three cases applies (the terminated state, in which the root has no unvisited children, carries no loss term). Summing over the steps and taking the supremum over graphs and traces,
\begin{align}
    \gL_{\text{gen}} \leq 2(N-1)\,C\,(\alpha t)^{-\frac{N-1}{N}}
\end{align}
for all $t\geq t_\ast/\alpha$, where $t_\ast$ is such that all the $O(1)$ estimates above are in force. This proves~\cref{eq:gen_rate}.
\end{proof}

%% file: content/appendix/addition_exp.tex
\subsection{Additional experimental results}
\label{add_exp}

\paragraph{Protocol.}
All runs use float64. The fixed first layer is implemented by its exact output on each token type (\cref{eq:hidden_state_after_first_layer}); the second layer, its value and output matrices, and the dynamic readout are those of~\cref{sec:training_dynamics}, and the softmax is taken over the $N$ vertex tokens. The reduced flow integrates the eight-scalar system of~\cref{lemma:equivalence_parameter} with an implicit solver in logarithmic time (relative tolerance $10^{-9}$), replacing each indicator $\ind_{x\geq0}$ by the steep ramp $\min\{\max\{x/\epsilon+1,0\},1\}$ with $\epsilon=10^{-6}$; changing $\epsilon$ by two orders of magnitude moves the trajectory by $O(\epsilon)$, and plain gradient descent with $\eta=0.05$ agrees with it to within $0.012$ on $[1,5\times10^5]$. Full-matrix training uses $B=128$ random vertex permutations per gradient step, i.e., a stochastic version of the loss~\cref{eq:permutation_loss}; the effective scalars of a matrix are the mean of its diagonal and the mean of its off-diagonal entries. The implementation was checked against a direct autograd implementation, against finite differences, against the closed-form node logits of~\cref{tab:forward_head,tab:backward_head} on random graphs, and, for $N=7$, against the exact average over all $5040$ permutations.

The test set contains, for $N=16$, $300$ random graphs (Erd\H{o}s--R\'enyi graphs with edge probability in $\{0.1,0.2,0.4\}$, random out-trees with up to three extra edges, and ``root-fan'' graphs in which the root starts several chains and the target is unreachable or on a late chain), together with the full canonical trace and the single-edge graph $\{(n_\Root\to a)\}$. For each graph we enumerate all valid DFS prefixes (or sample $50$ traces when there are more), giving $64{,}325$ distinct states, of which $5{,}625$ are forward steps at a node that has an edge back to its stack parent.

\begin{figure}[tbp!]
  \begin{center}
    \begin{minipage}[t]{0.49\columnwidth}
      \centering
      \includegraphics[width=\linewidth]{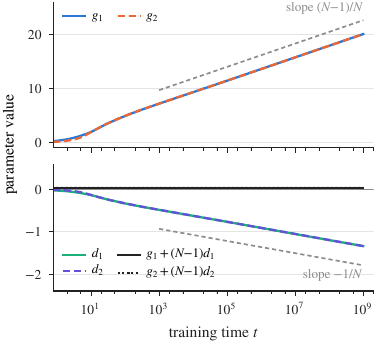}
    \end{minipage}\hfill
    \begin{minipage}[t]{0.49\columnwidth}
      \centering
      \includegraphics[width=\linewidth]{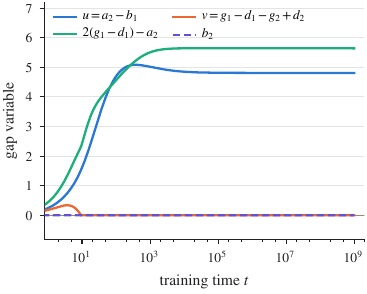}
    \end{minipage}
    \caption{\textbf{Left:} the individual parameters $g_1,g_2$ (fitted slopes $0.935$; predicted $(N-1)/N=0.9375$) and $d_1,d_2$ (fitted slopes $-0.062$; predicted $-1/N$), and the conserved combinations $g_i+(N-1)d_i$. \textbf{Right:} the gap variables $u=a_2-b_1$, $v=g_1-d_1-g_2+d_2$ and $2(g_1-d_1)-a_2$, which stay bounded as in~\cref{thm:four_parameter_log_convergence}, together with $b_2$, which is frozen at $0$ after the initial phase.}
    \label{fig:individual_gaps}
  \end{center}
\end{figure}

\paragraph{Individual parameters and gap variables.}
\Cref{fig:individual_gaps} shows the quantities behind~\cref{eq:gd_recovery} and~\cref{thm:four_parameter_log_convergence}. The combination $g_1+(N-1)d_1$ is conserved to $5\times10^{-12}$ along the reduced flow; $g_2+(N-1)d_2$ moves by $0.019$ during the initial phase, while $d_2>0$, and is constant afterwards, as predicted. The gap $u=a_2-b_1$ is nonnegative throughout and converges to about $4.8$ after a transient maximum of $5.1$ at $t\approx400$; $v$ rises to $0.34$ during the initial phase and then converges to $0$, i.e., the trajectory slides on the switching surface $v=0$ for all $t\gtrsim10^2$; and $2(g_1-d_1)-a_2$ converges to about $5.6$.

\begin{figure}[tbp!]
  \begin{center}
    \begin{minipage}[t]{0.49\columnwidth}
      \centering
      \includegraphics[width=\linewidth]{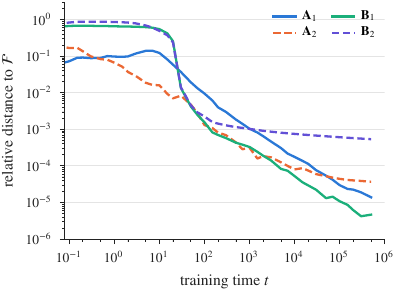}
    \end{minipage}\hfill
    \begin{minipage}[t]{0.49\columnwidth}
      \centering
      \includegraphics[width=\linewidth]{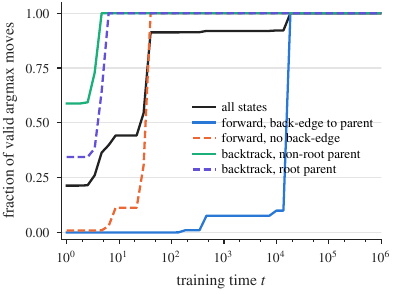}
    \end{minipage}
    \caption{\textbf{Left:} relative distance $\|\mX-(\bar x_{\rm diag}\mI+\bar x_{\rm off}\Eoff)\|_F/\|\mX\|_F$ of each trained matrix from the invariant subspace during full-matrix training with $B=128$ sampled permutations per step. \textbf{Right:} fraction of test states at which the argmax is a valid DFS move, by step type.}
    \label{fig:subspace_accuracy}
  \end{center}
\end{figure}

\paragraph{Invariant subspace under sampled permutations.}
\Cref{prop:symmetric_weight} concerns the exact permutation-averaged loss. With $B=128$ sampled permutations per step (\cref{fig:subspace_accuracy}, left), the iterate leaves the subspace during the initial phase, when the dynamics slide along the switching surfaces $b_2=0$ and $v=0$: the relative distance of $\mB_1$ and $\mB_2$ reaches $0.7$--$0.9$ and the effective scalars differ from the reduced flow by up to $0.9$ at $t\approx15$. The iterate then returns: the distance is below $10^{-2}$ from $t\approx70$ on and at most $6\times10^{-4}$ at $t=5\times10^5$, where the effective scalars agree with the reduced flow to within $0.03$. The excursion is a sampling effect: rerunning the early phase with $B\in\{512,2048\}$ reduces it only partially, whereas $B\geq8192$ keeps the distance below $0.03$ and the scalars within $0.007$ of the reduced flow at all times, and reducing $\eta$ tenfold at $B=128$ has no effect.

\begin{figure}[tbp!]
  \begin{center}
    \centerline{\includegraphics[width=0.5\columnwidth]{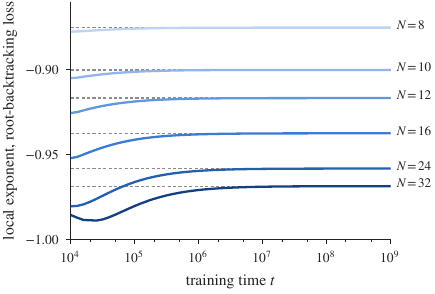}}
    \caption{Local log--log slope $\mathrm{d}\log\gL/\mathrm{d}\log t$ of the root-backtracking loss for several $N$ (reduced flow). Dashed lines mark the predicted exponents $-(N-1)/N$.}
    \label{fig:exponent_vs_N}
  \end{center}
\end{figure}

\paragraph{Exponent versus $N$.}
The table in~\cref{sec:experiment} reports the exponent of the root-backtracking loss fitted on $t\in[10^6,10^9]$ for $N\in\{8,10,12,16,24,32\}$; every fit lies within $3\times10^{-4}$ of $-(N-1)/N$. \Cref{fig:exponent_vs_N} shows the local log--log slope of the same curves: it approaches $-(N-1)/N$ from below, and the transient lasts longer for larger $N$, because the $O(1)$ terms in $d_2=-\frac1N\log(\alpha t)+O(1)$ then take longer to become negligible.

\paragraph{Accuracy and back-edge states.}
\Cref{fig:subspace_accuracy} (right) shows when the argmax becomes a valid DFS move. Backtracking steps are correct from $t\approx5$, forward steps without an edge back to the parent from $t\approx40$, and forward steps with such an edge only from $t\approx1.8\times10^4$, when $b_1$ exceeds $g_1-d_1$; the overall accuracy is $0.91$ at $t=100$ and equals $1$ from $t\approx2\times10^4$ on, and both training procedures give the same curves. The back-edge states are also the ones that attain the supremum $\gL_{\text{gen}}$ over the whole observed range: a visited child of the current node that is also its stack parent receives votes from both heads, $g_1-d_1$ from the backtracking head and $a_2-b_1$ from the forward head, so its logit gap to the correct unexplored child is $b_1-(g_1-d_1)=\log(\alpha t)-u+O(1)$ and the loss is $\Theta(e^{u}/t)$ with $e^u\approx e^{5}$. Both effects are consequences of the two-head mechanism itself rather than of the training horizon.

\begin{figure}[tbp!]
  \begin{center}
    \centerline{\includegraphics[width=\columnwidth]{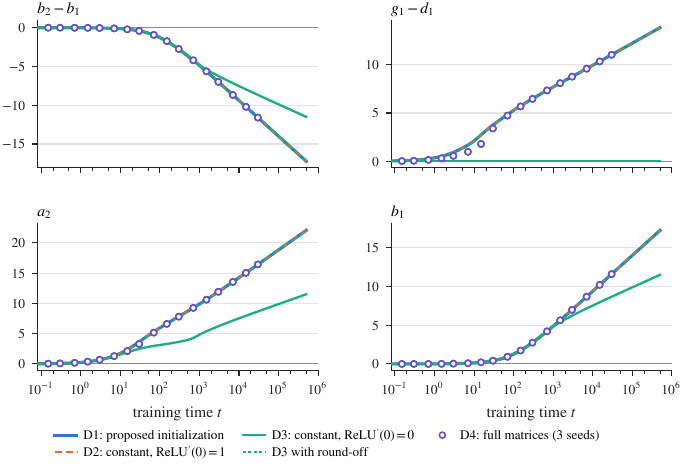}}
    \caption{Initialization ablation at $N=16$, $K=8$ (reduced flow by gradient descent with $\eta=0.05$; D4 is full-matrix training with three seeds). D1: the initialization~\cref{eq:initialization}. D2, D3: the all-constant initialization $c_0\mE$ with the conventions $\relu'(0)=1$ and $\relu'(0)=0$. With $\relu'(0)=0$ and exactly cancelling head sums, $g_1-d_1$ stays at $0$ forever and the backtracking head never activates; floating-point round-off in the head sums, or sampling noise (D4), releases it.}
    \label{fig:init_ablation}
  \end{center}
\end{figure}

\begin{figure}[tbp!]
  \begin{center}
    \centerline{\includegraphics[width=\columnwidth]{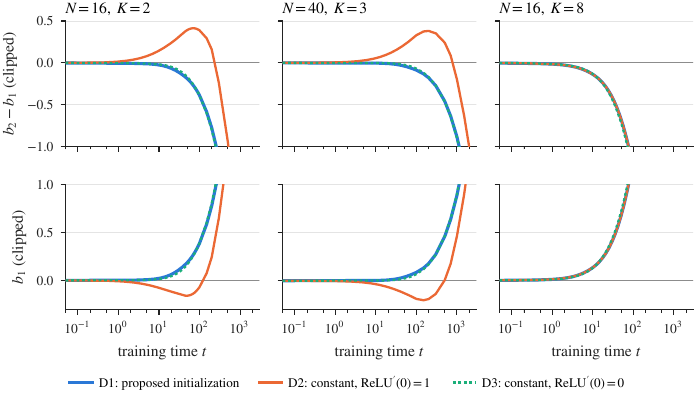}}
    \caption{The all-constant initialization $c_0\mE$ with $\relu'(0)=1$ at $(N,K)=(16,2)$, $(40,3)$, and $(16,8)$. When $N$ is large relative to $K$, the trajectory first leaves the switching surface $b_2=b_1$ towards $b_2>b_1$ and $b_1$ becomes negative, before re-entering the region $\{b_2<b_1,\ g_1>d_1\}$ at $t\approx300$ and $t\approx10^3$, respectively; at $(16,8)$ it enters the region immediately. The initialization~\cref{eq:initialization} stays in the region from $t=0$ in all three cases.}
    \label{fig:init_wrong_side}
  \end{center}
\end{figure}

\paragraph{Initialization.}
\Cref{fig:init_ablation,fig:init_wrong_side} examine the role of the initialization~\cref{eq:initialization}. The all-constant initialization $c_0\mE$ lies on the two switching surfaces $b_2=b_1$ and $g_1=d_1$, where the vector field is discontinuous, and its behaviour depends on the convention for $\relu'(0)$ and on $(N,K)$. With $\relu'(0)=0$ (the default of common autograd libraries) and exactly cancelling head sums, $g_1-d_1$ remains exactly $0$ and the backtracking head never activates; the forward parameters then grow only half as fast ($a_2,b_1\approx11.5$ at $t=5\times10^5$ against $22.0$ and $17.2$). With $\relu'(0)=1$ the outcome depends on $(N,K)$: at $(16,8)$ the trajectory enters the region $\{b_2<b_1,\ g_1>d_1\}$ at once, but at $(16,2)$ and $(40,3)$ it first exits towards $b_2>b_1$ with $b_1<0$ for a transient of several hundred time units, after which it approaches the trajectory from~\cref{eq:initialization} in $a_2,b_1,g_1-d_1$ but keeps a positive $b_2$ that decays only logarithmically slowly ($0.34$ at $t=5\times10^5$ for $(16,2)$), its gradient being $O(1/t)$. Across $60$ pairs $(N,K)$ with $8\leq N\leq64$ and $2\leq K\leq10$, the wrong-side exit occurs in all but one case in which $N>(K+2)^2/2$, and in none of the others. In practice, floating-point round-off or permutation sampling is enough to release the frozen case (\cref{fig:init_ablation}, D3 with round-off and D4), but the trajectory is then selected by numerical noise rather than by the gradient flow. The initialization~\cref{eq:initialization} avoids all of these effects: it starts inside the region that~\cref{lemma:scalar_grad} shows to be invariant and stays there from $t=0$ in every setting we tried.

\paragraph{Summary.}
The experiments agree with the training-dynamics theory in every quantitative prediction: the forward-head parameters grow at $2\log t$, the backtracking separation at $\log t$, the individual parameters at $\frac{N-1}{N}\log t$ and $-\frac1N\log t$, the gap variables remain bounded, and backtracking to the root decays at exactly $t^{-(N-1)/N}$ for $N$ between $8$ and $32$. They also show which parts of the statements are asymptotic: the supremum loss is dominated over all practical horizons by $t^{-1}$ terms with large constants, and sampled-permutation training approaches the invariant subspace only after the initial phase.